\pdfoutput=1
\documentclass[11pt,letterpaper]{article}

\usepackage{rit_arxiv}

\usepackage{cuted}

\usepackage{pifont}

\usepackage{array}
\usepackage{aliascnt}
\newtheorem{theorem}{Theorem}[section]
\newcommand{\sharedtheorem}[2]{%
  \newaliascnt{#1}{theorem}\newtheorem{#1}[#1]{#2}\aliascntresetthe{#1}}
\sharedtheorem{proposition}{Proposition}
\sharedtheorem{lemma}{Lemma}
\sharedtheorem{corollary}{Corollary}
\theoremstyle{definition}
\sharedtheorem{definition}{Definition}
\sharedtheorem{assumption}{Assumption}
\sharedtheorem{example}{Example}
\theoremstyle{remark}
\sharedtheorem{remark}{Remark}
\newcommand{\E}{\mathbb{E}}
\DeclareMathOperator{\Var}{Var}
\DeclareMathOperator{\Cov}{Cov}

\usepackage{colortbl}
\definecolor{vira9}{HTML}{F76902}
\definecolor{vira27}{HTML}{DE5E02}
\colorlet{vira9bg}{vira9!16}
\colorlet{vira27bg}{vira27!16}
\newcommand{\tok}[1]{{\fontsize{7.5}{8}\selectfont\color{black!60}\,(#1)}}
\newlength{\accw}\newlength{\tokw}\newlength{\numw}\newlength{\imgw}
\newcommand{\nb}[1]{\makebox[\numw][r]{#1}}
\newcommand{\cstrut}{\rule{0pt}{1.02em}}
\newcommand{\shaderow}[1]{\rowcolor{#1}\cstrut}
\definecolor{deltac}{HTML}{1A7F3C}
\newcommand{\dgain}[1]{\makebox[\numw][l]{\textcolor{deltac}{#1\%}}}
\definecolor{gainc}{HTML}{1A9850}
\definecolor{lossc}{HTML}{D73027}
\usepackage{listings}
\lstdefinestyle{code}{language=Python, keywordstyle=\bfseries, commentstyle=\color{black!55}}
\lstdefinestyle{prompt}{language={}, breakatwhitespace=true}
\usepackage{titletoc}

\makeatletter
\renewcommand\@makefntext[1]{\noindent\@makefnmark#1}
\makeatother

\title{ViRA: Toward Ultra-High-Resolution Visual Agents\\via Cross-Input Relative Policy Optimization}

\newcommand{\ArxivLogoRow}{%
  \PreprintLogo{rit}{18pt}\hfill\PreprintLogo{purdue}{18pt}\hfill\PreprintLogo{utah}{18pt}\hfill\PreprintLogo{nvidia}{18pt}}
\newcommand{\AuthorNames}{%
  \mbox{Huawei Lin\textsuperscript{1,\dag}}\quad
  \mbox{Jinting Liu\textsuperscript{1}}\quad
  \mbox{Tongyan Wang\textsuperscript{2}}\quad
  \mbox{Xuyang Chen\textsuperscript{1}}\\
  \mbox{Ruihua Shi\textsuperscript{1}}\quad
  \mbox{Zhichao Xu\textsuperscript{3}}\quad
  \mbox{Jiaxin Qin\textsuperscript{1}}\quad
  \mbox{Guoshenghui Zhao\textsuperscript{1}}\\
  \mbox{Baijian Yang\textsuperscript{2}}\quad
  \mbox{Yingjie Victor Chen\textsuperscript{2}}\quad
  \mbox{Tan Yu\textsuperscript{4}}\quad
  \mbox{Weijie Zhao\textsuperscript{1,\ddag}}}
\newcommand{\AuthorAffils}{%
  \mbox{\textsuperscript{1}Rochester Institute of Technology}\quad
  \mbox{\textsuperscript{2}Purdue University}\quad
  \mbox{\textsuperscript{3}University of Utah}\quad
  \mbox{\textsuperscript{4}NVIDIA}}
\newcommand{\AuthorNotes}{\textsuperscript{\dag}Project Lead\qquad\textsuperscript{\ddag}Corresponding Author: \href{mailto:wjz@cs.rit.edu}{wjz@cs.rit.edu}}
\newcommand{\ArxivRows}{%
  \ArxivRow{Code}{\href{https://github.com/huawei-lin/ViRA}{github.com/huawei-lin/ViRA}}%
  \ArxivRow{Data}{\href{https://huggingface.co/datasets/huaweilin/ViRA-Train}{huggingface.co/datasets/huaweilin/ViRA-Train}}%
  \ArxivRow{Models}{\href{https://huggingface.co/huaweilin/ViRA-9B}{huggingface.co/huaweilin/ViRA-9B}\quad
    \href{https://huggingface.co/huaweilin/ViRA-27B}{huggingface.co/huaweilin/ViRA-27B}}%
  \ArxivRow{Bench}{\href{https://huggingface.co/datasets/huaweilin/ViRA-Bench}{huggingface.co/datasets/huaweilin/ViRA-Bench}}}
\author{Huawei Lin, Jinting Liu, Tongyan Wang, Xuyang Chen, Ruihua Shi, Zhichao Xu, Jiaxin Qin, Guoshenghui Zhao, Baijian Yang, Yingjie Victor Chen, Tan Yu, Weijie Zhao}

\begin{document}
\maketitle

\begin{strip}
  \centering
  \includegraphics[width=\textwidth]{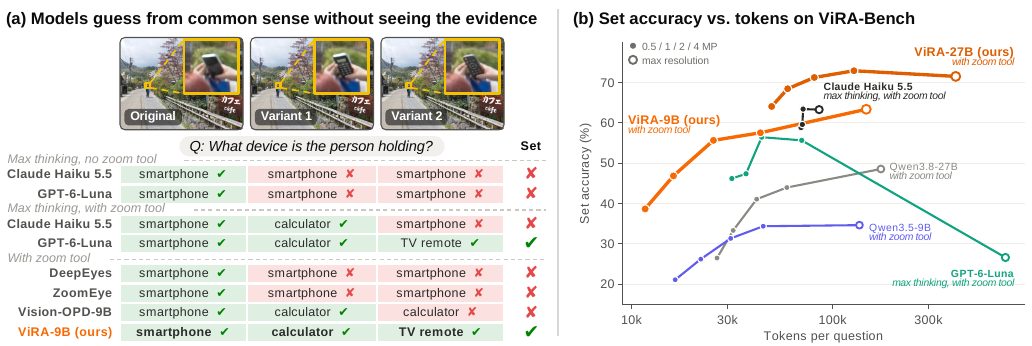}

  \captionof{figure}{\textbf{ViRA-27B leads at every budget.}
  (a)~Three versions of a 45-megapixel (MP) ViRA-Bench photo differ only in the device a woman holds up (yellow box).
  Rows give models' answers at 1\,MP budget, and \emph{Set} is correct only if all three are.
  (b)~Set accuracy is the share of sets with all versions correct, and tokens are summed over all turns and versions.
  Filled markers are pixel budgets of 0.5--4\,MP, which cap every image a model sees, and hollow markers are models' max resolution.
  Claude Haiku~5.5 has no 4\,MP marker as its max resolution is 3.8\,MP.
  \emph{Max thinking} is the highest thinking level, and \emph{with zoom tool}, a model draws a box to~see~that~region.}
  \label{fig:teaser}
\end{strip}

\begin{abstract}
Visual agents answer questions about ultra-high-resolution photos by zooming into the relevant regions.
However, existing agents often answer from common sense instead of reading the evidence.
Reinforcement learning (RL) on the final answer cannot correct this, because a guess earns the same reward as reading.
We propose Cross-Input Relative Policy Optimization (CrossRPO), which trains agents on \emph{contrast sets}.
A contrast set asks one question about several versions of an image that differ only in a small region with the evidence and have different answers, so only reading the evidence solves the whole set.
CrossRPO rewards solving the whole set, and we prove that it does not reinforce a set answered from common sense.
We build contrast sets for training (ViRA-Train) and evaluation (ViRA-Bench) by editing one small object in images of 21 to 120 megapixels, and we photograph further sets with real changes (ViRA-Bench-Real) to exclude editing traces. We train two Visual Reasoning Agents, ViRA-9B and ViRA-27B, with CrossRPO on only about 3,500 items of ViRA-Train each.
CrossRPO achieves 63.3\% set accuracy on ViRA-Bench at the maximum resolution, against 59.3\% for GRPO.
ViRA-9B is the most accurate open model of its size on all benchmarks, and ViRA-27B outperforms Claude Haiku~5.5 and GPT-6-Luna, both at their highest thinking level, on most benchmarks.
\end{abstract}

\flushbottom
\section{Introduction}
\label{sec:intro}

Photos from modern cameras have tens of megapixels (MP), but the detail that answers a question about them is often only a few dozen pixels wide, such as what a pedestrian down the street is holding.
Such photos exceed the input size of vision-language models.
For example, Qwen3.5 turns each patch of $32\times32$ pixels into one token, so a 45\,MP photo would require about 44,000 tokens.
Models therefore cap the input size, even with high-resolution encoders~\cite{li2024monkey,guo2024llavauhd,bai2025qwen25vl,bai2025qwen3vl}, and shrink larger photos to fit.
The detail shrinks with the photo and often becomes unreadable~\cite{wu2024vstar,wang2025hrbench,zhang2025mmerealworld}.
At the 3.8\,MP limit of Claude Haiku~5.5, a detail 50 pixels wide in a 45\,MP photo becomes about 15 pixels wide.
To recover such details, visual agents view the whole photo at low resolution and zoom into the regions they need~\cite{openai2025thinking,su2025thinkingsurvey}.
Some follow a fixed search procedure~\cite{shen2025zoomeye,li2025dyfo,zhang2025mllmsknow,wang2025rap,lin2025agentomni,lin2026museautoskill}, and others learn where to zoom~\cite{shao2024visualcot,zheng2025deepeyes,wang2025pixelreasoner,lai2025minio3,zhang2025thyme,hong2025deepeyesv2,zhang2025chainfocus,fan2025grit}.

\textbf{Correct without reading.} However, a correct answer does not show that the agent read the evidence, because common sense often gives the same answer~\cite{agrawal2018dont,niu2021counterfactual,tong2024eyes,li2023pope,guan2024hallusionbench,xu2025morethinking}.
For example, a person who holds up a small device as if taking a photo usually holds a smartphone.
An agent can therefore answer ``smartphone'' correctly without reading the device.
Figure~\ref{fig:teaser}a tests this with three versions of a 45\,MP street photo, in which a woman holds up a smartphone, a calculator or a TV remote.
The device fits in a square of 110 pixels, so a model must zoom in to read it.
Without a zoom tool, Claude Haiku~5.5 and GPT-6-Luna answer ``smartphone'' for all three versions.
The released agents DeepEyes~\cite{zheng2025deepeyes} and ZoomEye~\cite{shen2025zoomeye} give the same answer, although they can zoom.
Even with a zoom tool and its highest thinking level, Claude Haiku~5.5 still identifies the remote as a smartphone.
Being able to zoom therefore does not make an agent read the evidence.

\textbf{Existing methods.} The agents that learn where to zoom are trained with reinforcement learning (RL) on the final answer~\cite{guo2025deepseekr1,jin2025searchr1}, typically with Group Relative Policy Optimization (GRPO)~\cite{shao2024deepseekmath}.
This training cannot correct the guess, because on a single image, a guess earns the same reward as reading.
The guess is also easier, because it requires no zoom, so it becomes a shortcut that the reward cannot detect~\cite{geirhos2020shortcut,langosco2022goal,ye2026samereward,yao2026mmrewardhacking}.
Recent work therefore also rewards the zoom, such as a tool call~\cite{zheng2025deepeyes}, a crop that covers annotated evidence~\cite{zhao2026lfpc,dong2026toolsrl} or a call that improves the answer~\cite{peng2026countercredit,feng2026taco}.
These rewards credit viewing the evidence rather than reading it, and agents learn to call the zoom tool without using what it shows~\cite{hong2025deepeyesv2,wang2026illusion,zhu2026visualflip,ma2026vision}.
For example, on the TV remote version of Figure~\ref{fig:teaser}a, DeepEyes zooms onto the whole remote but writes that ``\textit{the zoomed-in image confirms that the person is holding a smartphone.}''
Given only this zoom in a new conversation, it answers ``\textit{TV remote.}''
Other methods mask the evidence and reward abstention or penalize an answer~\cite{xu2026defacto,liu2026evan}.
However, a masked image has no answer to read, so they cannot check whether the answer changes with the evidence.
Both still judge one image at a time.

\begin{figure*}[t]
  \centering
  \resizebox{\textwidth}{!}{\includegraphics{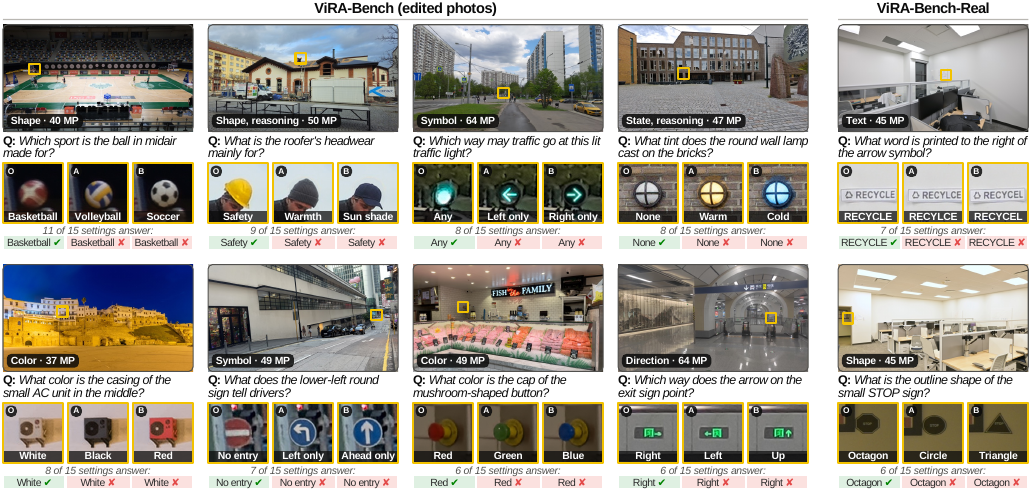}}
  \caption{\textbf{Examples of ViRA-Bench and ViRA-Bench-Real.}
  Each set shows the original photo (O, cut to 16:9) with the object in a yellow box.
  The insets show the object in O and the two variants (A, B), with the correct answers.
  Below each set are the original's answer and the number of baseline settings, out of 15 (Appendix~\ref{sec:supp-baseline-list}), that give this answer to all three versions at 1\,MP.
  Questions are shortened.}
  \label{fig:bench}
\end{figure*}

\textbf{Contrast sets.} On one image, reading and guessing give the same answer, so no reward on that image can separate them.
Several versions of the same photo can.
A \emph{contrast set}~\cite{gardner2020evaluating} asks the same question about several versions of a photo.
Only a small region with the evidence differs between them, so the versions have different answers, as in Figure~\ref{fig:teaser}a.
An agent that gives the same answer to every version can be correct on only one of them, so only reading the evidence can solve the whole set.
We call the share of sets that an agent solves its \textit{set accuracy}.
Contrast sets and similar minimal pairs have been used to evaluate vision-language models~\cite{goyal2017making,thrush2022winoground,li2024naturalbench} and to train them with supervised losses~\cite{kaushik2020learning,teney2020learning,chen2020counterfactual}.
We build such sets for zoom agents by editing one small object in real photos of 21 to 120\,MP.
ViRA-Train has 9,261 sets with $K=1$, 2 or 3 versions for training, and ViRA-Bench has 762 sets for evaluation.
In ViRA-Bench, the edited object covers a median of 0.0075\% of the photo.
We also photograph 78 sets in which only the real object changes between shots (ViRA-Bench-Real), so they involve no editing.
Figure~\ref{fig:bench} shows examples of both benchmarks, on which many of the models we test give the same answer to all three versions.

\textbf{CrossRPO.} GRPO samples several answers (rollouts) for one input and compares them with each other (Figure~\ref{fig:method}, top).
If it trains on each version of a contrast set separately, no reward depends on solving the whole set.
If it compares rollouts across versions (pooled GRPO), a guess that is correct on one version scores above the wrong answers on the others and is rewarded.
We therefore propose Cross-Input Relative Policy Optimization (CrossRPO, Figure~\ref{fig:method}, bottom).
It compares each rollout only with rollouts of the same version, so a guess that is always correct on one version receives no credit.
It then scales the result with a cross-input weight.
The weight grows when the other versions are answered correctly, because a correct answer on this version then solves the set.
CrossRPO applies to sets of any size, including single images, whereas recent methods for two inputs~\cite{huang2025taming,zhu2026comet,zou2026consistency,li2026cfcamo,du2026crpo} handle only pairs.
We prove that its update is unbiased and does not reinforce a set answered from common sense, and that the objectives of GRPO and of these two-input methods are special cases~of~its~objective.

\textbf{ViRA.} With CrossRPO, we train two Visual Reasoning Agents, ViRA-9B and ViRA-27B, on about 3,500 items of ViRA-Train each.
With the same data and reward, CrossRPO achieves 63.3\% set accuracy on ViRA-Bench at the maximum resolution, against 59.3\% for GRPO on each version separately and 52.6\% for pooled GRPO.
Training on each version separately, without the cross-input weight, achieves 58.7\%.
CrossRPO loses its gain over separate training when the cross-input weight comes from photos of other questions, so the gain depends on the contrast sets.
ViRA-9B zooms onto the object as often as the agent trained on each version separately, but it reads the object correctly more often.
ViRA-9B and ViRA-27B are the most accurate open agents of their sizes on ten and nine of the ten benchmarks we test, respectively.
On ViRA-Bench, ViRA-27B also outperforms all proprietary models we test at every resolution (Figure~\ref{fig:teaser}b).

\textbf{Contributions.} Our contributions are as follows.
\begin{itemize}
  \item We propose CrossRPO, which trains on sets of inputs of any size. We prove that its update is unbiased, that it does not reinforce a set answered from common sense, and that the objectives of GRPO and of recent two-input methods are special cases of its objective.
  \item We build ViRA-Train and ViRA-Bench, contrast sets made by editing one small object in photos of 21 to 120\,MP, and ViRA-Bench-Real, photographed sets in which only the real object changes.
  \item We train ViRA-9B and 27B with CrossRPO. Our experiments show that ViRA models outperform the models with similar size on most benchmarks. We will release all data, code and models.
\end{itemize}

\section{Related Work}
\label{sec:related}

\textbf{Zooming agents.}
Thinking-with-images agents~\cite{openai2025thinking,su2025thinkingsurvey} crop and zoom over several turns.
They are trained with supervised data~\cite{shao2024visualcot} or with RL on the final answer~\cite{zheng2025deepeyes,wang2025pixelreasoner,lai2025minio3,zhang2025thyme,hong2025deepeyesv2,zhang2025chainfocus,fan2025grit}.
Other methods search the image without training~\cite{shen2025zoomeye,li2025dyfo,zhang2025mllmsknow,wang2025rap} or learn to see fine details without zooming~\cite{wei2026zooming,yuan2026visionopd}.
Recent rewards credit crops that cover the evidence~\cite{zhao2026lfpc,dong2026toolsrl} or calls that outperform a random crop or no call~\cite{peng2026countercredit,feng2026taco}, but they still judge one image at a time.
Audits find that the returned crops often matter little~\cite{shao2026textcall,wang2026illusion,zhu2026visualflip} and that the gains come mostly from the model~\cite{ma2026vision}.
Except for PerceptRL~\cite{guan2026perceptrl}, these works compare with proprietary models that view the image once and have no zoom tool~\cite{wang2025pixelreasoner,wei2026zooming,wei2026zoomingv2,yuan2026visionopd,peng2026countercredit}, whereas we give them our zoom tool at their highest thinking level.

\textbf{Learning from contrasts.}
Contrast sets and similar minimal pairs~\cite{goyal2017making,gardner2020evaluating,thrush2022winoground,li2024naturalbench} test whether answers follow the image, and they credit a model only if it answers every member correctly.
Counterfactual examples also serve as supervised training data~\cite{kaushik2020learning,teney2020learning,chen2020counterfactual}.
For RL, a reward on the final answer cannot distinguish an answer that is read from the image from an answer that is based on priors~\cite{ye2026samereward,liu2026evan,yao2026mmrewardhacking}.
DeFacto~\cite{xu2026defacto} and EVAN~\cite{liu2026evan} therefore add training signals from inputs without the evidence, whereas our versions replace the evidence, so every version has an answer to read.
GRPO~\cite{shao2024deepseekmath,guo2025deepseekr1} and its variants~\cite{ahmadian2024back,liu2025understanding,yu2025dapo,zhou2026demystifying} compare rollouts of one prompt, also for vision-language models~\cite{huang2026visionr1,shen2025vlmr1,liu2025visualrft,wang2025thinklite}.
Recent methods train on augmented or rephrased copies of an input with the same answer~\cite{liu2025noisyrollout,zou2026consistency,zhu2026comet,yao2025sharevl}, on counterfactual pairs~\cite{du2026crpo,li2026cfcamo,huang2025taming} or on packs of inputs~\cite{kang2026bpo,jiang2026cargo}.
CRPO~\cite{du2026crpo} weights an input and its transformed copy by the accuracy on the other, and CrossRPO extends this coupling to any number of inputs (Appendix~\ref{sec:theory-related}).

\section{Cross-Input Relative Policy Optimization}
\label{sec:method}

On a single image, a correct guess earns the same reward as reading, so RL on final answers cannot distinguish them.
A contrast set can distinguish them, because a guess is correct on only one of its versions.
CrossRPO turns this difference into a training signal for visual agents.
Figure~\ref{fig:method} compares its pipeline with GRPO. The theorems, propositions and their proofs are in Appendix~\ref{sec:theory}.

\subsection{Preliminaries}
\label{sec:method-setup}
\textbf{Reward.}
The agent receives a question $q$ and an image $s$ downsampled to a pixel budget of $x$\,MP.
At each turn, it either answers or calls a zoom tool with a box on the image.
The tool crops that region from the full-resolution image and resizes it to fit the same budget (Appendix~\ref{sec:supp-tool}).
A rollout $o\sim P_\theta(\cdot\mid s)$ is the full sequence of turns on the input $s$ under the policy $\pi_\theta$.
For the correct answer $y$, its correctness is $c(o)=\mathbf 1[\mathrm{ans}(o)=y]$, and its~reward~is~$r(o)=c(o)$.

\textbf{Group-based RL.}
CrossRPO builds on RL methods that compare several rollouts of the same input.
These methods sample $o_1,\dots,o_n\sim P_\theta(\cdot\mid s)$ and update $\theta$ along $\hat g=\sum_iA_i\nabla_\theta\log P_\theta(o_i\mid s)$, where the advantage $A_i$ compares $c_i=c(o_i)$ with the other rollouts.
GRPO~\cite{shao2024deepseekmath} uses $A_i=(c_i-\mu)/\sigma$ with $\mu=\frac1n\sum_jc_j$ and $\sigma^2=\frac1n\sum_j(c_j-\mu)^2$, and $A_i=0$ if all $c_j$ are equal.
RLOO~\cite{ahmadian2024back} uses $A_i=c_i-\bar c_{-i}$ with $\bar c_{-i}=\frac1{n-1}\sum_{j\neq i}c_j$.
Both depend only on the rollouts of one input, so a correct guess and a correct reading receive the same advantage.

\subsection{Contrast sets and the objective}
\label{sec:method-sets}
A contrast set~\cite{gardner2020evaluating} is a tuple $\mathcal S=(q,(s_v,y_v)_{v=1}^K)$ of a question $q$ and $K$ versions $s_v$ of an image with correct answers $y_v$.
The versions are identical except in a small region with the evidence, and $y_u\neq y_v$ for $u\neq v$.
A single image is a set with $K=1$.
The agent sees one version at a time.
Let $o_v\sim P_\theta(\cdot\mid s_v)$ be independent rollouts, one on each version, with $c_v=\mathbf 1[\mathrm{ans}(o_v)=y_v]$.
The agent \emph{solves} the set if $c_1=\dots=c_K=1$.
The accuracy on version $v$ and the \emph{set accuracy} are
\begin{equation}
  p_v=\Pr[c_v=1],\quad\Pr[c_1=\dots=c_K=1]=\prod_vp_v.
  \label{eq:method-setacc}
\end{equation}
Consider an agent that does not read the region that differs.
Since the versions are identical outside this region, its answer has the same distribution on every version, $\rho(a)=\Pr[\mathrm{ans}(o_v)=a]$ for all $v$, and hence $p_v=\rho(y_v)$.
Since the $y_v$ are pairwise different, $\sum_vp_v=\sum_v\rho(y_v)\le\sum_a\rho(a)\le1$, and hence $\prod_vp_v\le\big(\frac1K\sum_vp_v\big)^K\le K^{-K}$ (Proposition~\ref{prop:blind}).
If the agent always gives the same answer and $K\ge2$, then $\prod_vp_v=0$.
In contrast, an agent that reads can reach $p_v=1$ for all $v$ and $\prod_vp_v=1$.
Set accuracy alone gives a weak signal early in training.
The term of version $v$ in its gradient is $\prod_{u\neq v}p_u\nabla_\theta p_v$, which is zero when another version is never answered correctly.
CrossRPO therefore maximizes the accuracy together with the set accuracy.
For $K\le3$, which covers all our image sets, the~objective~is
\begin{equation}
  J(\theta)=\sum_{v=1}^Kp_v+\lambda\prod_{v=1}^Kp_v,\qquad\lambda\ge0.
  \label{eq:method-objective}
\end{equation}
We call $\lambda\prod_vp_v$ the \emph{joint term}, which rewards the agent for answering all versions correctly.

\subsection{The CrossRPO advantage}
\label{sec:method-crossrpo}
\begin{figure*}[t]
  \centering
  \includegraphics[width=\textwidth]{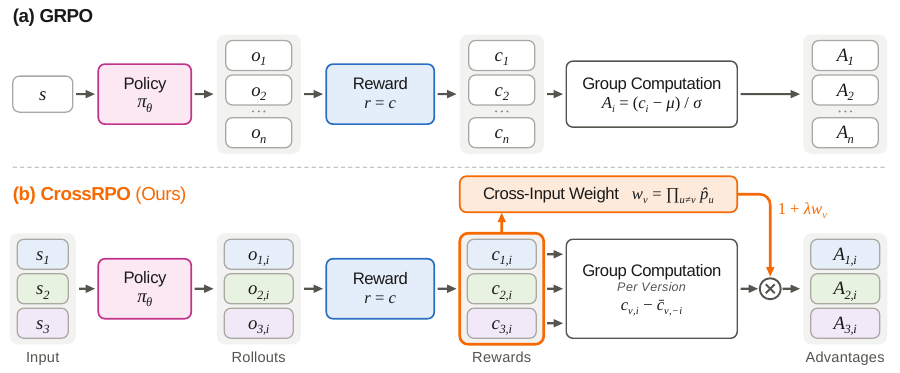}
  \caption{\textbf{GRPO and CrossRPO.}
  The policy $\pi_\theta$ samples the rollouts, and the reward $r$ is the correctness $c$ of the answer.
  (a)~GRPO compares the rollouts of one input $s$.
  (b)~CrossRPO compares rollouts within each version $s_v$ of a set and multiplies the result by $1+\lambda w_v$, where $w_v$ comes from the rewards of~the~other~versions.}
  \label{fig:method}
\end{figure*}

We sample $n$ rollouts $o_{v,1},\dots,o_{v,n}\sim P_\theta(\cdot\mid s_v)$ on each version and write $c_{v,i}=c(o_{v,i})$, $\hat p_v=\frac1n\sum_ic_{v,i}$ and $\bar c_{v,-i}=\frac1{n-1}\sum_{j\neq i}c_{v,j}$.
GRPO can be applied to a set in two ways.
With one group per version, $A_{v,i}=(c_{v,i}-\mu_v)/\sigma_v$, where $\mu_v$ and $\sigma_v$ are the mean and standard deviation of the rollouts of version $v$.
This advantage does not depend on the other versions, so the update contains no joint term.
With one group for all $Kn$ rollouts (pooled GRPO), $A_{v,i}=(c_{v,i}-\mu)/\sigma$ with the mean $\mu$ and standard deviation $\sigma$ of the whole set.
A guess that is correct on one version then has $c_{v,i}=1>\mu$ and receives a positive advantage.
CrossRPO keeps one group per version and adds the joint term through a per-version weight (Figure~\ref{fig:method}, bottom).

By the product rule, the gradient of Eq.~(\ref{eq:method-objective}) is
\begin{equation}
  \nabla_\theta J=\sum_{v=1}^K\Big(1+\lambda\prod_{u\neq v}p_u\Big)\nabla_\theta p_v.
  \label{eq:method-grad}
\end{equation}
The factor $\prod_{u\neq v}p_u$ is the probability that all other versions are answered correctly.
We estimate this probability by the \emph{cross-input weight}
\begin{equation}
  w_v=\prod_{u\neq v}\hat p_u,
  \label{eq:method-weight}
\end{equation}
which satisfies $\E[w_v]=\prod_{u\neq v}p_u$ and is independent of the rollouts of version $v$.
We estimate $\nabla_\theta p_v$ as in RLOO.
This gives the \emph{CrossRPO advantage} and the gradient estimate $\hat g$ of a set,
\begin{align}
  A_{v,i}&=(1+\lambda w_v)\,\big(c_{v,i}-\bar c_{v,-i}\big),\label{eq:method-advantage}\\
  \hat g&=\sum_{v=1}^K\sum_{i=1}^nA_{v,i}\,\nabla_\theta\log P_\theta(o_{v,i}\mid s_v),\label{eq:method-ghat}
\end{align}
with $\E[\hat g]=n\,\nabla_\theta J$ (Theorem~\ref{thm:unbiased}).
Eq.~(\ref{eq:advantage}) in Appendix~\ref{sec:theory-advantage} generalizes Eq.~(\ref{eq:method-advantage}) to any reward.
With $\lambda=0$, Eq.~(\ref{eq:method-advantage}) is RLOO on each version separately.
For $K=1$, $\prod_vp_v=p_1$ repeats the first term of Eq.~(\ref{eq:method-objective}), so we set $w_1=0$, and $A_{1,i}=c_{1,i}-\bar c_{1,-i}$ is RLOO.

Eq.~(\ref{eq:method-advantage}) has three properties.
(a)~$A_{v,i}=0$ for all $i$ if and only if $c_{v,1}=\dots=c_{v,n}$.
(b)~Otherwise, $A_{v,i}>0$ if $c_{v,i}=1$ and $A_{v,i}<0$ if $c_{v,i}=0$.
(c)~$1\le1+\lambda w_v\le1+\lambda$, with $w_v=0$ if $\hat p_u=0$ for some $u\neq v$ and $w_v=1$ if $\hat p_u=1$ for all $u\neq v$.
A version therefore has nonzero advantages whenever its own rollouts differ, and they are largest when all other versions are answered correctly.
In Figure~\ref{fig:teaser}a ($K=3$), an agent that always answers ``smartphone'' has $c_{1,i}=1$ and $c_{2,i}=c_{3,i}=0$ for all $i$, so $A_{v,i}=0$ by (a), whereas pooled GRPO gives $\sqrt2\approx1.41$ to every guess on the original.
If one rollout on the second version is correct, it receives $A_{2,i}=1$.
Once $\hat p_1=\hat p_2=1$, a correct rollout on the third version receives $(1+\lambda)(1-\bar c_{3,-i})$, the largest advantage in the set.
In Eq.~(\ref{eq:method-advantage}), $c_{v,i}-\bar c_{v,-i}$ compares a rollout only with rollouts of the same version, and the factor $1+\lambda w_v$ grows when the other versions are correct, so the answer that solves the set receives the most credit.

\textbf{Batch scale.}
Each training step uses a batch $\mathcal B$ of sets, with rollouts on every version of every set.
Before the update, the advantages of these rollouts are divided by a scale.
GRPO uses one scale per group: $A_i=(c_i-\mu)/\sigma$ divides by the standard deviation $\sigma$ of the group's own rollouts.
In CrossRPO, a group is one version, and all its rollouts share the factor $1+\lambda w_v$.
A scale per group would cancel this factor.
CrossRPO therefore uses one scale for the whole batch.
Let $A_1,\dots,A_M$ be the advantages of all $M$ rollouts in $\mathcal B$.
We divide each of them by their root mean square,
\begin{equation}
  \tilde A_m=\frac{A_m}{s_{\mathcal B}},\qquad s_{\mathcal B}=\Big(\frac1M\sum_{m=1}^MA_m^2\Big)^{1/2}.
  \label{eq:method-scale}
\end{equation}
All advantages are divided by the same number, so their ratios are kept, and the scale changes the length of the step but not its direction (Proposition~\ref{prop:scale}).

\textbf{Scaling to larger sets.}
Computing Eq.~(\ref{eq:method-advantage}) takes $O(nK)$ operations per set (Appendix~\ref{sec:supp-loop}).
If $p_u=p$ for all $u$, the weight in Eq.~(\ref{eq:method-grad}) is $\prod_{u\neq v}p_u=p^{K-1}$, which vanishes as $K$ grows (Proposition~\ref{prop:order}).
For $K>3$, CrossRPO therefore uses the accuracy on triples of versions as the joint term, $\lambda\binom{K-1}2^{-1}\sum_{t<u<v}p_tp_up_v$, which equals $\lambda\prod_vp_v$ for $K=3$ (Eq.~(\ref{eq:objective})).
The weight $w_v$ becomes the mean of $\hat p_u\hat p_{u'}$ over the $\binom{K-1}2$ pairs of other versions, which is of order $p^2$ for every $K$.
Appendix~\ref{sec:theory-general} treats sets of any size, and Section~\ref{sec:exp-general} trains CrossRPO on text sets with larger $K=4$, 8 and 16.

\subsection{Guarantees}
\label{sec:method-props}
Appendix~\ref{sec:theory} proves five properties of this update.
(i)~\emph{Unbiasedness.}
$\E[\hat g]=n\,\nabla_\theta J$, and the batch scale changes only the step length (Theorem~\ref{thm:unbiased}, Remark~\ref{rem:unbiased}).
(ii)~\emph{Variance.}
$w_v$ averages over all $n^{K-1}$ combinations of rollouts on the other versions.
For every fixed pairing of rollouts across versions, such as pairing by index, the estimate $\hat g_{\mathrm{pair}}$ has the same mean and $\Cov(\hat g_{\mathrm{pair}})\succeq\Cov(\hat g)$ (Theorem~\ref{thm:rb}).
(iii)~\emph{No credit for a guess.}
If $c_{v^\star,i}=1$ and $c_{v,i}=0$ for all $v\neq v^\star$ and all $i$, then $A_{v,i}=0$ for all $v$ and $i$, whereas pooled GRPO gives $\sqrt{K-1}$ to every rollout of $v^\star$ (Proposition~\ref{prop:prior}).
(iv)~\emph{Special cases.}
Eq.~(\ref{eq:method-objective}) contains the objectives of GRPO and RLOO ($K=1$) and of recent two-input methods ($K=2$) (Proposition~\ref{prop:related}).
For $K\ge3$, $\prod_vc_v$ cannot be written as $\sum_v\phi_v(c_v)+\sum_{u<v}\phi_{uv}(c_u,c_v)$, so rewards on pairs of versions cannot express the joint term (Proposition~\ref{prop:not-pairwise}).
(v)~\emph{Which inputs to group.}
If the $K\le3$ inputs of a set are grouped at random with mean accuracy $\bar p$, then $\nabla_\theta\E[J]=K(1+\lambda\bar p^{K-1})\nabla_\theta\bar p$, a positive multiple of the gradient of mean accuracy.
The joint term thus helps only if the members share a shortcut (Propositions~\ref{prop:independent} and~\ref{prop:shortcut}).

\section{ViRA-Bench and ViRA-Train}
\label{sec:data}

CrossRPO requires sets whose versions differ only in the evidence and have different correct answers.
This section introduces three datasets of such sets, ViRA-Bench, ViRA-Bench-Real and ViRA-Train, which are built from ultra-high-resolution photos of 21\,MP to 120\,MP.

\textbf{Building a contrast set.}
The original photos are color photos from Wikimedia Commons.
We use a vision-language model to find an object in a photo that can be changed into $K-1$ other versions.
For $K=3$, a smartphone in a hand (the original $s_1$) can be changed into a calculator (variant $s_2$) and a TV remote (variant $s_3$), as in Figure~\ref{fig:teaser}a.
An image editor redraws the object for each variant, and only the region of the old and new object is pasted back into the photo.
A language model writes a question $q$ about the object with four options, including the correct answer $y_v$ of every version and at least one distractor.
The question first states the location of the object and never names the changed attribute.
Some variants contradict common sense, so a model that guesses gives the original's answer (Figure~\ref{fig:bench}).
The same versions can serve several sets: another question $q'$ with answers $y'_v$ gives the set $(q',(s_v,y'_v)_{v=1}^K)$.
After automatic checks, human reviewers check every ViRA-Bench question (Appendix~\ref{sec:supp-bench}).

\begin{wrapfigure}{R}{0.50\textwidth}
  \centering
  \includegraphics[width=.95\linewidth]{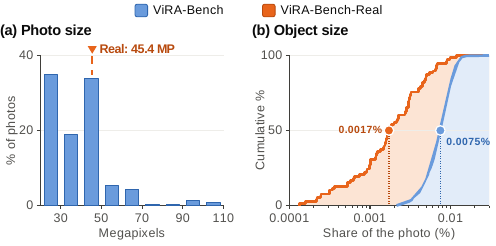}
  \caption{\textbf{Statistics of ViRA-Bench and ViRA-Bench-Real.}
  (a)~Size of the 466 ViRA-Bench photos, and every ViRA-Bench-Real photo has 45.4\,MP.
  (b)~Cumulative distribution of the object's area as a share of the photo, with dots at the medians.}
  \label{fig:data-stats}
\end{wrapfigure}
\textbf{ViRA-Bench} contains 762 contrast sets on 466 photos of 24 to 108\,MP (Figure~\ref{fig:bench}).
Each set has $K=3$ versions, the original O and the two variants A and B.
An object is kept only if its longer side is at most 13 pixels when the photo is reduced to 1\,MP and its shorter side is at least 28 pixels at full resolution, so a model must zoom in to read it.
The edited object covers a median of 0.0075\% of the photo (Figure~\ref{fig:data-stats}b).
Of the questions, 458 ask for a visible property, 209 add a step of reasoning, and 95 involve several objects.
The metric is set accuracy, the share of sets with all three versions answered correctly.
The first zoom of every agent covers the object equally often on the edited variants and on the original (Appendix~\ref{sec:supp-traces}): the two shares differ by at most 2.4 percentage \mbox{points at every budget}.

\textbf{ViRA-Bench-Real.}
To exclude editing traces, ViRA-Bench-Real contains 78 sets that were photographed without any editing (Figure~\ref{fig:bench}, right; Appendix~\ref{sec:supp-real}).
Its 234 photos were taken with a Nikon Z8 camera.
Between the three shots of a set, only the object was swapped.
The swapped objects are posters, signs, stickers, labels and small everyday objects.
The photos have 45.4\,MP, and the objects cover a median of 0.0017\% of the photo.

\textbf{ViRA-Train} contains 6,457 contrast sets on 3,797 Commons photos of 21 to 120\,MP that are not in ViRA-Bench (Appendix~\ref{sec:supp-train}).
Of these sets, 4,110 have three versions and 2,347 have two.
The edited region has a shorter side of at least 20 pixels at full resolution, and its longer side has a median of 16 pixels at 1\,MP.
ViRA-Train also contains 2,804 single images as sets with $K=1$.
They are questions about details in Commons photos, questions about several objects in photos and maps, math problems from ThinkLite-VL~\cite{wang2025thinklite} and synthetic charts.

\begin{table*}[t]
\caption{\textbf{Set accuracy (\%)} at five pixel budgets, with the tokens per set in gray.
\emph{Max}: each model's maximum resolution (\emph{Max Res.}), which also caps the other budgets.
$\Delta$: gain of ViRA over its base model with a zoom tool.
Green and red cells: ViRA-27B is more or less accurate than the proprietary model.
Bold: best in each group.
ViRA results are means of three runs. Tables~\ref{tab:image-acc} and~\ref{tab:ablation-bench} give the per-image accuracy and the ablations, and Appendix~\ref{sec:supp-eval} gives the protocol.}
\label{tab:main}\label{tab:full}\label{tab:real-full}
\centering
\scriptsize
\setlength{\tabcolsep}{2.2pt}
\settowidth{\accw}{\textbf{00.0}}\settowidth{\tokw}{\tok{000K}}\settowidth{\numw}{\textbf{00.0}}
\resizebox{\textwidth}{!}{%
\begin{tabular}{l*{11}{c}}
\toprule
\midrule
 & Max Res. & \multicolumn{5}{c}{ViRA-Bench} & \multicolumn{5}{c}{ViRA-Bench-Real} \\
\cmidrule(lr){3-7}\cmidrule(l){8-12}
Model & (MP) & 0.5\,MP & 1\,MP & 2\,MP & 4\,MP & Max & 0.5\,MP & 1\,MP & 2\,MP & 4\,MP & Max \\
\midrule
\multicolumn{12}{l}{\emph{Open 7--9B models}} \\
Qwen3.5-9B & \nb{16.8} & \makebox[\accw][r]{0.9}\makebox[\tokw][l]{\tok{2K}} & \makebox[\accw][r]{2.2}\makebox[\tokw][l]{\tok{4K}} & \makebox[\accw][r]{3.8}\makebox[\tokw][l]{\tok{6K}} & \makebox[\accw][r]{7.0}\makebox[\tokw][l]{\tok{12K}} & \makebox[\accw][r]{12.2}\makebox[\tokw][l]{\tok{49K}} & \makebox[\accw][r]{0.0}\makebox[\tokw][l]{\tok{2K}} & \makebox[\accw][r]{0.0}\makebox[\tokw][l]{\tok{3K}} & \makebox[\accw][r]{0.0}\makebox[\tokw][l]{\tok{6K}} & \makebox[\accw][r]{2.6}\makebox[\tokw][l]{\tok{12K}} & \makebox[\accw][r]{7.7}\makebox[\tokw][l]{\tok{49K}} \\
ZwZ2-9B & \nb{16.8} & \makebox[\accw][r]{2.1}\makebox[\tokw][l]{\tok{4K}} & \makebox[\accw][r]{3.8}\makebox[\tokw][l]{\tok{5K}} & \makebox[\accw][r]{9.2}\makebox[\tokw][l]{\tok{8K}} & \makebox[\accw][r]{13.9}\makebox[\tokw][l]{\tok{14K}} & \makebox[\accw][r]{24.1}\makebox[\tokw][l]{\tok{51K}} & \makebox[\accw][r]{0.0}\makebox[\tokw][l]{\tok{3K}} & \makebox[\accw][r]{0.0}\makebox[\tokw][l]{\tok{5K}} & \makebox[\accw][r]{0.0}\makebox[\tokw][l]{\tok{8K}} & \makebox[\accw][r]{2.6}\makebox[\tokw][l]{\tok{14K}} & \makebox[\accw][r]{17.9}\makebox[\tokw][l]{\tok{50K}} \\
Vision-OPD-9B & \nb{16.8} & \makebox[\accw][r]{2.1}\makebox[\tokw][l]{\tok{2K}} & \makebox[\accw][r]{5.1}\makebox[\tokw][l]{\tok{3K}} & \makebox[\accw][r]{8.0}\makebox[\tokw][l]{\tok{6K}} & \makebox[\accw][r]{12.3}\makebox[\tokw][l]{\tok{12K}} & \makebox[\accw][r]{20.5}\makebox[\tokw][l]{\tok{49K}} & \makebox[\accw][r]{0.0}\makebox[\tokw][l]{\tok{2K}} & \makebox[\accw][r]{0.0}\makebox[\tokw][l]{\tok{3K}} & \makebox[\accw][r]{1.3}\makebox[\tokw][l]{\tok{6K}} & \makebox[\accw][r]{2.6}\makebox[\tokw][l]{\tok{12K}} & \makebox[\accw][r]{24.4}\makebox[\tokw][l]{\tok{49K}} \\
\midrule[0.3pt]
\multicolumn{12}{l}{\emph{Open 7--9B models + tool}} \\
Qwen3.5-9B + tool & \nb{16.8} & \makebox[\accw][r]{21.1}\makebox[\tokw][l]{\tok{16K}} & \makebox[\accw][r]{26.2}\makebox[\tokw][l]{\tok{22K}} & \makebox[\accw][r]{31.4}\makebox[\tokw][l]{\tok{31K}} & \makebox[\accw][r]{34.4}\makebox[\tokw][l]{\tok{45K}} & \makebox[\accw][r]{34.6}\makebox[\tokw][l]{\tok{136K}} & \makebox[\accw][r]{10.3}\makebox[\tokw][l]{\tok{22K}} & \makebox[\accw][r]{16.7}\makebox[\tokw][l]{\tok{30K}} & \makebox[\accw][r]{20.5}\makebox[\tokw][l]{\tok{36K}} & \makebox[\accw][r]{23.1}\makebox[\tokw][l]{\tok{55K}} & \makebox[\accw][r]{34.6}\makebox[\tokw][l]{\tok{141K}} \\
ZwZ2-9B + tool & \nb{16.8} & \makebox[\accw][r]{25.3}\makebox[\tokw][l]{\tok{15K}} & \makebox[\accw][r]{34.1}\makebox[\tokw][l]{\tok{18K}} & \makebox[\accw][r]{36.0}\makebox[\tokw][l]{\tok{26K}} & \makebox[\accw][r]{40.4}\makebox[\tokw][l]{\tok{40K}} & \makebox[\accw][r]{44.2}\makebox[\tokw][l]{\tok{128K}} & \makebox[\accw][r]{25.6}\makebox[\tokw][l]{\tok{17K}} & \makebox[\accw][r]{21.8}\makebox[\tokw][l]{\tok{23K}} & \makebox[\accw][r]{37.2}\makebox[\tokw][l]{\tok{30K}} & \makebox[\accw][r]{35.9}\makebox[\tokw][l]{\tok{46K}} & \makebox[\accw][r]{46.2}\makebox[\tokw][l]{\tok{131K}} \\
Vision-OPD-9B + tool & \nb{16.8} & \makebox[\accw][r]{2.5}\makebox[\tokw][l]{\tok{4K}} & \makebox[\accw][r]{5.9}\makebox[\tokw][l]{\tok{6K}} & \makebox[\accw][r]{8.5}\makebox[\tokw][l]{\tok{8K}} & \makebox[\accw][r]{12.1}\makebox[\tokw][l]{\tok{15K}} & \makebox[\accw][r]{20.1}\makebox[\tokw][l]{\tok{52K}} & \makebox[\accw][r]{2.6}\makebox[\tokw][l]{\tok{7K}} & \makebox[\accw][r]{5.1}\makebox[\tokw][l]{\tok{9K}} & \makebox[\accw][r]{3.8}\makebox[\tokw][l]{\tok{12K}} & \makebox[\accw][r]{5.1}\makebox[\tokw][l]{\tok{21K}} & \makebox[\accw][r]{24.4}\makebox[\tokw][l]{\tok{65K}} \\
\midrule[0.3pt]
\multicolumn{12}{l}{\emph{Released zoom agents (7--9B)}} \\
Mini-o3 & \nb{12.8} & \makebox[\accw][r]{17.2}\makebox[\tokw][l]{\tok{41K}} & \makebox[\accw][r]{18.9}\makebox[\tokw][l]{\tok{59K}} & \makebox[\accw][r]{17.3}\makebox[\tokw][l]{\tok{102K}} & \makebox[\accw][r]{21.1}\makebox[\tokw][l]{\tok{165K}} & \makebox[\accw][r]{18.8}\makebox[\tokw][l]{\tok{396K}} & \makebox[\accw][r]{16.7}\makebox[\tokw][l]{\tok{200K}} & \makebox[\accw][r]{26.9}\makebox[\tokw][l]{\tok{158K}} & \makebox[\accw][r]{25.6}\makebox[\tokw][l]{\tok{298K}} & \makebox[\accw][r]{23.1}\makebox[\tokw][l]{\tok{287K}} & \makebox[\accw][r]{26.9}\makebox[\tokw][l]{\tok{781K}} \\
DeepEyes & \nb{12.8} & \makebox[\accw][r]{0.8}\makebox[\tokw][l]{\tok{9K}} & \makebox[\accw][r]{2.9}\makebox[\tokw][l]{\tok{13K}} & \makebox[\accw][r]{4.3}\makebox[\tokw][l]{\tok{21K}} & \makebox[\accw][r]{6.6}\makebox[\tokw][l]{\tok{37K}} & \makebox[\accw][r]{9.7}\makebox[\tokw][l]{\tok{106K}} & \makebox[\accw][r]{2.6}\makebox[\tokw][l]{\tok{9K}} & \makebox[\accw][r]{5.1}\makebox[\tokw][l]{\tok{13K}} & \makebox[\accw][r]{2.6}\makebox[\tokw][l]{\tok{22K}} & \makebox[\accw][r]{2.6}\makebox[\tokw][l]{\tok{38K}} & \makebox[\accw][r]{12.8}\makebox[\tokw][l]{\tok{106K}} \\
DeepEyesV2 & \nb{12.8} & \makebox[\accw][r]{0.3}\makebox[\tokw][l]{\tok{24K}} & \makebox[\accw][r]{1.3}\makebox[\tokw][l]{\tok{27K}} & \makebox[\accw][r]{2.2}\makebox[\tokw][l]{\tok{37K}} & \makebox[\accw][r]{5.2}\makebox[\tokw][l]{\tok{55K}} & \makebox[\accw][r]{9.6}\makebox[\tokw][l]{\tok{137K}} & \makebox[\accw][r]{1.3}\makebox[\tokw][l]{\tok{26K}} & \makebox[\accw][r]{0.0}\makebox[\tokw][l]{\tok{36K}} & \makebox[\accw][r]{1.3}\makebox[\tokw][l]{\tok{50K}} & \makebox[\accw][r]{2.6}\makebox[\tokw][l]{\tok{72K}} & \makebox[\accw][r]{7.7}\makebox[\tokw][l]{\tok{180K}} \\
Thyme & \nb{12.8} & \makebox[\accw][r]{0.8}\makebox[\tokw][l]{\tok{5K}} & \makebox[\accw][r]{2.9}\makebox[\tokw][l]{\tok{8K}} & \makebox[\accw][r]{3.7}\makebox[\tokw][l]{\tok{13K}} & \makebox[\accw][r]{6.0}\makebox[\tokw][l]{\tok{22K}} & \makebox[\accw][r]{9.8}\makebox[\tokw][l]{\tok{59K}} & \makebox[\accw][r]{1.3}\makebox[\tokw][l]{\tok{7K}} & \makebox[\accw][r]{5.1}\makebox[\tokw][l]{\tok{9K}} & \makebox[\accw][r]{1.3}\makebox[\tokw][l]{\tok{15K}} & \makebox[\accw][r]{2.6}\makebox[\tokw][l]{\tok{26K}} & \makebox[\accw][r]{11.5}\makebox[\tokw][l]{\tok{67K}} \\
Pixel Reasoner & \nb{12.8} & \makebox[\accw][r]{0.5}\makebox[\tokw][l]{\tok{7K}} & \makebox[\accw][r]{3.1}\makebox[\tokw][l]{\tok{10K}} & \makebox[\accw][r]{2.8}\makebox[\tokw][l]{\tok{16K}} & \makebox[\accw][r]{6.6}\makebox[\tokw][l]{\tok{30K}} & \makebox[\accw][r]{9.3}\makebox[\tokw][l]{\tok{74K}} & \makebox[\accw][r]{0.0}\makebox[\tokw][l]{\tok{6K}} & \makebox[\accw][r]{1.3}\makebox[\tokw][l]{\tok{9K}} & \makebox[\accw][r]{0.0}\makebox[\tokw][l]{\tok{12K}} & \makebox[\accw][r]{1.3}\makebox[\tokw][l]{\tok{25K}} & \makebox[\accw][r]{12.8}\makebox[\tokw][l]{\tok{61K}} \\
ZoomEye & \nb{12.8} & \makebox[\accw][r]{12.6}\makebox[\tokw][l]{\tok{303K}} & \makebox[\accw][r]{14.6}\makebox[\tokw][l]{\tok{393K}} & \makebox[\accw][r]{16.4}\makebox[\tokw][l]{\tok{516K}} & \makebox[\accw][r]{17.1}\makebox[\tokw][l]{\tok{696K}} & \makebox[\accw][r]{20.5}\makebox[\tokw][l]{\tok{1.1M}} & \makebox[\accw][r]{17.9}\makebox[\tokw][l]{\tok{314K}} & \makebox[\accw][r]{17.9}\makebox[\tokw][l]{\tok{398K}} & \makebox[\accw][r]{16.7}\makebox[\tokw][l]{\tok{529K}} & \makebox[\accw][r]{20.5}\makebox[\tokw][l]{\tok{754K}} & \makebox[\accw][r]{28.2}\makebox[\tokw][l]{\tok{1.2M}} \\
\midrule[0.3pt]
\multicolumn{12}{l}{\emph{Ours + our zoom tool}} \\
\shaderow{vira9bg}\textbf{ViRA-9B} (ours) & \nb{16.8} & \makebox[\accw][r]{\textbf{38.7}}\makebox[\tokw][l]{\tok{12K}} & \makebox[\accw][r]{\textbf{46.9}}\makebox[\tokw][l]{\tok{16K}} & \makebox[\accw][r]{\textbf{55.6}}\makebox[\tokw][l]{\tok{26K}} & \makebox[\accw][r]{\textbf{57.5}}\makebox[\tokw][l]{\tok{44K}} & \makebox[\accw][r]{\textbf{63.3}}\makebox[\tokw][l]{\tok{147K}} & \makebox[\accw][r]{\textbf{31.2}}\makebox[\tokw][l]{\tok{11K}} & \makebox[\accw][r]{\textbf{49.6}}\makebox[\tokw][l]{\tok{15K}} & \makebox[\accw][r]{\textbf{57.7}}\makebox[\tokw][l]{\tok{23K}} & \makebox[\accw][r]{\textbf{68.8}}\makebox[\tokw][l]{\tok{41K}} & \makebox[\accw][r]{\textbf{77.4}}\makebox[\tokw][l]{\tok{135K}} \\
\quad $\Delta$ vs.\ Qwen3.5-9B + tool &  & \dgain{+17.6}\makebox[\tokw]{} & \dgain{+20.7}\makebox[\tokw]{} & \dgain{+24.2}\makebox[\tokw]{} & \dgain{+23.1}\makebox[\tokw]{} & \dgain{+28.7}\makebox[\tokw]{} & \dgain{+20.9}\makebox[\tokw]{} & \dgain{+32.9}\makebox[\tokw]{} & \dgain{+37.2}\makebox[\tokw]{} & \dgain{+45.7}\makebox[\tokw]{} & \dgain{+42.8}\makebox[\tokw]{} \\
\midrule[1pt]
\multicolumn{12}{l}{\emph{Open 27B models}} \\
Qwen3.8-27B & \nb{16.8} & \makebox[\accw][r]{1.4}\makebox[\tokw][l]{\tok{2K}} & \makebox[\accw][r]{3.7}\makebox[\tokw][l]{\tok{4K}} & \makebox[\accw][r]{7.5}\makebox[\tokw][l]{\tok{7K}} & \makebox[\accw][r]{13.3}\makebox[\tokw][l]{\tok{13K}} & \makebox[\accw][r]{22.3}\makebox[\tokw][l]{\tok{50K}} & \makebox[\accw][r]{0.0}\makebox[\tokw][l]{\tok{2K}} & \makebox[\accw][r]{0.0}\makebox[\tokw][l]{\tok{3K}} & \makebox[\accw][r]{0.0}\makebox[\tokw][l]{\tok{7K}} & \makebox[\accw][r]{2.6}\makebox[\tokw][l]{\tok{12K}} & \makebox[\accw][r]{15.4}\makebox[\tokw][l]{\tok{49K}} \\
ZwZ2-27B & \nb{16.8} & \makebox[\accw][r]{2.4}\makebox[\tokw][l]{\tok{3K}} & \makebox[\accw][r]{5.6}\makebox[\tokw][l]{\tok{4K}} & \makebox[\accw][r]{11.5}\makebox[\tokw][l]{\tok{7K}} & \makebox[\accw][r]{15.1}\makebox[\tokw][l]{\tok{13K}} & \makebox[\accw][r]{28.3}\makebox[\tokw][l]{\tok{50K}} & \makebox[\accw][r]{0.0}\makebox[\tokw][l]{\tok{3K}} & \makebox[\accw][r]{0.0}\makebox[\tokw][l]{\tok{4K}} & \makebox[\accw][r]{1.3}\makebox[\tokw][l]{\tok{7K}} & \makebox[\accw][r]{5.1}\makebox[\tokw][l]{\tok{13K}} & \makebox[\accw][r]{19.2}\makebox[\tokw][l]{\tok{50K}} \\
\midrule[0.3pt]
\multicolumn{12}{l}{\emph{Open 27B models + tool}} \\
Qwen3.8-27B + tool & \nb{16.8} & \makebox[\accw][r]{26.5}\makebox[\tokw][l]{\tok{27K}} & \makebox[\accw][r]{33.3}\makebox[\tokw][l]{\tok{32K}} & \makebox[\accw][r]{41.1}\makebox[\tokw][l]{\tok{42K}} & \makebox[\accw][r]{44.0}\makebox[\tokw][l]{\tok{59K}} & \makebox[\accw][r]{48.6}\makebox[\tokw][l]{\tok{174K}} & \makebox[\accw][r]{20.5}\makebox[\tokw][l]{\tok{30K}} & \makebox[\accw][r]{30.8}\makebox[\tokw][l]{\tok{36K}} & \makebox[\accw][r]{33.3}\makebox[\tokw][l]{\tok{44K}} & \makebox[\accw][r]{44.9}\makebox[\tokw][l]{\tok{65K}} & \makebox[\accw][r]{59.0}\makebox[\tokw][l]{\tok{181K}} \\
ZwZ2-27B + tool & \nb{16.8} & \makebox[\accw][r]{24.5}\makebox[\tokw][l]{\tok{35K}} & \makebox[\accw][r]{32.8}\makebox[\tokw][l]{\tok{45K}} & \makebox[\accw][r]{35.7}\makebox[\tokw][l]{\tok{59K}} & \makebox[\accw][r]{40.3}\makebox[\tokw][l]{\tok{81K}} & \makebox[\accw][r]{44.5}\makebox[\tokw][l]{\tok{219K}} & \makebox[\accw][r]{26.9}\makebox[\tokw][l]{\tok{40K}} & \makebox[\accw][r]{28.2}\makebox[\tokw][l]{\tok{51K}} & \makebox[\accw][r]{38.5}\makebox[\tokw][l]{\tok{67K}} & \makebox[\accw][r]{47.4}\makebox[\tokw][l]{\tok{83K}} & \makebox[\accw][r]{46.2}\makebox[\tokw][l]{\tok{226K}} \\
\midrule[0.3pt]
\multicolumn{12}{l}{\emph{Released zoom agents (27--32B)}} \\
LBS-27B & \nb{16.8} & \makebox[\accw][r]{42.8}\makebox[\tokw][l]{\tok{28K}} & \makebox[\accw][r]{51.2}\makebox[\tokw][l]{\tok{34K}} & \makebox[\accw][r]{57.3}\makebox[\tokw][l]{\tok{46K}} & \makebox[\accw][r]{63.4}\makebox[\tokw][l]{\tok{69K}} & \makebox[\accw][r]{68.6}\makebox[\tokw][l]{\tok{149K}} & \makebox[\accw][r]{43.6}\makebox[\tokw][l]{\tok{30K}} & \makebox[\accw][r]{48.7}\makebox[\tokw][l]{\tok{36K}} & \makebox[\accw][r]{64.1}\makebox[\tokw][l]{\tok{48K}} & \makebox[\accw][r]{69.2}\makebox[\tokw][l]{\tok{68K}} & \makebox[\accw][r]{71.8}\makebox[\tokw][l]{\tok{147K}} \\
SenseNova-MARS-32B & \nb{16.8} & \makebox[\accw][r]{26.1}\makebox[\tokw][l]{\tok{33K}} & \makebox[\accw][r]{32.9}\makebox[\tokw][l]{\tok{45K}} & \makebox[\accw][r]{36.1}\makebox[\tokw][l]{\tok{62K}} & \makebox[\accw][r]{38.5}\makebox[\tokw][l]{\tok{87K}} & \makebox[\accw][r]{40.8}\makebox[\tokw][l]{\tok{207K}} & \makebox[\accw][r]{25.6}\makebox[\tokw][l]{\tok{39K}} & \makebox[\accw][r]{30.8}\makebox[\tokw][l]{\tok{52K}} & \makebox[\accw][r]{43.6}\makebox[\tokw][l]{\tok{76K}} & \makebox[\accw][r]{37.2}\makebox[\tokw][l]{\tok{102K}} & \makebox[\accw][r]{48.7}\makebox[\tokw][l]{\tok{205K}} \\
\midrule[0.3pt]
\multicolumn{12}{l}{\emph{Ours + our zoom tool}} \\
\shaderow{vira27bg}\textbf{ViRA-27B} (ours) & \nb{16.8} & \makebox[\accw][r]{\textbf{64.0}}\makebox[\tokw][l]{\tok{50K}} & \makebox[\accw][r]{\textbf{68.4}}\makebox[\tokw][l]{\tok{60K}} & \makebox[\accw][r]{\textbf{71.2}}\makebox[\tokw][l]{\tok{81K}} & \makebox[\accw][r]{\textbf{72.9}}\makebox[\tokw][l]{\tok{128K}} & \makebox[\accw][r]{\textbf{71.5}}\makebox[\tokw][l]{\tok{409K}} & \makebox[\accw][r]{\textbf{72.6}}\makebox[\tokw][l]{\tok{41K}} & \makebox[\accw][r]{\textbf{75.2}}\makebox[\tokw][l]{\tok{50K}} & \makebox[\accw][r]{\textbf{79.1}}\makebox[\tokw][l]{\tok{70K}} & \makebox[\accw][r]{\textbf{80.8}}\makebox[\tokw][l]{\tok{109K}} & \makebox[\accw][r]{\textbf{76.5}}\makebox[\tokw][l]{\tok{354K}} \\
\quad $\Delta$ vs.\ Qwen3.8-27B + tool &  & \dgain{+37.5}\makebox[\tokw]{} & \dgain{+35.1}\makebox[\tokw]{} & \dgain{+30.1}\makebox[\tokw]{} & \dgain{+28.9}\makebox[\tokw]{} & \dgain{+22.9}\makebox[\tokw]{} & \dgain{+52.1}\makebox[\tokw]{} & \dgain{+44.4}\makebox[\tokw]{} & \dgain{+45.8}\makebox[\tokw]{} & \dgain{+35.9}\makebox[\tokw]{} & \dgain{+17.5}\makebox[\tokw]{} \\
\midrule[1pt]
\multicolumn{12}{l}{\emph{Proprietary models + our zoom tool}} \\
\cstrut Claude Haiku~5.5, max thinking & \nb{3.8} & \cellcolor{gainc!8}\makebox[\accw][r]{\textbf{58.8}}\makebox[\tokw][l]{\tok{70K}} & \cellcolor{gainc!10}\makebox[\accw][r]{\textbf{59.6}}\makebox[\tokw][l]{\tok{71K}} & \cellcolor{gainc!10}\makebox[\accw][r]{\textbf{63.4}}\makebox[\tokw][l]{\tok{71K}} & \cellcolor{gainc!11}\makebox[\accw][r]{\textbf{63.3}}\makebox[\tokw][l]{\tok{86K}} & \cellcolor{gainc!10}\makebox[\accw][r]{\textbf{63.3}}\makebox[\tokw][l]{\tok{86K}} & \cellcolor{lossc!6}\makebox[\accw][r]{\textbf{73.1}}\makebox[\tokw][l]{\tok{58K}} & \cellcolor{gainc!9}\makebox[\accw][r]{\textbf{66.7}}\makebox[\tokw][l]{\tok{61K}} & \cellcolor{gainc!10}\makebox[\accw][r]{\textbf{69.2}}\makebox[\tokw][l]{\tok{64K}} & \cellcolor{gainc!11}\makebox[\accw][r]{\textbf{67.9}}\makebox[\tokw][l]{\tok{85K}} & \cellcolor{gainc!9}\makebox[\accw][r]{\textbf{67.9}}\makebox[\tokw][l]{\tok{85K}} \\
\cstrut Claude Haiku~5.5, medium thinking & \nb{3.8} & \cellcolor{gainc!16}\makebox[\accw][r]{43.3}\makebox[\tokw][l]{\tok{18K}} & \cellcolor{gainc!16}\makebox[\accw][r]{48.2}\makebox[\tokw][l]{\tok{21K}} & \cellcolor{gainc!16}\makebox[\accw][r]{49.7}\makebox[\tokw][l]{\tok{27K}} & \cellcolor{gainc!15}\makebox[\accw][r]{54.5}\makebox[\tokw][l]{\tok{41K}} & \cellcolor{gainc!14}\makebox[\accw][r]{54.5}\makebox[\tokw][l]{\tok{41K}} & \cellcolor{gainc!7}\makebox[\accw][r]{69.2}\makebox[\tokw][l]{\tok{19K}} & \cellcolor{gainc!12}\makebox[\accw][r]{60.3}\makebox[\tokw][l]{\tok{24K}} & \cellcolor{gainc!11}\makebox[\accw][r]{65.4}\makebox[\tokw][l]{\tok{31K}} & \cellcolor{gainc!13}\makebox[\accw][r]{64.1}\makebox[\tokw][l]{\tok{43K}} & \cellcolor{gainc!11}\makebox[\accw][r]{64.1}\makebox[\tokw][l]{\tok{43K}} \\
\cstrut Claude Haiku~5.5, no thinking & \nb{3.8} & \cellcolor{gainc!22}\makebox[\accw][r]{31.2}\makebox[\tokw][l]{\tok{14K}} & \cellcolor{gainc!21}\makebox[\accw][r]{37.5}\makebox[\tokw][l]{\tok{19K}} & \cellcolor{gainc!20}\makebox[\accw][r]{41.3}\makebox[\tokw][l]{\tok{26K}} & \cellcolor{gainc!21}\makebox[\accw][r]{41.7}\makebox[\tokw][l]{\tok{39K}} & \cellcolor{gainc!20}\makebox[\accw][r]{41.7}\makebox[\tokw][l]{\tok{39K}} & \cellcolor{gainc!15}\makebox[\accw][r]{50.0}\makebox[\tokw][l]{\tok{16K}} & \cellcolor{gainc!13}\makebox[\accw][r]{57.7}\makebox[\tokw][l]{\tok{21K}} & \cellcolor{gainc!16}\makebox[\accw][r]{53.8}\makebox[\tokw][l]{\tok{28K}} & \cellcolor{gainc!14}\makebox[\accw][r]{61.5}\makebox[\tokw][l]{\tok{42K}} & \cellcolor{gainc!12}\makebox[\accw][r]{61.5}\makebox[\tokw][l]{\tok{42K}} \\
\cstrut GPT-6-Luna, max thinking & \nb{30.7} & \cellcolor{gainc!15}\makebox[\accw][r]{46.2}\makebox[\tokw][l]{\tok{32K}} & \cellcolor{gainc!16}\makebox[\accw][r]{47.4}\makebox[\tokw][l]{\tok{37K}} & \cellcolor{gainc!13}\makebox[\accw][r]{56.4}\makebox[\tokw][l]{\tok{44K}} & \cellcolor{gainc!14}\makebox[\accw][r]{55.6}\makebox[\tokw][l]{\tok{70K}} & \cellcolor{gainc!28}\makebox[\accw][r]{26.6}\makebox[\tokw][l]{\tok{723K}} & \cellcolor{gainc!14}\makebox[\accw][r]{52.6}\makebox[\tokw][l]{\tok{28K}} & \cellcolor{gainc!19}\makebox[\accw][r]{43.6}\makebox[\tokw][l]{\tok{39K}} & \cellcolor{gainc!17}\makebox[\accw][r]{51.3}\makebox[\tokw][l]{\tok{47K}} & \cellcolor{gainc!18}\makebox[\accw][r]{50.0}\makebox[\tokw][l]{\tok{76K}} & \cellcolor{gainc!18}\makebox[\accw][r]{47.4}\makebox[\tokw][l]{\tok{877K}} \\
\cstrut GPT-6-Luna, medium thinking & \nb{30.7} & \cellcolor{gainc!19}\makebox[\accw][r]{37.7}\makebox[\tokw][l]{\tok{20K}} & \cellcolor{gainc!18}\makebox[\accw][r]{43.6}\makebox[\tokw][l]{\tok{24K}} & \cellcolor{gainc!15}\makebox[\accw][r]{51.7}\makebox[\tokw][l]{\tok{28K}} & \cellcolor{gainc!17}\makebox[\accw][r]{50.3}\makebox[\tokw][l]{\tok{54K}} & \cellcolor{gainc!29}\makebox[\accw][r]{22.8}\makebox[\tokw][l]{\tok{568K}} & \cellcolor{gainc!18}\makebox[\accw][r]{42.3}\makebox[\tokw][l]{\tok{20K}} & \cellcolor{gainc!14}\makebox[\accw][r]{56.4}\makebox[\tokw][l]{\tok{22K}} & \cellcolor{gainc!15}\makebox[\accw][r]{56.4}\makebox[\tokw][l]{\tok{29K}} & \cellcolor{gainc!21}\makebox[\accw][r]{42.3}\makebox[\tokw][l]{\tok{59K}} & \cellcolor{gainc!30}\makebox[\accw][r]{17.9}\makebox[\tokw][l]{\tok{708K}} \\
\cstrut GPT-6-Luna, no thinking & \nb{30.7} & \cellcolor{gainc!30}\makebox[\accw][r]{8.4}\makebox[\tokw][l]{\tok{13K}} & \cellcolor{gainc!30}\makebox[\accw][r]{18.4}\makebox[\tokw][l]{\tok{17K}} & \cellcolor{gainc!27}\makebox[\accw][r]{28.0}\makebox[\tokw][l]{\tok{23K}} & \cellcolor{gainc!30}\makebox[\accw][r]{16.4}\makebox[\tokw][l]{\tok{46K}} & \cellcolor{gainc!30}\makebox[\accw][r]{6.7}\makebox[\tokw][l]{\tok{344K}} & \cellcolor{gainc!30}\makebox[\accw][r]{6.4}\makebox[\tokw][l]{\tok{12K}} & \cellcolor{gainc!30}\makebox[\accw][r]{14.1}\makebox[\tokw][l]{\tok{17K}} & \cellcolor{gainc!27}\makebox[\accw][r]{25.6}\makebox[\tokw][l]{\tok{22K}} & \cellcolor{gainc!30}\makebox[\accw][r]{12.8}\makebox[\tokw][l]{\tok{49K}} & \cellcolor{gainc!30}\makebox[\accw][r]{16.7}\makebox[\tokw][l]{\tok{399K}} \\
\midrule
\bottomrule
\end{tabular}%
}
\end{table*}

\section{Experiments}
\label{sec:exp}

In this section, we test whether ViRA-9B and ViRA-27B are more accurate than existing agents on contrast sets and general benchmarks, whether the gain comes from the joint term, and whether CrossRPO works on text and larger $K$.

\begin{table*}[tp]
\caption{\textbf{General benchmarks, all models}: accuracy (\%) at 1\,MP and at each model's maximum resolution (\emph{Max}).
Tools, +\,tool, $\Delta$ and the colors are as in Table~\ref{tab:main}.
O3-Bench photos have a median of 19\,MP, and V$^*$ and HR-Bench photos have 3.4 to 38.8\,MP.
Most MMStar and HallusionBench images are below 1\,MP (medians 0.18 and 0.53\,MP), so we report them only at the maximum resolution.
ViRA results on V$^*$ and HR-Bench are means of three runs.
Bold: best in each group.}
\label{tab:general-full}\label{tab:general}
\centering
\scriptsize
\setlength{\tabcolsep}{5.60pt}
\settowidth{\accw}{\textbf{00.0}}\settowidth{\tokw}{\tok{000K}}\settowidth{\numw}{\textbf{00.0}}\settowidth{\imgw}{(00.0)}
\begin{tabular}{l*{14}{c}}
\toprule
\midrule
 & \multicolumn{2}{c}{V$^*$} & \multicolumn{2}{c}{HR-Bench 4K} & \multicolumn{2}{c}{HR-Bench 8K} & \multicolumn{2}{c}{MME-RW-Lite} & \multicolumn{2}{c}{TreeBench} & \multicolumn{2}{c}{O3-Bench} & MMStar & Hallusion \\
\cmidrule(lr){2-3}\cmidrule(lr){4-5}\cmidrule(lr){6-7}\cmidrule(lr){8-9}\cmidrule(lr){10-11}\cmidrule(lr){12-13}\cmidrule(lr){14-14}\cmidrule(l){15-15}
Model & 1\,MP & Max & 1\,MP & Max & 1\,MP & Max & 1\,MP & Max & 1\,MP & Max & 1\,MP & Max & Max & Max \\
\midrule
\multicolumn{15}{l}{\emph{Open 7--9B models}} \\
Qwen3.5-9B & \nb{74.9} & \nb{82.2} & \nb{64.9} & \nb{83.2} & \nb{60.2} & \nb{79.2} & \nb{52.8} & \nb{55.5} & \nb{44.2} & \nb{46.7} & \nb{27.2} & \nb{53.3} & \nb{70.8} & \nb{74.1} \\
ZwZ2-9B & \nb{82.2} & \nb{90.1} & \nb{72.9} & \nb{87.9} & \nb{65.4} & \nb{84.8} & \nb{57.1} & \nb{61.3} & \nb{49.6} & \nb{48.4} & \nb{26.1} & \nb{59.4} & \nb{76.3} & \nb{78.5} \\
Vision-OPD-9B & \nb{80.6} & \nb{94.2} & \nb{67.9} & \nb{83.0} & \nb{63.2} & \nb{80.9} & \nb{57.6} & \nb{61.5} & \nb{46.7} & \nb{47.7} & \nb{30.1} & \nb{53.6} & \nb{72.7} & \nb{74.1} \\
\midrule[0.3pt]
\multicolumn{15}{l}{\emph{Open 7--9B models + tool}} \\
Qwen3.5-9B + tool & \nb{83.8} & \nb{90.6} & \nb{76.9} & \nb{82.5} & \nb{70.6} & \nb{77.1} & \nb{55.0} & \nb{57.5} & \nb{48.9} & \nb{49.4} & \nb{34.5} & \nb{45.2} & \nb{70.7} & \nb{73.2} \\
ZwZ2-9B + tool & \nb{86.9} & \nb{94.8} & \nb{82.0} & \nb{88.0} & \nb{76.2} & \nb{87.9} & \nb{61.2} & \nb{63.8} & \nb{50.6} & \nb{48.9} & \nb{40.0} & \nb{50.7} & \nb{75.0} & \nb{75.8} \\
Vision-OPD-9B + tool & \nb{77.0} & \nb{92.7} & \nb{74.9} & \nb{83.9} & \nb{69.0} & \nb{80.9} & \nb{55.4} & \nb{57.6} & \nb{46.7} & \nb{47.4} & \nb{26.7} & \nb{49.9} & \nb{70.3} & \nb{74.4} \\
\midrule[0.3pt]
\multicolumn{15}{l}{\emph{Released zoom agents (7--9B)}} \\
Mini-o3 & \nb{86.4} & \nb{86.9} & \nb{72.0} & \nb{71.8} & \nb{68.5} & \nb{64.1} & \nb{45.0} & \nb{47.5} & \nb{40.0} & \nb{41.2} & \nb{18.8} & \nb{18.6} & \nb{58.7} & \nb{62.3} \\
DeepEyes & \nb{73.8} & \nb{83.2} & \nb{68.4} & \nb{74.0} & \nb{60.9} & \nb{69.9} & \nb{49.3} & \nb{55.0} & \nb{39.8} & \nb{41.2} & \nb{26.7} & \nb{27.2} & \nb{61.2} & \nb{65.9} \\
DeepEyesV2 & \nb{72.8} & \nb{83.8} & \nb{64.9} & \nb{75.4} & \nb{61.0} & \nb{69.1} & \nb{49.5} & \nb{53.4} & \nb{40.0} & \nb{40.7} & \nb{24.9} & \nb{29.6} & \nb{63.2} & \nb{69.8} \\
Thyme & \nb{75.9} & \nb{82.7} & \nb{71.9} & \nb{79.1} & \nb{64.8} & \nb{71.8} & \nb{50.1} & \nb{54.1} & \nb{41.2} & \nb{39.8} & \nb{24.3} & \nb{30.7} & \nb{63.9} & \nb{68.8} \\
Pixel Reasoner & \nb{74.9} & \nb{84.8} & \nb{68.6} & \nb{73.6} & \nb{60.6} & \nb{67.8} & \nb{48.8} & \nb{52.9} & \nb{41.7} & \nb{41.5} & \nb{26.1} & \nb{26.4} & \nb{62.7} & \nb{68.6} \\
ZoomEye & \nb{81.7} & \nb{86.4} & \nb{71.9} & \nb{71.8} & \nb{67.5} & \nb{70.2} & \nb{43.1} & \nb{46.4} & \nb{40.0} & \nb{41.0} & \nb{29.0} & \nb{30.1} & \nb{61.8} & \nb{67.4} \\
\midrule[0.3pt]
\multicolumn{15}{l}{\emph{Ours + our zoom tool}} \\
\shaderow{vira9bg}\textbf{ViRA-9B} (ours) & \nb{\textbf{92.0}} & \nb{\textbf{96.2}} & \nb{\textbf{87.5}} & \nb{\textbf{91.0}} & \nb{\textbf{84.7}} & \nb{\textbf{90.1}} & \nb{\textbf{63.6}} & \nb{\textbf{65.3}} & \nb{\textbf{56.3}} & \nb{\textbf{57.0}} & \nb{\textbf{50.4}} & \nb{\textbf{67.0}} & \nb{\textbf{77.1}} & \nb{\textbf{79.6}} \\
\quad $\Delta$ vs.\ Qwen3.5-9B + tool & \dgain{+8.2} & \dgain{+5.6} & \dgain{+10.6} & \dgain{+8.5} & \dgain{+14.1} & \dgain{+13.0} & \dgain{+8.6} & \dgain{+7.8} & \dgain{+7.4} & \dgain{+7.6} & \dgain{+15.9} & \dgain{+21.8} & \dgain{+6.4} & \dgain{+6.4} \\
\midrule[1pt]
\multicolumn{15}{l}{\emph{Open 27B models}} \\
Qwen3.8-27B & \nb{73.8} & \nb{83.8} & \nb{69.1} & \nb{79.4} & \nb{61.4} & \nb{76.8} & \nb{54.5} & \nb{57.7} & \nb{51.9} & \nb{54.3} & \nb{27.5} & \nb{58.8} & \nb{68.0} & \nb{72.8} \\
ZwZ2-27B & \nb{79.6} & \nb{93.7} & \nb{79.9} & \nb{90.8} & \nb{73.6} & \nb{87.8} & \nb{61.8} & \nb{\textbf{65.6}} & \nb{54.3} & \nb{56.3} & \nb{46.1} & \nb{74.8} & \nb{79.7} & \nb{80.3} \\
\midrule[0.3pt]
\multicolumn{15}{l}{\emph{Open 27B models + tool}} \\
Qwen3.8-27B + tool & \nb{84.8} & \nb{92.1} & \nb{84.0} & \nb{88.5} & \nb{75.9} & \nb{81.4} & \nb{55.6} & \nb{59.8} & \nb{52.1} & \nb{55.6} & \nb{47.2} & \nb{59.7} & \nb{77.0} & \nb{80.3} \\
ZwZ2-27B + tool & \nb{83.2} & \nb{88.5} & \nb{85.4} & \nb{86.9} & \nb{82.5} & \nb{82.5} & \nb{62.5} & \nb{63.3} & \nb{54.6} & \nb{55.1} & \nb{53.9} & \nb{58.3} & \nb{76.5} & \nb{77.8} \\
\midrule[0.3pt]
\multicolumn{15}{l}{\emph{Released zoom agents (27--32B)}} \\
LBS-27B & \nb{92.7} & \nb{94.2} & \nb{88.2} & \nb{89.2} & \nb{89.0} & \nb{89.6} & \nb{\textbf{63.7}} & \nb{65.3} & \nb{65.2} & \nb{67.4} & \nb{53.6} & \nb{78.3} & \nb{79.7} & \nb{83.0} \\
SenseNova-MARS-32B & \nb{90.6} & \nb{92.7} & \nb{86.8} & \nb{90.0} & \nb{81.9} & \nb{85.0} & \nb{59.9} & \nb{61.3} & \nb{53.8} & \nb{53.8} & \nb{45.2} & \nb{61.4} & \nb{75.3} & \nb{75.6} \\
\midrule[0.3pt]
\multicolumn{15}{l}{\emph{Ours + our zoom tool}} \\
\shaderow{vira27bg}\textbf{ViRA-27B} (ours) & \nb{\textbf{95.5}} & \nb{\textbf{97.6}} & \nb{\textbf{91.0}} & \nb{\textbf{92.0}} & \nb{\textbf{90.5}} & \nb{\textbf{91.0}} & \nb{63.4} & \nb{64.5} & \nb{\textbf{67.4}} & \nb{\textbf{68.1}} & \nb{\textbf{64.3}} & \nb{\textbf{84.1}} & \nb{\textbf{82.1}} & \nb{\textbf{86.0}} \\
\quad $\Delta$ vs.\ Qwen3.8-27B + tool & \dgain{+10.7} & \dgain{+5.5} & \dgain{+7.0} & \dgain{+3.5} & \dgain{+14.6} & \dgain{+9.6} & \dgain{+7.8} & \dgain{+4.7} & \dgain{+15.3} & \dgain{+12.5} & \dgain{+17.1} & \dgain{+24.4} & \dgain{+5.1} & \dgain{+5.7} \\
\midrule[1pt]
\multicolumn{15}{l}{\emph{Proprietary models + our zoom tool}} \\
\cstrut Claude Haiku~5.5, max thinking & \cellcolor{gainc!13}\nb{\textbf{89.5}} & \cellcolor{gainc!10}\nb{\textbf{94.2}} & \cellcolor{lossc!7}\nb{\textbf{91.8}} & \cellcolor{gainc!6}\nb{\textbf{91.9}} & \cellcolor{gainc!6}\nb{\textbf{90.2}} & \cellcolor{lossc!8}\nb{\textbf{92.2}} & \cellcolor{gainc!13}\nb{57.9} & \cellcolor{gainc!12}\nb{60.0} & \cellcolor{gainc!11}\nb{\textbf{63.7}} & \cellcolor{gainc!8}\nb{\textbf{66.4}} & \cellcolor{lossc!23}\nb{\textbf{78.3}} & \cellcolor{lossc!7}\nb{85.2} & \cellcolor{gainc!10}\nb{79.0} & \cellcolor{gainc!9}\nb{\textbf{83.9}} \\
\cstrut Claude Haiku~5.5, medium thinking & \cellcolor{gainc!18}\nb{85.9} & \cellcolor{gainc!10}\nb{\textbf{94.2}} & \cellcolor{gainc!9}\nb{88.8} & \cellcolor{gainc!8}\nb{90.1} & \cellcolor{gainc!11}\nb{86.4} & \cellcolor{gainc!6}\nb{90.9} & \cellcolor{gainc!12}\nb{58.3} & \cellcolor{gainc!9}\nb{\textbf{62.1}} & \cellcolor{gainc!13}\nb{61.5} & \cellcolor{gainc!12}\nb{63.5} & \cellcolor{lossc!23}\nb{\textbf{78.3}} & \cellcolor{lossc!11}\nb{\textbf{87.8}} & \cellcolor{gainc!11}\nb{78.3} & \cellcolor{gainc!9}\nb{\textbf{83.9}} \\
\cstrut Claude Haiku~5.5, no thinking & \cellcolor{gainc!24}\nb{81.2} & \cellcolor{gainc!14}\nb{91.1} & \cellcolor{gainc!12}\nb{86.4} & \cellcolor{gainc!11}\nb{88.1} & \cellcolor{gainc!17}\nb{81.4} & \cellcolor{gainc!9}\nb{88.4} & \cellcolor{gainc!14}\nb{57.0} & \cellcolor{gainc!13}\nb{58.9} & \cellcolor{gainc!21}\nb{55.6} & \cellcolor{gainc!18}\nb{58.5} & \cellcolor{gainc!9}\nb{61.7} & \cellcolor{gainc!18}\nb{74.2} & \cellcolor{gainc!13}\nb{76.5} & \cellcolor{gainc!11}\nb{82.1} \\
\cstrut GPT-6-Luna, max thinking & \cellcolor{gainc!20}\nb{84.3} & \cellcolor{gainc!21}\nb{85.9} & \cellcolor{lossc!6}\nb{91.2} & \cellcolor{gainc!9}\nb{89.2} & \cellcolor{gainc!12}\nb{85.4} & \cellcolor{gainc!30}\nb{66.6} & \cellcolor{gainc!13}\nb{57.6} & \cellcolor{gainc!17}\nb{55.5} & \cellcolor{gainc!13}\nb{61.5} & \cellcolor{gainc!13}\nb{62.2} & \cellcolor{lossc!21}\nb{76.5} & \cellcolor{gainc!11}\nb{80.3} & \cellcolor{gainc!9}\nb{79.3} & \cellcolor{gainc!12}\nb{81.4} \\
\cstrut GPT-6-Luna, medium thinking & \cellcolor{gainc!23}\nb{81.7} & \cellcolor{gainc!15}\nb{90.1} & \cellcolor{gainc!7}\nb{90.6} & \cellcolor{gainc!8}\nb{90.0} & \cellcolor{gainc!11}\nb{86.8} & \cellcolor{gainc!30}\nb{64.9} & \cellcolor{gainc!8}\nb{\textbf{62.0}} & \cellcolor{gainc!12}\nb{60.0} & \cellcolor{gainc!17}\nb{58.3} & \cellcolor{gainc!17}\nb{59.3} & \cellcolor{lossc!21}\nb{75.9} & \cellcolor{gainc!14}\nb{77.7} & \cellcolor{gainc!8}\nb{\textbf{80.1}} & \cellcolor{gainc!12}\nb{81.1} \\
\cstrut GPT-6-Luna, no thinking & \cellcolor{gainc!30}\nb{65.4} & \cellcolor{gainc!30}\nb{74.9} & \cellcolor{gainc!26}\nb{74.9} & \cellcolor{gainc!30}\nb{72.8} & \cellcolor{gainc!30}\nb{66.0} & \cellcolor{gainc!30}\nb{56.8} & \cellcolor{gainc!22}\nb{50.4} & \cellcolor{gainc!23}\nb{50.9} & \cellcolor{gainc!30}\nb{47.7} & \cellcolor{gainc!27}\nb{50.9} & \cellcolor{gainc!30}\nb{33.9} & \cellcolor{gainc!30}\nb{40.6} & \cellcolor{gainc!26}\nb{66.3} & \cellcolor{gainc!23}\nb{72.3} \\
\midrule
\bottomrule
\end{tabular}
\end{table*}

\begin{figure*}[!t]
  \centering
  \includegraphics[width=\textwidth]{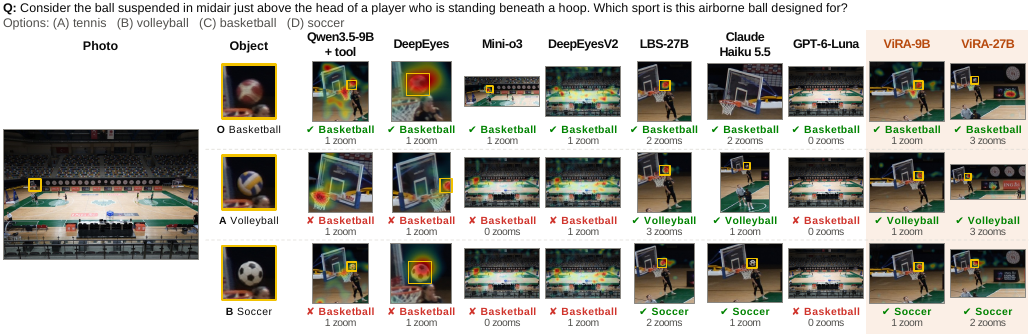}
  \caption{\textbf{Where the agents look} on one contrast set of ViRA-Bench at 1\,MP.
  Each tile is the agent's last crop, or the whole photo if it received no crop.
  For the open models, colors show the attention of the final reply to each part of the tile, relative to a request for a general description of the image~\cite{zhang2025mllmsknow}, so red marks where the reply looks more than usual.
  Appendix~\ref{sec:supp-cases} gives the details and 14 other sets.}
  \label{fig:heat}
\end{figure*}

\subsection{Setup}
\label{sec:exp-setup}

\textbf{Training.}
We train Qwen3.5-9B and Qwen3.8-27B with CrossRPO ($\lambda=0.5$) and obtain ViRA-9B and ViRA-27B.
For each base model, we first remove the contrast sets that it solves without zooming at 12.8\,MP and the single images that it answers correctly in at least seven or in none of eight attempts, because the advantages in Eq.~(\ref{eq:method-advantage}) are zero when the rollouts of a version are all correct or all wrong.
ViRA-9B then trains on 3,642 items and ViRA-27B on 3,401 (Table~\ref{tab:train-mix}), fewer than any trained baseline reports (Table~\ref{tab:models}).
For ViRA-9B, these are 1,664 contrast sets and 1,978 single images, and for ViRA-27B, 1,338 contrast sets and 2,063 single images.
A contrast set counts as one item: we sample $n=12/K$ rollouts per version, so a set receives 12 rollouts in total, the same as one input of GRPO.
Each photo is viewed at a budget drawn log-uniformly between 0.5 and 16.8\,MP with up to six zooms (Appendix~\ref{sec:supp-impl}).
The reward is the correctness of the final answer, and GPT-4o judges open answers against the reference answer.
Each model makes one pass over its items with 16 items per step, which is 227 steps for ViRA-9B and 212 steps for ViRA-27B, and training takes 12.7 and 29.6 hours on eight H200 GPUs (Appendix~\ref{sec:supp-training}).

\textbf{Benchmarks.}
We evaluate on ten benchmarks (Appendix~\ref{sec:supp-benchmark-list}).
ViRA-Bench and ViRA-Bench-Real measure set accuracy at 0.5, 1, 2 and 4\,MP and at each model's maximum.
ViRA-Bench has 762 sets, and ViRA-Bench-Real has 78.
A model's maximum is the largest input that it accepts, which is 16.8\,MP for the open models of the Qwen3 family, 12.8\,MP for those based on Qwen2.5-VL, 3.8\,MP for Claude Haiku~5.5 and 30.7\,MP for GPT-6-Luna.
The other eight are \emph{general benchmarks}, which measure accuracy on single-image questions at 1\,MP and the maximum: V$^*$~\cite{wu2024vstar}, HR-Bench 4K and 8K~\cite{wang2025hrbench}, MME-RealWorld-Lite~\cite{zhang2025mmerealworld}, TreeBench~\cite{wang2026treebench}, O3-Bench~\cite{li2026insighto3}, MMStar~\cite{chen2024mmstar} and HallusionBench~\cite{guan2024hallusionbench}.
Most images of MMStar and HallusionBench are below 1\,MP, so we report these two benchmarks only at the maximum.
The budget caps the pixels of the first view and of every crop.
We also report the tokens per set, which are the prompt tokens of all turns plus the generated tokens, summed over the versions of the set.

\textbf{Baselines.}
The open baselines Qwen3.5-9B, ZwZ2-9B~\cite{wei2026zoomingv2}, Vision-OPD-9B~\cite{yuan2026visionopd}, Qwen3.8-27B and ZwZ2-27B run without a tool and with a zoom tool (\emph{+\,tool}).
ZwZ2 and Vision-OPD are trained to read fine details without a tool, and with the zoom tool every open baseline may zoom up to six times.
The released agents use their own tools: Mini-o3~\cite{lai2025minio3}, DeepEyes~\cite{zheng2025deepeyes}, DeepEyesV2~\cite{hong2025deepeyesv2}, Thyme~\cite{zhang2025thyme}, Pixel Reasoner~\cite{wang2025pixelreasoner} and ZoomEye~\cite{shen2025zoomeye} (7--9B), and LBS-27B~\cite{yao2026lbs} and SenseNova-MARS-32B~\cite{chng2025sensenovamars} (27--32B).
The 7--9B agents are based on Qwen2.5-VL-7B and keep the tools, prompts and turn limits they were trained with, and ZoomEye searches a tree of crops without training.
Claude Haiku~5.5 and GPT-6-Luna call our zoom tool through the function calling of their APIs, at max, medium and no thinking.
Decoding is greedy, and ViRA and the ablations are averaged over three runs on ViRA-Bench, ViRA-Bench-Real (Appendices~\ref{sec:supp-baseline-list} and~\ref{sec:supp-eval}).

\begin{table*}[t]
\caption{\textbf{Ablations}: set accuracy (\%) on ViRA-Bench and ViRA-Bench-Real, and accuracy (\%) on the general benchmarks.
The column heads are pixel budgets in MP, and \emph{Max} is the maximum resolution.
All models are trained from Qwen3.5-9B.
Results on ViRA-Bench, ViRA-Bench-Real are means~of~three~runs.}
\label{tab:ablation}
\centering
\scriptsize
\setlength{\tabcolsep}{1.80pt}
\settowidth{\numw}{\textbf{00.0}}
\resizebox{\textwidth}{!}{%
\begin{tabular}{l*{24}{c}}
\toprule
\midrule
 & \multicolumn{5}{c}{ViRA-Bench} & \multicolumn{5}{c}{ViRA-Bench-Real} & \multicolumn{2}{c}{V$^*$} & \multicolumn{2}{c}{HR-4K} & \multicolumn{2}{c}{HR-8K} & \multicolumn{2}{c}{MME-RW} & \multicolumn{2}{c}{TreeBench} & \multicolumn{2}{c}{O3-Bench} & MMStar & Hallusion \\
\cmidrule(lr){2-6}\cmidrule(lr){7-11}\cmidrule(lr){12-13}\cmidrule(lr){14-15}\cmidrule(lr){16-17}\cmidrule(lr){18-19}\cmidrule(lr){20-21}\cmidrule(lr){22-23}\cmidrule(lr){24-24}\cmidrule(l){25-25}
Advantage & 0.5 & 1 & 2 & 4 & Max & 0.5 & 1 & 2 & 4 & Max & 1 & Max & 1 & Max & 1 & Max & 1 & Max & 1 & Max & 1 & Max & Max & Max \\
\midrule
\shaderow{vira9bg}\textbf{CrossRPO} ($\lambda=0.5$) & \nb{38.7} & \nb{46.9} & \nb{\textbf{55.6}} & \nb{\textbf{57.5}} & \nb{\textbf{63.3}} & \nb{31.2} & \nb{49.6} & \nb{57.7} & \nb{68.8} & \nb{\textbf{77.4}} & \nb{92.0} & \nb{\underline{96.2}} & \nb{87.5} & \nb{\underline{91.0}} & \nb{\underline{84.7}} & \nb{\textbf{90.1}} & \nb{\textbf{63.6}} & \nb{\textbf{65.3}} & \nb{\textbf{56.3}} & \nb{\textbf{57.0}} & \nb{\underline{50.4}} & \nb{\underline{67.0}} & \nb{77.1} & \nb{79.6} \\
\midrule[0.3pt]
\multicolumn{25}{l}{\emph{Joint term}} \\
$\lambda=0$ (per-version RLOO) & \nb{31.3} & \nb{41.2} & \nb{50.0} & \nb{52.9} & \nb{58.7} & \nb{23.5} & \nb{41.9} & \nb{56.8} & \nb{65.4} & \nb{70.1} & \nb{\textbf{93.0}} & \nb{95.6} & \nb{\underline{87.8}} & \nb{\textbf{92.9}} & \nb{\textbf{85.7}} & \nb{\underline{89.8}} & \nb{\underline{62.4}} & \nb{\underline{64.1}} & \nb{54.3} & \nb{55.3} & \nb{43.5} & \nb{64.1} & \nb{\underline{78.8}} & \nb{\underline{80.2}} \\
$\lambda=1$ & \nb{33.4} & \nb{44.2} & \nb{51.3} & \nb{56.1} & \nb{56.6} & \nb{32.9} & \nb{49.6} & \nb{59.0} & \nb{67.1} & \nb{69.2} & \nb{89.5} & \nb{94.6} & \nb{85.4} & \nb{89.8} & \nb{83.4} & \nb{89.1} & \nb{60.2} & \nb{64.0} & \nb{52.8} & \nb{53.6} & \nb{47.8} & \nb{62.3} & \nb{77.4} & \nb{78.3} \\
$\lambda=2$ & \nb{\underline{40.9}} & \nb{\underline{47.9}} & \nb{\underline{53.8}} & \nb{55.7} & \nb{58.2} & \nb{\underline{52.1}} & \nb{\underline{59.0}} & \nb{\underline{68.8}} & \nb{69.7} & \nb{74.8} & \nb{87.3} & \nb{93.9} & \nb{82.4} & \nb{88.9} & \nb{81.8} & \nb{86.8} & \nb{60.3} & \nb{62.6} & \nb{52.3} & \nb{53.1} & \nb{40.3} & \nb{58.3} & \nb{76.5} & \nb{77.0} \\
Random partners & \nb{33.8} & \nb{42.1} & \nb{47.5} & \nb{51.7} & \nb{57.3} & \nb{32.5} & \nb{48.3} & \nb{52.1} & \nb{65.0} & \nb{71.4} & \nb{89.2} & \nb{95.8} & \nb{86.0} & \nb{90.4} & \nb{82.7} & \nb{89.5} & \nb{58.4} & \nb{60.8} & \nb{\underline{54.8}} & \nb{54.3} & \nb{\textbf{53.0}} & \nb{\underline{67.0}} & \nb{75.3} & \nb{79.9} \\
\midrule[0.3pt]
\multicolumn{25}{l}{\emph{Existing RL methods}} \\
Per-version GRPO & \nb{\textbf{44.0}} & \nb{\textbf{50.2}} & \nb{53.4} & \nb{\underline{56.3}} & \nb{59.3} & \nb{\textbf{55.6}} & \nb{\textbf{66.2}} & \nb{\textbf{74.4}} & \nb{\textbf{73.5}} & \nb{\underline{75.6}} & \nb{89.5} & \nb{95.3} & \nb{86.0} & \nb{90.8} & \nb{84.6} & \nb{88.8} & \nb{59.5} & \nb{60.6} & \nb{53.6} & \nb{54.8} & \nb{43.8} & \nb{58.6} & \nb{75.1} & \nb{79.2} \\
Pooled GRPO & \nb{38.0} & \nb{44.9} & \nb{51.6} & \nb{53.9} & \nb{52.6} & \nb{44.9} & \nb{53.8} & \nb{59.8} & \nb{65.4} & \nb{70.1} & \nb{88.3} & \nb{94.8} & \nb{\textbf{88.2}} & \nb{89.5} & \nb{84.2} & \nb{87.0} & \nb{61.0} & \nb{61.8} & \nb{52.1} & \nb{54.1} & \nb{48.1} & \nb{61.2} & \nb{75.7} & \nb{79.4} \\
Pooled RLOO & \nb{37.2} & \nb{43.7} & \nb{52.2} & \nb{55.7} & \nb{\underline{61.0}} & \nb{44.9} & \nb{58.1} & \nb{68.4} & \nb{\underline{70.9}} & \nb{\underline{75.6}} & \nb{87.3} & \nb{94.2} & \nb{82.8} & \nb{87.8} & \nb{81.0} & \nb{84.5} & \nb{59.6} & \nb{60.6} & \nb{54.3} & \nb{\underline{55.6}} & \nb{40.3} & \nb{58.8} & \nb{75.2} & \nb{78.2} \\
ConsistRoll~\cite{zou2026consistency} & \nb{23.5} & \nb{33.4} & \nb{40.8} & \nb{47.1} & \nb{53.1} & \nb{16.2} & \nb{29.9} & \nb{49.6} & \nb{56.0} & \nb{61.5} & \nb{\underline{92.5}} & \nb{\textbf{97.9}} & \nb{85.2} & \nb{90.6} & \nb{81.2} & \nb{89.2} & \nb{58.7} & \nb{62.3} & \nb{54.1} & \nb{52.6} & \nb{\underline{50.4}} & \nb{\textbf{69.0}} & \nb{\textbf{79.3}} & \nb{\textbf{82.1}} \\
\midrule
\bottomrule
\end{tabular}%
}
\end{table*}

\subsection{Main results}
\label{sec:exp-main}

Table~\ref{tab:main} gives the set accuracy and the tokens per set of every model on ViRA-Bench and ViRA-Bench-Real, and Table~\ref{tab:general} gives the accuracy of every model on the general benchmarks.
Both tables group the models into the 7--9B models, the 27--32B models and the proprietary models, and a $\Delta$ row gives the gain of ViRA over its base model with the zoom tool.
ViRA-9B and ViRA-27B are the most accurate open models of their sizes on ten and nine of~the~ten~\mbox{benchmarks}.

\textbf{ViRA-Bench.}
Without a zoom tool, no open model solves more than 15.1\% of the sets below its maximum resolution (Table~\ref{tab:main}).
The gain of ViRA-9B over its base model with the zoom tool grows with the budget, from 17.6 to 28.7 points.
The released 7--9B agents stay below the untrained Qwen3.5-9B with the zoom tool at every budget: they reach a per-image accuracy of 30 to 51\% (Table~\ref{tab:image-acc}) but solve at most 21.1\% of the sets.
ViRA-9B is also ahead of the open 27B models with the zoom tool, and ViRA-27B, after RL on only 3,401 items, is ahead of every proprietary model at every budget.
Claude Haiku~5.5 at max thinking uses about 2 to 6 times as many tokens as ViRA-9B (Figure~\ref{fig:teaser}b) and is ahead of it below the maximum resolution.
Both proprietary models are less accurate at every budget with medium thinking and again without thinking, and GPT-6-Luna at max thinking falls from 55.6\% at 4\,MP to 26.6\% at its maximum resolution.
Figure~\ref{fig:heat} shows where nine agents look on one contrast set.

\textbf{ViRA-Bench-Real.}
ViRA-9B and ViRA-27B also lead the open models of their sizes on photographed sets (Table~\ref{tab:real-full}), so their lead does not depend on edited images.
Without a zoom tool, no open model solves more than 5.1\% of the sets below its maximum resolution.
The released 7--9B agents reach a per-image accuracy of 30 to 60\% (Table~\ref{tab:image-acc}) but solve at most 28.2\% of the sets.
ViRA-27B is within 0.5 points of Claude Haiku~5.5 at 0.5\,MP and ahead of~it~from~1\,MP~on.

\textbf{General benchmarks.}
ViRA-9B gains 5.6 to 21.8 points over its base model with the zoom tool, and ViRA-27B gains 3.5 to 24.4 points (Table~\ref{tab:general}).
Among the open models of their sizes, ViRA-9B is the most accurate on all eight benchmarks and ViRA-27B on seven.
ZoomEye searches a tree of crops with Qwen2.5-VL-7B and is not trained.
On V$^*$ and HR-Bench, its accuracy is within the range of the five trained agents, but it is 9.8 to 19.9 points below ViRA-9B.
ViRA-27B is ahead of both proprietary models on five of the eight benchmarks.

\subsection{Ablations}
\label{sec:exp-ablation}

All ablations (Table~\ref{tab:ablation}) start from Qwen3.5-9B with the same data, reward, protocol and number of steps as ViRA-9B, and only the advantage changes (Appendix~\ref{sec:supp-ablation-settings}).

\textbf{Joint term.}
Without the joint term ($\lambda=0$), each version is trained separately with RLOO, and set accuracy on ViRA-Bench drops at every budget, by 4.6 to 7.4 points.
On the general benchmarks, accuracy changes by at most 2 points, except on O3-Bench, where it drops by 6.9 points at 1\,MP.
Without the joint term, the last zoom covers the object as often, but accuracy at 1\,MP is 2.4 points lower on the originals and 5.2 and 5.3 points lower on the variants (Appendix~\ref{sec:supp-reading}), so the joint term improves reading, not finding.

\textbf{Random partners.}
The versions in a batch are assigned to the sets at random, so the weight $w_v$ of an image is computed from images of other questions.
Set accuracy then stays close to that of $\lambda=0$, as Proposition~\ref{prop:independent} predicts.

\textbf{Value of $\lambda$.}
With $\lambda=2$, set accuracy is higher at low budgets but lower at the maximum resolution and on every general benchmark, and $\lambda=1$ is lower on ViRA-Bench at every budget.
From 2\,MP on, $\lambda=0.5$ is the most accurate variant~on~\mbox{ViRA-Bench}.
With $\lambda=2$, the agent also zooms 2.7 times per image and uses 34K to 346K tokens per set, against 12K to 147K (Table~\ref{tab:ablation-bench}).

\textbf{One group per version.}
Per-version GRPO is $\lambda=0$ with the advantages of each version divided by their standard deviation.
It is behind CrossRPO from 2\,MP on ViRA-Bench and on every general benchmark, but ahead at 0.5 and 1\,MP on ViRA-Bench and up to 4\,MP on ViRA-Bench-Real.
Its agent zooms two to three times per image where CrossRPO zooms once, and uses about twice as many tokens (Appendix~\ref{sec:supp-box}).

\textbf{One group per set.}
Pooled RLOO is $\lambda=0$ with the leave-one-out mean taken over all rollouts of the set.
On ViRA-Bench, pooled GRPO and pooled RLOO are behind CrossRPO at every budget.
On ViRA-Bench-Real, both are ahead up to 2\,MP, and pooled RLOO is ahead because it draws smaller zoom boxes (Appendix~\ref{sec:supp-box}).
ConsistRoll~\cite{zou2026consistency}, pooled GRPO with a pair bonus, is behind on both benchmarks at every budget.

\subsection{Generalizability and Scalability}
\label{sec:exp-general}

\textbf{Generalization to text.}
The joint term does not depend on images.
It needs only sets of inputs that share one question and have different answers.
Questions about passages that contradict known facts give such sets in text.
There, the model's memory plays the role of common sense, and the passage plays the role of the evidence.
We therefore train Qwen3.5-9B on text contrast sets, without images and without the zoom tool.

\emph{Data.}
ConFiQA~\cite{bi2024contextdpo} asks questions about facts from Wikidata, and its passages were written by an LLM from these facts.
We use its three subsets, which have single-hop questions (QA), multi-hop questions (MR) and multi-hop questions with several replaced facts (MC).
Each ConFiQA item has the original passage and a copy in which the answer entity is replaced.
A set with $K=3$ has the original passage (O), this copy (A) and one more copy (B), in which we replace the answer of A with another entity of the same Wikidata relation.
No substituted passage mentions the original answer, so an answer from the model's memory is correct only on O.
We hold out 1,660 questions of the three subsets for testing.
Of the other questions, we train only on the 2,856 that the base model answers correctly without the passage, so its memory gives the original answer on every version.
The 493 test questions with this property form the strong-prior subset in Table~\ref{tab:text}.

\emph{Training.}
The model reads the passage and the question, reasons, and then answers with a word or phrase in a single turn.
The reward is accuracy.
An answer is correct if it contains the answer of its version or an alias, and it is wrong if it also names the answer of another version.
Each step samples 16 sets, and each version has four rollouts, so a set receives 12 rollouts as in the image runs.
CrossRPO uses $\lambda=0.5$, and $\lambda=0$, pooled GRPO and pooled RLOO are defined as in Appendix~\ref{sec:supp-ablation-settings}.
All runs use the same data and 178 steps, which is about one pass over the training questions.
We train each method three times with different random seeds.

\emph{Test sets.}
The ConFiQA test questions come from the same source as the training questions.
NQ-Swap~\cite{longpre2021entity} and FaithEval~\cite{ming2025faitheval} come from other sources, and we did not train on them.
NQ-Swap takes questions and Wikipedia passages from Natural Questions and replaces the answer entity in the passage.
We test 1,000 random NQ-Swap questions, each with the substituted and the original passage.
FaithEval (counterfactual task) has 1,000 multiple-choice science questions from ARC-Challenge.
For each question, an LLM wrote a passage that supports a counterfactual answer, and this answer is the correct option.
The model answers with the letter of an option.
ConFiQA and NQ-Swap answers are scored with the rule of the reward.

\emph{Results.}
Table~\ref{tab:text} gives, for ConFiQA, the share of test sets with all three versions answered correctly, the accuracy on each version, and the rate at which an answer on A or B is the original fact.
For NQ-Swap, it gives the accuracy on substituted and on original passages and the rate of answers with the original fact, and for FaithEval the accuracy.
Averaged over the three training runs, CrossRPO answers all three versions of 81.9\% of the test sets, against 81.0\% for $\lambda=0$, 79.6\% for pooled GRPO, 80.2\% for pooled RLOO and 58.6\% for the base model.
It also gives the original fact on a substituted passage least often.
On the strong-prior questions, it is on par with pooled GRPO.
On NQ-Swap with substituted passages and on FaithEval, all four methods are within 2.0 points.

\begin{table*}[t]
\caption{\textbf{Text contrast sets} (ConFiQA, $K=3$, Qwen3.5-9B): accuracy (\%) and the rate of answering with the original fact (lower is better). NQ-Swap and FaithEval were not trained on. Strong-prior questions: the base model answers them correctly without the passage.
Each method is trained three times with different random seeds, and each run and the base model are evaluated three times.
Cells give the mean and the standard deviation over the three training runs. Bold: best in each row.}
\label{tab:text}
\centering
\footnotesize
\newcommand{\sd}[1]{{\tiny$\,\pm$#1}}
\setlength{\tabcolsep}{3.2pt}
\settowidth{\numw}{\textbf{00.0}}
\begin{tabular*}{\textwidth}{@{\extracolsep{\fill}}lccccc@{}}
\toprule
\midrule
 & Base & $\lambda=0$ & Pooled GRPO & Pooled RLOO & CrossRPO \\
\midrule
ConFiQA, all three correct & \nb{58.6} & \nb{81.0}\sd{1.3} & \nb{79.6}\sd{1.8} & \nb{80.2}\sd{1.7} & \nb{\textbf{81.9}}\sd{3.1} \\
\quad strong-prior questions (493) & \nb{64.3} & \nb{90.5}\sd{0.4} & \nb{\textbf{91.5}}\sd{3.5} & \nb{90.3}\sd{1.0} & \nb{\textbf{91.5}}\sd{2.0} \\
\quad O correct & \nb{76.0} & \nb{87.1}\sd{1.7} & \nb{86.4}\sd{2.2} & \nb{87.0}\sd{1.6} & \nb{\textbf{87.8}}\sd{2.3} \\
\quad A correct & \nb{75.7} & \nb{90.6}\sd{1.2} & \nb{90.5}\sd{1.1} & \nb{90.0}\sd{1.2} & \nb{\textbf{91.1}}\sd{1.5} \\
\quad B correct & \nb{71.8} & \nb{90.6}\sd{1.5} & \nb{\textbf{91.0}}\sd{0.4} & \nb{89.4}\sd{1.7} & \nb{\textbf{91.0}}\sd{1.6} \\
\quad A or B with the original fact & \nb{2.4} & \nb{0.1}\sd{0.1} & \nb{0.1}\sd{0.1} & \nb{0.2}\sd{0.1} & \nb{\textbf{0.0}}\sd{0.0} \\
NQ-Swap, substituted passage & \nb{70.8} & \nb{\textbf{86.4}}\sd{4.1} & \nb{86.3}\sd{1.2} & \nb{84.9}\sd{1.0} & \nb{85.2}\sd{3.0} \\
NQ-Swap, original fact given & \nb{4.8} & \nb{2.1}\sd{1.0} & \nb{1.8}\sd{0.2} & \nb{2.3}\sd{1.2} & \nb{\textbf{1.7}}\sd{0.7} \\
NQ-Swap, original passage & \nb{90.1} & \nb{92.0}\sd{0.3} & \nb{90.4}\sd{1.6} & \nb{91.6}\sd{1.1} & \nb{\textbf{93.1}}\sd{1.0} \\
FaithEval, counterfactual & \nb{62.2} & \nb{\textbf{79.6}}\sd{6.1} & \nb{78.7}\sd{2.8} & \nb{77.6}\sd{7.0} & \nb{78.7}\sd{3.8} \\
\midrule
\bottomrule
\end{tabular*}

\end{table*}

\textbf{Scaling to larger sets.}
Image contrast sets with more than three versions are hard to build, because each further version needs another edit of the same small object that gives a different answer and passes review.
Text sets have no such limit, so we test how CrossRPO scales with $K$ on ConFiQA.
For $K>3$, each set adds versions in which the answer entity is replaced by other entities of the same Wikidata relation, as in B.
The training sets have 2,645 strong-prior questions for $K=4$ and 8 and 2,627 for $K=16$.
Every version has four rollouts, and at each $K$ the four methods use the same data and number of steps, about one pass over the training questions.
The weight of a version in the joint term is then computed from pairs of the other versions (Eq.~(\ref{eq:objective})).
We test on 1,660 questions, and on 1,653 for $K=16$.
Again, each method is trained three times.
Table~\ref{tab:textk} gives the share of test sets with all versions answered correctly and the accuracy per version, and it repeats $K=3$ for comparison.
At $K=4$, 8 and 16, CrossRPO answers the most sets correctly on average.
It is 1.0 to 8.0 points ahead of $\lambda=0$, 0.8 to 2.4 points ahead of pooled GRPO and 0.2 to 4.6 points ahead of pooled RLOO.
Its accuracy per version is also the highest for every $K$, at 91.5\%, 93.4\% and 94.6\%, which is 0.8 to 1.6 points above pooled GRPO and 0.1 to 2.1 points above pooled RLOO.
These text experiments are small, and we leave larger ones to future work.

\begin{table*}[t]
\caption{\textbf{Text sets with $K$ versions} (ConFiQA, Qwen3.5-9B): accuracy (\%) on the test questions at the $K$ used in training.
For $K>3$, the joint term of CrossRPO uses pairs of other versions (Eq.~(\ref{eq:objective})).
Each method is trained three times with different random seeds, and each run is evaluated three times.
Cells give the mean and the standard deviation over the three training runs. Bold: best in each row and group.}
\label{tab:textk}
\centering
\footnotesize
\newcommand{\sd}[1]{{\tiny$\,\pm$#1}}
\setlength{\tabcolsep}{2.4pt}
\settowidth{\numw}{\textbf{00.0}}
\begin{tabular*}{\textwidth}{@{\extracolsep{\fill}}lcccc@{\hspace{18pt}}cccc@{}}
\toprule
\midrule
 & \multicolumn{4}{c}{All versions correct} & \multicolumn{4}{c}{Accuracy per version} \\
\cmidrule(lr{19.6pt}){2-5}\cmidrule(l{1.6pt}){6-9}
$K$ & $\lambda=0$ & Pooled GRPO & Pooled RLOO & CrossRPO & $\lambda=0$ & Pooled GRPO & Pooled RLOO & CrossRPO \\
\midrule
3 & \nb{81.0}\sd{1.3} & \nb{79.6}\sd{1.8} & \nb{80.2}\sd{1.7} & \nb{\textbf{81.9}}\sd{3.1} & \nb{89.4}\sd{1.1} & \nb{89.3}\sd{1.2} & \nb{88.8}\sd{1.4} & \nb{\textbf{90.0}}\sd{1.8} \\
4 & \nb{79.2}\sd{3.0} & \nb{81.2}\sd{2.4} & \nb{79.0}\sd{1.2} & \nb{\textbf{82.0}}\sd{3.1} & \nb{89.7}\sd{1.3} & \nb{90.7}\sd{1.0} & \nb{89.4}\sd{0.7} & \nb{\textbf{91.5}}\sd{1.0} \\
8 & \nb{73.4}\sd{3.8} & \nb{79.7}\sd{0.7} & \nb{76.8}\sd{4.2} & \nb{\textbf{81.4}}\sd{3.5} & \nb{91.0}\sd{1.4} & \nb{92.6}\sd{0.8} & \nb{91.4}\sd{0.9} & \nb{\textbf{93.4}}\sd{1.3} \\
16 & \nb{77.9}\sd{1.7} & \nb{76.5}\sd{2.3} & \nb{78.7}\sd{5.6} & \nb{\textbf{78.9}}\sd{1.0} & \nb{93.2}\sd{1.1} & \nb{93.0}\sd{1.4} & \nb{94.5}\sd{0.9} & \nb{\textbf{94.6}}\sd{0.8} \\
\midrule
\bottomrule
\end{tabular*}

\end{table*}

\section{Conclusion}
\label{sec:conclusion}

Visual agents often answer from common sense.
In this paper, we propose CrossRPO, which trains visual agents on contrast sets, rewards solving the whole set, and provably does not reinforce a set answered from common sense.
We build ViRA-Train and ViRA-Bench by editing one small object in photos of 21 to 120\,MP, and photograph ViRA-Bench-Real.
CrossRPO achieves 63.3\% set accuracy on ViRA-Bench at maximum resolution, against 59.3\% for GRPO.
After RL on about 3,500 items each, ViRA-9B leads open models of its size on all benchmarks, and ViRA-27B outperforms Claude Haiku~5.5 and GPT-6-Luna at their highest thinking level on most benchmarks.

\section*{Limitations}

\textbf{Data.}
ViRA-Bench and ViRA-Train are made by image editing, so traces inside an edited object could affect the answer.
The first zoom covers the object equally often on the edited and the unedited versions (Appendix~\ref{sec:supp-traces}), so the edits do not guide the search, but the test does not cover reading.
ViRA-Bench-Real has no edits, but it has only 78 sets, which were shot indoors with one camera and show mostly posters, signs, stickers and labels.
A photographed benchmark at the scale of ViRA-Bench, with outdoor scenes and more kinds of objects, would be a stronger test.
Each question also describes where the object is, which makes the search easier than a question without this description would.
The image sets have at most three versions, so larger sets are tested only on text.

\textbf{Method.}
CrossRPO requires sets whose versions differ only in the evidence and have different answers (Proposition~\ref{prop:shortcut}).
Such sets need an image editor or a second photo of the scene, so they cost more to build than single images.
Its gain over $\lambda=0$ is mainly on contrast sets.
A single image has no other version, so its cross-input weight is zero, and CrossRPO then trains it as RLOO does.

\textbf{Agents.}
The reward is accuracy only, so nothing encourages tight boxes or further zooms.
ViRA-9B usually zooms once with a large box (Appendix~\ref{sec:supp-box}).
At low budgets, a very small object then stays small in the crop, and a reward on the object's size in the last view could lead to tighter boxes.

\textbf{Evaluation.}
Only one family of base models is trained.
ViRA-9B and ViRA-27B start from Qwen3.5-9B and Qwen3.8-27B, and we have not tested CrossRPO on other architectures.
The baselines run with their own tools and prompts, and the proprietary models use our tool through their APIs.
The ablations of Table~\ref{tab:ablation} share the data, the reward and the tool, so they isolate the effect of the advantage.

\textbf{Text.}
The text experiments are small.
They train one base model for one pass over fewer than 3,000 questions of one dataset.
They show that the joint term transfers to text, not the accuracy that it can reach there.
We leave text experiments at a larger scale to future work.

\section*{LLM Usage Statement}
LLMs and image editing models are part of our data pipeline and training.
Appendices~\ref{sec:supp-bench}, \ref{sec:supp-train} and~\ref{sec:supp-training} name each model and its role in our pipeline.
Human reviewers check every ViRA-Bench question.
For ViRA-Bench-Real, Claude~Code (Anthropic) proposed and designed the scenes and changes, and humans selected the proposals, took the photos, and verified and reviewed every set.
Claude~Code and Codex (OpenAI) also assisted with the prototype, the implementation and the writing of this paper.
The authors reviewed all AI-assisted work and take full responsibility for the code, data, claims, contributions, results and text.

\pagebreak

{
    \small
    \bibliographystyle{ieeenat_fullname}
    \bibliography{main}
}

\clearpage
\maketitlesupplementary
\appendix
\raggedbottom
\makeatletter
\@beginparpenalty=\@M
\makeatother

\startcontents[appendix]
\section*{Contents}
\begingroup
\def\numberline#1{#1\hskip1em}
\makeatletter\let\tocaddvspace\addvspace\def\addvspace#1{\tocaddvspace{0.3em}}\makeatother
\printcontents[appendix]{}{1}{\setcounter{tocdepth}{2}}
\endgroup
\newpage

\section{Theory of CrossRPO}
\label{sec:theory}

This section derives CrossRPO from first principles and proves every statement.
The argument has four steps.
(i)~We formalize policies, sets of inputs and rollout sampling (Appendix~\ref{sec:theory-setup}) and prove the score-function identities behind group-based policy gradients (Appendix~\ref{sec:theory-pg}).
(ii)~We show that accuracy on each version cannot distinguish an answer that is read from the evidence from an answer that is taken from the prior, whereas the accuracy on a whole set can.
We define the objective accordingly (Appendix~\ref{sec:theory-objective}).
(iii)~For an arbitrary set-level reward, we construct a family of unbiased gradient estimators.
The CrossRPO advantage is its closed form for our objective (Appendix~\ref{sec:theory-advantage}).
(iv)~The estimator involves three design choices, and prior work makes different ones.
We compare them under four requirements: unbiasedness (Appendix~\ref{sec:theory-unbiased}), variance (Appendix~\ref{sec:theory-variance}), credit assignment (Appendix~\ref{sec:theory-credit}) and scale (Appendix~\ref{sec:theory-scale}).
Appendix~\ref{sec:theory-general} extends the construction to all correctness-based set rewards, shows that for $K\ge3$ the joint term is not a sum of pairwise terms, and recovers RLOO and GRPO as special cases.
Appendix~\ref{sec:theory-grouping} shows which inputs should form a set.
The joint term helps only if the members share a shortcut and have different answers, and for members grouped at random it reduces in expectation to a constant reweighting of RLOO.
Appendix~\ref{sec:theory-related} locates prior methods in the framework (Table~\ref{tab:related}).
Appendix~\ref{sec:theory-check} checks three of the results on logged rollouts, and Appendix~\ref{sec:theory-summary} summarizes the argument (Table~\ref{tab:summary}).
Only elementary probability is assumed.

\begin{table*}[htbp]
\caption{Prior methods within the framework of Appendix~\ref{sec:theory}.
$K$, \emph{Answers} and $\lambda$ give the special case of the objective Eq.~(\ref{eq:objective}) that a method optimizes (Proposition~\ref{prop:related}).
(D1)--(D3) are the design choices of Appendix~\ref{sec:theory-advantage}: how the credit for answering all versions is computed (Theorem~\ref{thm:rb} and Corollary~\ref{cor:set-accuracy}), which rollouts form the baseline (Propositions~\ref{prop:baseline}, \ref{prop:balance} and~\ref{prop:prior}), and what the advantages are divided by (Proposition~\ref{prop:scale}).
A group mean and a leave-one-out mean differ by a constant factor (Lemma~\ref{lem:loo}).
$^\dagger$Not covered by Eq.~(\ref{eq:objective}): NoisyRollout scores rollouts sampled on a noisy copy under the clean input, and BPO rewards a rollout for agreeing or disagreeing with an anchor rollout.
CRPO weights the copy by $w_{\mathrm{aug}}$ and matches Eq.~(\ref{eq:objective}) for $w_{\mathrm{aug}}=1$.}
\label{tab:related}
\centering
\scriptsize
\begin{tabular*}{\textwidth}{@{\extracolsep{\fill}}lcccccc@{}}
\toprule
\midrule
Method & $K$ & Answers & $\lambda$ & Joint term (D1) & Baseline (D2) & Scale (D3) \\
\midrule
RLOO~\cite{ahmadian2024back}             & 1 & --           & 0              & --               & leave-one-out            & none \\
Dr.~GRPO~\cite{liu2025understanding}     & 1 & --           & 0              & --               & group mean               & none \\
GRPO~\cite{shao2024deepseekmath}         & 1 & --           & 0              & --               & group mean               & per group \\
NoisyRollout$^\dagger$~\cite{liu2025noisyrollout} & 1 & same & 0           & --               & pooled with noisy copies & per group \\
COMET~\cite{zhu2026comet}                & 2 & same         & 0              & --               & pooled over versions     & per group \\
DNA-Train~\cite{huang2025taming}         & 2 & differ       & 0              & --               & per version              & per version, $\ell_1$ \\
ConsistRoll~\cite{zou2026consistency}    & 2 & same         & 0.5            & index pairing    & pooled over versions     & per group \\
CSPO~\cite{li2026cfcamo}                 & 2 & differ       & $1/(1+\gamma)$ & index pairing    & over pairs               & not stated \\
CRPO~\cite{du2026crpo}                   & 2 & per question & $\lambda_z$    & all combinations & per version              & per set \\
BPO$^\dagger$~\cite{kang2026bpo}         & 4 & differ       & --             & --               & anchor rollout           & not stated \\
\midrule
\textbf{CrossRPO}                        & any & any        & any            & all combinations & per version              & per batch \\
\midrule
\bottomrule
\end{tabular*}
\end{table*}

\begin{table*}[htbp]
\caption{Summary of Appendix~\ref{sec:theory}: each requirement, the choice CrossRPO makes, the alternative used in prior work, and what the alternative loses, with the result that shows it.}
\label{tab:summary}
\centering
\scriptsize
\begin{tabular*}{\textwidth}{@{\extracolsep{\fill}}>{\raggedright\arraybackslash}m{2.3cm}>{\raggedright\arraybackslash}m{3.1cm}>{\raggedright\arraybackslash}m{3.4cm}>{\raggedright\arraybackslash}m{7.3cm}@{}}
\toprule
\midrule
Requirement & CrossRPO & Alternative in prior work & What the alternative loses \\
\midrule
Target & set accuracy $\prod_vp_v$ for $K\le3$, all-correct triples for $K>3$ & per-version accuracy; agreement &
  a blind policy reaches $p_v=1$ on one version, and agreement is maximized by blind policies; the set accuracy loses its weight as $K$ grows (Propositions~\ref{prop:blind} and~\ref{prop:agreement}, Corollary~\ref{cor:threshold} and Proposition~\ref{prop:order}) \\
\midrule[0.3pt]
Unbiasedness & leave-one-out within each version & division by group statistics &
  a bias from statistics that contain the rollout itself~\cite{liu2025understanding}; CrossRPO is unbiased for every $R$ (Theorem~\ref{thm:unbiased}) \\
\midrule[0.3pt]
Variance (D1) & all combinations & pairing by index &
  covariance larger by $\E[\Cov(\hat g^b_\Pi\mid O)]\succeq0$; first-order variance ratio $\rho_K(p)\to0$ as $K$ grows (Theorem~\ref{thm:rb} and Corollary~\ref{cor:set-accuracy}) \\
\midrule[0.3pt]
Variance (D2) & per-version baseline & baseline pooled over versions &
  an extra $\Delta_v^2$ from the difficulty gap between versions (Proposition~\ref{prop:baseline}) \\
\midrule[0.3pt]
Credit (D2) & per-version baseline and $w_v$ & baseline pooled over versions &
  answers from the prior get $+\sqrt{K-1}$ instead of 0, and credit shifts toward easier versions (Propositions~\ref{prop:balance} and~\ref{prop:prior}) \\
\midrule[0.3pt]
Scale (D3) & one statistic per batch & standard deviation per group or set &
  a cost difference of any size $\epsilon>0$ becomes the advantage $-\sqrt{N-1}$ (Proposition~\ref{prop:scale}) \\
\midrule[0.3pt]
Generality & any $K$, any set-level reward $R$ & $K\le2$, or a sum over all pairs &
  prior objectives are special cases and RLOO and GRPO are recovered (Propositions~\ref{prop:multilinear}, \ref{prop:special}, \ref{prop:related} and~\ref{prop:crpo}); for $K\ge3$ no sum over pairs gives the joint term, which bounds blind policies by $K^{-3}$ instead of $K^{-2}$ (Lemma~\ref{lem:additivity}, Proposition~\ref{prop:not-pairwise} and Corollary~\ref{cor:pairwise}) \\
\midrule[0.3pt]
Grouping & members share a shortcut, answers differ & copies with the same answer; unrelated inputs &
  with equal answers a shortcut can answer every copy; with unrelated members the joint term is, in expectation, a constant reweighting of RLOO plus variance (Propositions~\ref{prop:shortcut} and~\ref{prop:independent}) \\
\midrule
\bottomrule
\end{tabular*}
\end{table*}

\begin{example}[Running example]
  \label{ex:running}
  Consider a contrast set with $K=2$ variants of one photo that differ only in the year written on a sign above a window.
  The question is ``What year is written on the sign above the window?'', with $y_{\mathrm A}=1987$ and $y_{\mathrm B}=2015$.
  A policy that outputs ``1987'' without reading the sign has accuracies $(p_{\mathrm A},p_{\mathrm B})=(1,0)$, whereas a policy that reads the sign has $(1,1)$.
\end{example}

\subsection{Probabilistic setting}
\label{sec:theory-setup}

\begin{definition}[Policy and rollouts]
  \label{def:policy}
  A rollout on an input $s$ is a sequence $o=(a_1,\dots,a_T)$ of $T\le T_{\max}$ generated tokens from a finite vocabulary.
  After token $a_t$, the environment returns an observation $e_t=\tau(s,a_{1:t})$ from a deterministic map $\tau$.
  The observation is a crop of $s$ resized to the budget when $a_{1:t}$ completes a zoom call, and it is empty otherwise.
  With the history $h_t=(s,a_{1:t-1},e_{1:t-1})$, the policy $\pi_\theta(\cdot\mid h_t)$ is a distribution over tokens that is positive and continuously differentiable in $\theta\in\mathbb R^d$.
  The probability of a rollout is
  \begin{equation}
    P_\theta(o\mid s)=\prod_{t=1}^{T}\pi_\theta(a_t\mid h_t).
    \label{eq:rollout-prob}
  \end{equation}
  The set $\mathcal O$ of rollouts is finite.
\end{definition}

\begin{definition}[Sets, correctness and rewards]
  \label{def:set}
  A \emph{set} $\mathcal S=(q,(s_v,y_v)_{v\in\mathcal V})$ with $\mathcal V=\{1,\dots,K\}$ consists of a question $q$ and $K$ inputs $s_v$ with correct answers $y_v$.
  We call these inputs the \emph{versions} of the set.
  A rollout $o$ on $s_v$ has correctness $c(o)=\mathbf 1[\mathrm{ans}(o)=y_v]\in\{0,1\}$ and a bounded reward $r(o)$, for example $r=c-\kappa(\#\text{zooms})$.
  $\mathcal S$ is a \emph{contrast set}~\cite{gardner2020evaluating} if its versions coincide outside a region $\Omega$ and $y_u\neq y_v$ for $u\neq v$.
  Natural images that share a question, transformed copies of one image, and single inputs ($K=1$) also form sets.
  For version $v$ we write $P_v=P_\theta(\cdot\mid s_v)$ and
  \begin{equation}
    p_v=\E_{P_v}[c],\qquad V_v=\E_{P_v}[r],\qquad \sigma_v^2=\Var_{P_v}[r].
    \label{eq:version-stats}
  \end{equation}
\end{definition}

\begin{assumption}[Sampling]
  \label{asm:sampling}
  For each set, we draw $n\ge2$ rollouts per version.
  The rollouts $o_{v,i}$ with $v\in\mathcal V$ and $i\in[n]$ are mutually independent, and $o_{v,i}\sim P_v$ under the current parameters.
  No rollout is discarded for its outcome.
\end{assumption}

In our experiments $n=12/K$.
We write $O=(o_{v,i})_{v,i}$, $O_v=(o_{v,i})_{i\in[n]}$ and $O_{-v}=(O_u)_{u\neq v}$, as well as $r_{v,i}=r(o_{v,i})$ and $c_{v,i}=c(o_{v,i})$.
For any array $x_{v,i}$, we write
\begin{equation}
  \bar x_v=\frac1n\sum_{i=1}^nx_{v,i},\qquad \bar x_{v,-i}=\frac1{n-1}\sum_{j\neq i}x_{v,j},\qquad \hat p_v=\bar c_v.
  \label{eq:means}
\end{equation}
We call $\bar x_{v,-i}$ the \emph{leave-one-out} mean.

\subsection{Score-function calculus}
\label{sec:theory-pg}

The \emph{score} of a rollout on version $v$ is
\begin{equation}
  \psi_v(o)=\nabla_\theta\log P_\theta(o\mid s_v)=\sum_{t=1}^{T}\nabla_\theta\log\pi_\theta(a_t\mid h_t),
  \label{eq:score}
\end{equation}
where the second equality follows from Eq.~(\ref{eq:rollout-prob}), because the observations $e_t$ are deterministic functions of $(s_v,a_{1:t})$ and do not depend on $\theta$.
We write $\psi_{v,i}=\psi_v(o_{v,i})$.

\begin{lemma}[Score identity]
  \label{lem:score}
  For every $f:\mathcal O\to\mathbb R$ that does not depend on $\theta$,
  \begin{equation}
    \E_{P_v}[\psi_v f]=\nabla_\theta\E_{P_v}[f],\qquad\text{in particular}\qquad\E_{P_v}[\psi_v]=0.
    \label{eq:score-id}
  \end{equation}
\end{lemma}

\begin{proof}
  $\mathcal O$ is finite and $P_\theta(o\mid s_v)>0$ is differentiable, so gradient and sum commute and $\nabla_\theta P_\theta=P_\theta\,\nabla_\theta\log P_\theta$:
  \begin{align*}
    \nabla_\theta\E_{P_v}[f]
    &=\sum_{o\in\mathcal O}f(o)\,\nabla_\theta P_\theta(o\mid s_v)\\
    &=\sum_{o\in\mathcal O}f(o)\,P_\theta(o\mid s_v)\,\psi_v(o)
    =\E_{P_v}[\psi_vf].
  \end{align*}
  For $f\equiv1$ the left side is $\nabla_\theta1=0$.
\end{proof}

\begin{corollary}[Baselines]
  \label{cor:baseline}
  Let $o\sim P_v$ and let $b$ be a random variable independent of $o$.
  Then $\E[\psi_v(o)\,(f(o)-b)]=\nabla_\theta\E_{P_v}[f]$.
\end{corollary}

\begin{proof}
  By independence and Lemma~\ref{lem:score}, $\E[\psi_v(o)\,b]=\E[\psi_v(o)]\,\E[b]=0$.
\end{proof}

So $\psi_v(o)(f(o)-b)$ is an unbiased estimate of $\nabla_\theta\E_{P_v}[f]$ for any such $b$ (REINFORCE~\cite{williams1992simple}).
The coefficient $A=f(o)-b$ is the \emph{advantage}, and $b$ is the \emph{baseline}.
A baseline close to $\E_{P_v}[f]$ lowers the variance (Appendix~\ref{sec:theory-variance}).

\begin{lemma}[Leave-one-out]
  \label{lem:loo}
  (a)~For i.i.d.\ $o_1,\dots,o_n\sim P_v$, $\E\big[\sum_i\psi_v(o_i)\,(f(o_i)-\bar f_{-i})\big]=n\,\nabla_\theta\E_{P_v}[f]$.
  (b)~For any reals $x_1,\dots,x_n$, $x_i-\bar x=\frac{n-1}{n}\,(x_i-\bar x_{-i})$ and $\sum_i(x_i-\bar x_{-i})=0$.
\end{lemma}

\begin{proof}
  (a)~$\bar f_{-i}$ is a function of $(o_j)_{j\neq i}$, which is independent of $o_i$, so Corollary~\ref{cor:baseline} applies to each $i$.
  (b)~$x_i-\bar x=x_i-\frac1n\big(x_i+(n-1)\bar x_{-i}\big)=\frac{n-1}{n}(x_i-\bar x_{-i})$, and $\sum_i\bar x_{-i}=\frac1{n-1}\sum_i\sum_{j\neq i}x_j=\sum_jx_j$.
\end{proof}

\begin{definition}[Group-based advantages]
  \label{def:group}
  For the rewards $x_1,\dots,x_n$ of the rollouts of one group, RLOO~\cite{kool2019buy,ahmadian2024back} uses $A_i=x_i-\bar x_{-i}$, Dr.~GRPO~\cite{liu2025understanding} uses $A_i=x_i-\bar x$, and GRPO~\cite{shao2024deepseekmath} uses $A_i=(x_i-\bar x)/\mathrm{std}(x)$ with $\mathrm{std}(x)^2=\frac1n\sum_i(x_i-\bar x)^2$.
  Each updates $\theta$ along $\sum_iA_i\psi(o_i)$.
\end{definition}

By Lemma~\ref{lem:loo}(b), RLOO and Dr.~GRPO differ by the constant factor $\frac{n-1}{n}$, and both are unbiased by Lemma~\ref{lem:loo}(a).
GRPO divides by a statistic that depends on every rollout of the group, including $o_i$, which biases the estimate~\cite{liu2025understanding}.
When an input has several versions, a group-based method must also decide which rollouts form a group.
This decision is the subject of the rest of the section.
Its basic tool is the score identity for independent inputs.

\begin{lemma}[Product measures]
  \label{lem:product-score}
  Let $R:\mathcal O^K\to\mathbb R$ not depend on $\theta$, and let $o_v\sim P_v$, $v\in\mathcal V$, be independent.
  Then
  \begin{equation}
    \nabla_\theta\E[R(o_1,\dots,o_K)]=\sum_{v\in\mathcal V}\E\big[\psi_v(o_v)\,R(o_1,\dots,o_K)\big].
    \label{eq:product-score}
  \end{equation}
\end{lemma}

\begin{proof}
  The joint law is $\prod_vP_v(o_v)$, and by the product rule
  \begin{align*}
    \nabla_\theta\prod_{v}P_v(o_v)&=\sum_v\nabla_\theta P_v(o_v)\prod_{u\neq v}P_u(o_u)\\
    &=\Big(\sum_v\psi_v(o_v)\Big)\prod_{u}P_u(o_u).
  \end{align*}
  Multiply by $R$ and sum over the finite set $\mathcal O^K$.
\end{proof}

\subsection{What to optimize}
\label{sec:theory-objective}

\begin{definition}[Set-level reward and objective]
  \label{def:objective}
  A \emph{set-level reward} is a bounded map $R:\mathcal O^K\to\mathbb R$ that takes one rollout per version.
  Its objective is $J_R(\theta)=\E[R(o_1,\dots,o_K)]$ with independent $o_v\sim P_v$.
  Let $N=\sum_vc(o_v)$ and $k=\min(K,3)$.
  For $\lambda\ge0$, CrossRPO uses
  \begin{equation}
    R_\star(o_1,\dots,o_K)=\sum_{v\in\mathcal V}r(o_v)+\lambda\binom{N}{k}\Big/\binom{K-1}{k-1}.
    \label{eq:set-reward}
  \end{equation}
  By linearity and independence, its objective is
  \begin{equation}
    J(\theta)=J_{R_\star}(\theta)=\sum_{v\in\mathcal V}V_v+\lambda\,e_k(p)\Big/\binom{K-1}{k-1},
    \label{eq:objective}
  \end{equation}
  where $e_k(p)=\sum_{T\subseteq\mathcal V,\,|T|=k}\prod_{v\in T}p_v$ is the $k$-th elementary symmetric polynomial of $p=(p_v)_v$.
  For $K\le3$, the joint term is $\lambda\prod_vc(o_v)$ and $J=\sum_vV_v+\lambda\prod_vp_v$.
  Over a data distribution $\mathcal D$ of sets, CrossRPO maximizes $\E_{\mathcal S\sim\mathcal D}[J_{\mathcal S}(\theta)]$, and we use $\lambda=0.5$.
\end{definition}

Here $N$ is the number of versions answered correctly by one independent rollout each, and $\binom Nk$ counts the $k$-subsets of versions that are all answered correctly.
Since $\prod_vc(o_v)=\mathbf 1[N=K]$, the joint term for $K\le3$ is the \emph{set accuracy} $\prod_vp_v=\Pr(N=K)$.
For $K>3$, $e_3(p)/\binom K3$ is the probability that three versions drawn at random are all answered correctly.
We use triples instead of the whole set because the weight of the set accuracy vanishes as $K$ grows, whereas the weight of triples does not (Proposition~\ref{prop:order}).
We call a policy \emph{blind} on a set if $P_u=P_v$ for all $u,v\in\mathcal V$.
On a contrast set, this holds whenever the token distributions depend on the image and on every crop only through the pixels outside $\Omega$.
For instance, it holds when the policy answers from the prior without \mbox{inspecting}~the~region.

\begin{proposition}[Blind policies]
  \label{prop:blind}
  Let the correct answers of a set be pairwise distinct and the policy be blind on it.
  Then
  (a)~$\E[N]=\sum_vp_v\le1$;
  (b)~$\Pr(N\ge m)\le1/m$ for every $m\in[K]$;
  (c)~$\Pr(N=K)=\prod_vp_v\le K^{-K}$;
  (d)~if every $P_v$ is a point mass and $K\ge2$, then $\prod_vp_v=0$;
  (e)~$\E\big[\binom Nk\big]=e_k(p)\le\binom KkK^{-k}$ for every $k\in[K]$, which is a fraction $K^{-k}$ of its value $\binom Kk$ when every version is answered correctly.
  A policy that answers every version correctly has~$N=K$~almost~surely.
\end{proposition}

\begin{proof}
  Let $\rho(a)=P_v(\mathrm{ans}=a)$, which does not depend on $v$ by blindness, so $p_v=\rho(y_v)$.
  (a)~Since the $y_v$ are distinct, $\sum_v\rho(y_v)\le\sum_a\rho(a)\le1$, and the inequality is strict if rollouts may give no answer.
  (b)~By Markov's inequality, $\Pr(N\ge m)\le\E[N]/m\le1/m$.
  (c)~By independence $\Pr(N=K)=\prod_vp_v$, and by the inequality of arithmetic and geometric means $\prod_vp_v\le\big(\frac1K\sum_vp_v\big)^K\le K^{-K}$.
  (d)~A point mass gives one answer $a^\star$ on every version.
  Since $y_v=a^\star$ for at most one $v$, some $p_u=0$.
  (e)~By independence, $\E\big[\binom Nk\big]=\sum_{|T|=k}\prod_{v\in T}p_v=e_k(p)$.
  Maclaurin's inequality~\cite{hardy1952inequalities} gives $\big(e_k(p)/\binom Kk\big)^{1/k}\le\frac1K\sum_vp_v$, and the right side is at most $1/K$ by (a).
\end{proof}

On a single version, a blind policy can be perfect, so per-version accuracy cannot reveal blindness.
In contrast, the joint statistic $N$ can.
In Example~\ref{ex:running}, the policy that outputs ``1987'' attains $p_{\mathrm A}=1$, the bound $\E[N]=1$ of (a) and $\Pr(N=2)=0$, whereas reading attains $\E[N]=2$ and $\Pr(N=2)=1$.
Rewards for agreement~\cite{zou2026consistency} are a natural alternative to Eq.~(\ref{eq:set-reward}) but fail on such sets.

\begin{proposition}[Joint term versus agreement rewards]
  \label{prop:agreement}
  Let $K\ge2$ and let the correct answers of a set be pairwise distinct.
  (a)~The joint term $e_k(p)/\binom{K-1}{k-1}$ of Eq.~(\ref{eq:objective}) is nondecreasing in each $p_v$, and it attains its maximum $K/k$ if and only if $p_v=1$ for all $v$.
  (b)~The agreement reward $R_{\mathrm{agr}}(o_{1:K})=\mathbf 1[\mathrm{ans}(o_1)=\dots=\mathrm{ans}(o_K)]$ attains its maximum $J_{R_{\mathrm{agr}}}=1$ at every deterministic blind policy that always answers, and $R_{\mathrm{agr}}=1$ implies $N\le1$.
  (c)~$\binom Nk\,R_{\mathrm{agr}}\equiv0$ for every $k\ge2$, and in particular $\mathbf 1[N=K]\,R_{\mathrm{agr}}\equiv0$.
\end{proposition}

\begin{proof}
  (a)~Each monomial of $e_k$ is a product of factors in $[0,1]$, so it is nondecreasing in each factor and at most 1.
  Every $p_v$ appears in some monomial, so the sum of the $\binom Kk$ monomials equals $\binom Kk$ only if every $p_v$ is 1.
  Finally, $\binom Kk/\binom{K-1}{k-1}=K/k$.
  (b)~A deterministic blind policy gives the same answer on every version, so $R_{\mathrm{agr}}=1$ almost surely.
  If all answers equal $a$, version $v$ is correct only if $y_v=a$, which holds for at most one $v$.
  (c)~If $R_{\mathrm{agr}}=1$, then $N\le1$ by (b), so $\binom Nk=0$ for $k\ge2$.
\end{proof}

The joint term of Eq.~(\ref{eq:set-reward}) uses only the correctness of each version against its own answer, so it needs no assumption on how the answers relate and applies to any task with a correctness check.
Proposition~\ref{prop:related} shows that the objectives of prior two-version methods are $J_R$ for special cases of $R$.
These methods differ from CrossRPO mainly in their estimate of $\nabla_\theta J_R$, which the rest of~this~section~\mbox{compares}.

\subsection{From the objective to the CrossRPO advantage}
\label{sec:theory-advantage}

\paragraph{Projection onto one version.}
For a set-level reward $R$, we define the \emph{projection} of $R$ onto version $v$ as
\begin{equation}
  \begin{aligned}
    \bar R_v(o)&=\E\big[R(o_{1:K})\,\big|\,o_v=o\big]\\
    &=\sum_{o_{-v}\in\mathcal O^{K-1}}R(o,o_{-v})\prod_{u\neq v}P_u(o_u),
  \end{aligned}
  \label{eq:projection}
\end{equation}
It is the expected set reward when version $v$ produces rollout $o$ and the other versions are sampled from the policy.

\begin{lemma}[Gradient through projections]
  \label{lem:projection}
  $\nabla_\theta J_R=\sum_{v\in\mathcal V}\E_{o\sim P_v}\big[\psi_v(o)\,\bar R_v(o)\big]$.
\end{lemma}

\begin{proof}
  Since $\psi_v(o_v)$ is a function of $o_v$, Lemma~\ref{lem:product-score} and the tower property give
  \begin{align*}
    \E[\psi_v(o_v)R(o_{1:K})]&=\E\big[\psi_v(o_v)\,\E[R(o_{1:K})\mid o_v]\big]\\
    &=\E\big[\psi_v(o_v)\,\bar R_v(o_v)\big].\qedhere
  \end{align*}
\end{proof}

Let $N_{-v}$ be the number of other versions answered correctly, so that $N=c(o_v)+N_{-v}$.
Pascal's rule gives $\binom Nk=c(o_v)\binom{N_{-v}}{k-1}+\binom{N_{-v}}k$, and by independence $\E\big[\binom{N_{-v}}\ell\big]=e_\ell(p_{-v})$.
For $R_\star$, the projection is therefore
\begin{equation*}
  \begin{aligned}
    \bar R_v(o)&=r(o)+\sum_{u\neq v}V_u+\lambda\,w^\star_v\,c(o)+\lambda\,\frac{e_k(p_{-v})}{\binom{K-1}{k-1}},\\
    w^\star_v&=\frac{e_{k-1}(p_{-v})}{\binom{K-1}{k-1}}.
  \end{aligned}
\end{equation*}
The second and last terms are constant in $o$, so by Lemma~\ref{lem:score} they contribute nothing.
Lemma~\ref{lem:projection} then gives
\begin{equation}
  \nabla_\theta J=\sum_v\nabla_\theta V_v+\lambda\sum_vw^\star_v\,\nabla_\theta p_v.
  \label{eq:product-rule}
\end{equation}
The factor $w^\star_v$ is the marginal value of being correct on version $v$.
It is the probability that $k-1$ other versions drawn at random are all answered correctly.
For $K\le3$, these are all the other versions, so $w^\star_v=\prod_{u\neq v}p_u$.

\paragraph{Empirical projection.}
Lemma~\ref{lem:projection} involves the unknown distributions $P_u$ for $u\neq v$.
Replacing them by the empirical distributions of $O_u$ gives the \emph{empirical projection}
\begin{equation}
  \hat R_v(o)=\frac{1}{n^{K-1}}\sum_{\iota\in[n]^{\mathcal V\setminus\{v\}}}R\big(o,(o_{u,\iota_u})_{u\neq v}\big),
  \label{eq:empirical-projection}
\end{equation}
which is the average of $R$ over all $n^{K-1}$ \emph{combinations} of one rollout per other version.

\begin{lemma}[Conditional unbiasedness]
  \label{lem:empirical-projection}
  $\E\big[\hat R_v(o_{v,i})\mid o_{v,i}\big]=\bar R_v(o_{v,i})$.
\end{lemma}

\begin{proof}
  For every $\iota$, the tuple $(o_{u,\iota_u})_{u\neq v}$ is independent of $o_{v,i}$ and distributed as $\bigotimes_{u\neq v}P_u$, so each summand has conditional expectation $\bar R_v(o_{v,i})$ by Eq.~(\ref{eq:projection}).
\end{proof}

\begin{definition}[Cross-input advantage]
  \label{def:general-advantage}
  For a set-level reward $R$, the cross-input advantage of rollout $i$ of version $v$ and the gradient estimate are
  \begin{equation}
    \begin{aligned}
      A^R_{v,i}&=\hat R_v(o_{v,i})-\frac1{n-1}\sum_{j\neq i}\hat R_v(o_{v,j}),\\
      \hat g_R&=\sum_{v\in\mathcal V}\sum_{i=1}^nA^R_{v,i}\,\psi_{v,i}.
    \end{aligned}
    \label{eq:general-advantage}
  \end{equation}
\end{definition}

$A^R_{v,i}$ measures how much the outcome of rollout $(v,i)$ raises the set reward relative to the other rollouts of its version.
Here the set reward is averaged over all combinations of rollouts on~the~other~\mbox{versions}.

\begin{proposition}[Closed form for CrossRPO]
  \label{prop:closed-form}
  For $R=R_\star$,
  \begin{equation*}
    \hat R_v(o_{v,i})=r_{v,i}+\sum_{u\neq v}\bar r_u+\lambda\,w_v\,c_{v,i}+\hat\beta_v,\qquad w_v=\frac{e_{k-1}(\hat p_{-v})}{\binom{K-1}{k-1}},
  \end{equation*}
  where $\hat\beta_v=\lambda\,e_k(\hat p_{-v})/\binom{K-1}{k-1}$ does not depend on $i$.
  For $K\le3$, $w_v=\prod_{u\neq v}\hat p_u$, and for $K>3$, $w_v$ is the mean of $\hat p_u\hat p_{u'}$ over the pairs of other versions.
  The cross-input advantage is the \emph{CrossRPO advantage}
  \begin{equation}
    A_{v,i}=\big(r_{v,i}-\bar r_{v,-i}\big)+\lambda\,w_v\big(c_{v,i}-\bar c_{v,-i}\big),
    \label{eq:advantage}
  \end{equation}
  with $w_v=1$ (the empty product) when $K=1$.
\end{proposition}

\begin{proof}
  By linearity, the average in Eq.~(\ref{eq:empirical-projection}) splits into the terms of Eq.~(\ref{eq:set-reward}).
  The term $r(o_{v,i})$ does not depend on $\iota$.
  For $u\neq v$, the term $r(o_{u,\iota_u})$ depends only on $\iota_u$, and each value of $\iota_u$ occurs $n^{K-2}$ times, so its average is $\bar r_u$.
  For the joint term, Pascal's rule splits $\binom Nk$ into $c_{v,i}\binom{N_{-v}}{k-1}$ and $\binom{N_{-v}}k$.
  Here $\binom{N_{-v}}\ell$ is the sum of $\prod_{u\in T}c_{u,\iota_u}$ over the $\ell$-subsets $T$ of the other versions.
  Each such product involves distinct coordinates of $\iota$, so it factorizes:
  \begin{equation*}
    \frac{1}{n^{K-1}}\sum_{\iota}\prod_{u\in T}c_{u,\iota_u}=\prod_{u\in T}\Big(\frac1n\sum_{j=1}^nc_{u,j}\Big)=\prod_{u\in T}\hat p_u.
  \end{equation*}
  Summing over $T$ gives $e_{k-1}(\hat p_{-v})$ and $e_k(\hat p_{-v})$, and hence $w_v$ and $\hat\beta_v$.
  For $K\le3$, the only $(k-1)$-subset is $\mathcal V\setminus\{v\}$, and for $K>3$, $e_2(\hat p_{-v})/\binom{K-1}2$ averages $\hat p_u\hat p_{u'}$ over the pairs.
  In $A^{R_\star}_{v,i}$, the terms $\sum_{u\neq v}\bar r_u$ and $\hat\beta_v$ are the same for all $i$ and cancel.
  Since $w_v$ is common to the rollouts of version $v$, this gives Eq.~(\ref{eq:advantage}).
\end{proof}

\begin{definition}[CrossRPO update]
  \label{def:advantage}
  The gradient estimate of a set is $\hat g=\hat g_{R_\star}=\sum_{v,i}A_{v,i}\psi_{v,i}$.
  A training step on a batch $\mathcal B$ of sets moves $\theta$ along $s^{-1}\sum_{\mathcal S\in\mathcal B}\hat g_{\mathcal S}$, where $s$ is the root mean square of all advantages in the batch.
  \begin{equation}
    \hat g=\sum_{v\in\mathcal V}\sum_{i=1}^nA_{v,i}\,\psi_{v,i}.
    \label{eq:ghat}
  \end{equation}
\end{definition}

\paragraph{Interpretation.}
The first term of Eq.~(\ref{eq:advantage}) is RLOO on each version.
The second estimates $\lambda\,w^\star_v\nabla_\theta p_v$ from Eq.~(\ref{eq:product-rule}).
It replaces the marginal value $w^\star_v$ by $w_v$, which is the fraction of all-correct combinations of rollouts on $k-1$ other versions drawn at random.
If $K\le3$ and $\hat p_u>0$ for all $u$, then $w_v=\hat p_v^{-1}\prod_u\hat p_u$, so $w_u\ge w_v$ if and only if $\hat p_u\le\hat p_v$.
The weight is therefore largest for the version with the lowest accuracy.
For $K\le3$, it vanishes if another version is never answered correctly.
For $K>3$, such a version removes only the pairs that contain it, so the other versions keep part of their joint credit.

\begin{example}
  \label{ex:advantage}
  In Example~\ref{ex:running} with $n=4$, let $\hat p_{\mathrm A}=3/4$ and $\hat p_{\mathrm B}=0$.
  Then $w_{\mathrm A}=0$, so correct answers on A receive no joint credit.
  In contrast, $w_{\mathrm B}=3/4$, so once a correct answer on B is sampled, it receives the joint credit $\lambda\cdot\frac34\cdot(1-\bar c_{\mathrm B,-i})$.
\end{example}

\paragraph{Implementation.}
GRPO-style training maximizes the clipped surrogate of PPO~\cite{schulman2017ppo} with importance ratios $\rho_t=\pi_\theta(a_t\mid h_t)/\pi_{\theta_{\mathrm{old}}}(a_t\mid h_t)$.
Each of our batches gives one gradient step, so $\rho_t=1$ and the clipping is inactive when the gradient is taken.
The gradient of the surrogate is then $\sum A\,\nabla_\theta\log\pi_\theta(a_t\mid h_t)$, which is Eq.~(\ref{eq:ghat}), and there is no KL or entropy term.
The advantage of a rollout is assigned to all its generated tokens, and the loss is averaged over the tokens of the batch.
The step is therefore $\sum\hat g/s$ up to one positive factor for the whole batch.
Normalizing each rollout by its own length instead would add a length bias~\cite{liu2025understanding}.

\paragraph{Three design choices.}
Eq.~(\ref{eq:advantage}) fixes three choices, and each of them has an alternative in prior work:
\begin{itemize}
  \item[(D1)] the joint term averages over all combinations ($w_v$) instead of rollouts paired by index;
  \item[(D2)] the baseline of $(v,i)$ is computed from $O_v$ instead of from all versions of the set;
  \item[(D3)] the advantages are divided by one statistic per batch instead of the standard deviation~of~each~group.
\end{itemize}
Appendices~\ref{sec:theory-unbiased}, \ref{sec:theory-variance}, \ref{sec:theory-credit} and~\ref{sec:theory-scale} compare these design choices under four requirements, one~per~\mbox{subsection}.

\subsection{Requirement 1: unbiasedness}
\label{sec:theory-unbiased}

An estimate is unbiased if its expectation equals the gradient of the objective, so that training follows $\nabla_\theta J$ on average.

\begin{theorem}[Unbiasedness]
  \label{thm:unbiased}
  Under Assumption~\ref{asm:sampling}, $\E[\hat g_R]=n\,\nabla_\theta J_R$ for every set-level reward $R$.
  In particular $\E[\hat g]=n\,\nabla_\theta J$.
\end{theorem}

\begin{proof}
  Fix $v$ and $i$.
  \emph{Leading term.}
  By the tower property and Lemma~\ref{lem:empirical-projection},
  \begin{align*}
    \E\big[\psi_{v,i}\,\hat R_v(o_{v,i})\big]&=\E\big[\psi_{v,i}\,\E[\hat R_v(o_{v,i})\mid o_{v,i}]\big]\\
    &=\E_{o\sim P_v}\big[\psi_v(o)\,\bar R_v(o)\big].
  \end{align*}
  \emph{Baseline.}
  $\frac1{n-1}\sum_{j\neq i}\hat R_v(o_{v,j})$ is a function of $(o_{v,j})_{j\neq i}$ and $O_{-v}$, which are independent of $o_{v,i}$, so its product with $\psi_{v,i}$ has expectation 0 by Corollary~\ref{cor:baseline}.
  \emph{Sum.}
  Hence $\E[A^R_{v,i}\psi_{v,i}]=\E_{P_v}[\psi_v\bar R_v]$ for each of the $n$ rollouts of version $v$, and Lemma~\ref{lem:projection} gives $\E[\hat g_R]=n\sum_v\E_{P_v}[\psi_v\bar R_v]=n\,\nabla_\theta J_R$.
  The second claim is the case $R=R_\star$ by Proposition~\ref{prop:closed-form}.
\end{proof}

\begin{remark}
  \label{rem:unbiased}
  (i)~\emph{Unequal sample sizes.}
  If version $v$ has $n_v$ rollouts, weighting its terms by $1/n_v$ gives $\E[\hat g_R]=\nabla_\theta J_R$.
  With $n=12/K$, a set contributes $(12/K)\nabla_\theta J$, so the batch objective weights each set by $1/K$.
  (ii)~\emph{Batch scale.}
  Let the batch consist of $B$ i.i.d.\ sets with $N=Kn=12$ rollouts each, and let $\bar G_B=\frac1B\sum_b\hat g_b$ and $s_B^2=\frac1{BN}\sum_b\sum_{v,i}A_{b,v,i}^2$.
  By the strong law of large numbers, $\bar G_B\to\E[\hat g]$ and $s_B^2\to s_\infty^2=\E[\sum_{v,i}A_{v,i}^2]/N$ almost surely.
  If $s_\infty>0$, the scaled step $\bar G_B/s_B$ therefore converges almost surely to $\E[\hat g]/s_\infty$, which is a positive multiple of the gradient of the batch objective.
  For finite $B$, $s_B$ depends on the sample, so the scaled step is consistent but not exactly unbiased.
  However, $s_B$ changes only the length of the step and~not~its~direction.
\end{remark}

\paragraph{Unbiasedness does not decide (D1)--(D3).}
A baseline pooled over all versions does not depend on $o_{v,i}$ either, so it is unbiased by Corollary~\ref{cor:baseline}.
Pairing rollouts by index without looking at outcomes is unbiased as well (Theorem~\ref{thm:rb}).
GRPO's division by a group statistic adds a bias~\cite{liu2025understanding}, but the alternatives to (D1)--(D3) can be made unbiased.
The choices must therefore be compared by their variance and by which rollouts they credit.

\subsection{Requirement 2: low variance}
\label{sec:theory-variance}

A high variance makes each step noisy and requires more rollouts for the same progress.
Choices (D1) and (D2) affect the variance, as the next two results show.

\paragraph{(D1) All combinations versus a fixed pairing.}
Several methods pair rollouts by index~\cite{zou2026consistency,li2026cfcamo}.
The $k$-th rollouts of all versions form a combination, and its members are credited if the combination succeeds.
A \emph{pairing} $\sigma=(\sigma_v)_{v\in\mathcal V}$ is a tuple of permutations of $[n]$.
Combination $k$ is $o^\sigma_k=(o_{u,\sigma_u(k)})_{u\in\mathcal V}$, and rollout $(v,i)$ lies in combination $k_\sigma(v,i)=\sigma_v^{-1}(i)$.
Pairing by index is $\sigma=\mathrm{id}$.
Some methods credit version $v$ with a version-specific reward.
To cover them, let $R^{(v)}$ for $v\in\mathcal V$ be set-level rewards, with $R^{(v)}=R$ in the basic case.
For baselines $b_{v,i}$, define
\begin{align*}
  \hat g^b_\sigma&=\sum_v\sum_i\psi_{v,i}\big(R^{(v)}(o^\sigma_{k_\sigma(v,i)})-b_{v,i}\big),\\
  \hat g^b_{\mathrm{all}}&=\sum_v\sum_i\psi_{v,i}\big(\hat R^{(v)}_v(o_{v,i})-b_{v,i}\big),
\end{align*}
where $\hat R^{(v)}_v$ is Eq.~(\ref{eq:empirical-projection}) for $R^{(v)}$.
A baseline is \emph{symmetric} if $b_{v,i}$ does not depend on $o_{v,i}$ or on the pairing and depends on $O$ only through the multiset $\{o_{v,j}\}_{j\neq i}$ and the multisets $\{o_{u,j}\}_j$ for $u\neq v$.
The leave-one-out baseline of Definition~\ref{def:general-advantage} is symmetric, so $\hat g_R=\hat g^b_{\mathrm{all}}$ with $R^{(v)}=R$ and $b_{v,i}=\frac1{n-1}\sum_{j\neq i}\hat R_v(o_{v,j})$.

\begin{example}
  \label{ex:pairing}
  In Example~\ref{ex:running} with $n=4$, let rollouts 1--3 on A be correct and rollout 4 wrong, and let only rollout 4 on B be correct.
  Pairing by index puts the correct rollout on B with the wrong rollout on A, so its joint indicator is 0.
  However, 3 of the 4 combinations that contain it are correct on both, which gives $w_{\mathrm B}=3/4$.
\end{example}

\begin{lemma}[Random pairing]
  \label{lem:random-pairing}
  Let $\Pi$ have independent uniformly random components $\Pi_v$, independent of $O$.
  Then $\E\big[R^{(v)}(o^\Pi_{k_\Pi(v,i)})\mid O\big]=\hat R^{(v)}_v(o_{v,i})$.
\end{lemma}

\begin{proof}
  Given $\Pi_v$, rollout $(v,i)$ lies in combination $k=\Pi_v^{-1}(i)$.
  For $u\neq v$, $\Pi_u$ is uniform and independent of $\Pi_v$ and of $O$, so $(\Pi_u(k))_{u\neq v}$ is uniform on $[n]^{\mathcal V\setminus\{v\}}$ given $(O,\Pi_v)$.
  The conditional expectation is therefore the average of $R^{(v)}(o_{v,i},(o_{u,\iota_u})_{u\neq v})$ over $\iota\in[n]^{\mathcal V\setminus\{v\}}$, which is the empirical projection in Eq.~(\ref{eq:empirical-projection}).
\end{proof}

\begin{theorem}[All combinations]
  \label{thm:rb}
  For every fixed pairing $\sigma$ and every symmetric baseline $b$, $\E[\hat g^b_\sigma]=\E[\hat g^b_{\mathrm{all}}]$ and
  \begin{equation}
    \Cov(\hat g^b_\sigma)-\Cov(\hat g^b_{\mathrm{all}})=\E\big[\Cov(\hat g^b_\Pi\mid O)\big]\succeq0.
    \label{eq:rb}
  \end{equation}
  In particular $\E\|\hat g^b_\sigma-\E\hat g^b_\sigma\|^2\ge\E\|\hat g^b_{\mathrm{all}}-\E\hat g^b_{\mathrm{all}}\|^2$, with equality only if $\hat g^b_\sigma=\hat g^b_{\mathrm{all}}$ almost surely.
\end{theorem}

\begin{proof}
  \emph{Re-indexing.}
  Let $O^\sigma$ be $O$ re-indexed within each version by $o^\sigma_{v,k}=o_{v,\sigma_v(k)}$.
  Substituting $i=\sigma_v(k)$ and using the symmetry of $b$ gives $\hat g^b_\sigma(O)=\hat g^b_{\mathrm{id}}(O^\sigma)$.
  Under Assumption~\ref{asm:sampling}, the rollouts of each version are i.i.d.\ and the versions are independent, so $O^\sigma\overset{d}{=}O$ for every fixed $\sigma$.
  Hence $\hat g^b_\sigma(O)\overset{d}{=}\hat g^b_{\mathrm{id}}(O)\overset{d}{=}\hat g^b_\Pi(O)=:G$, where the last equality holds because $\Pi$ is independent of $O$.

  \emph{Conditioning.}
  Neither $\psi_{v,i}$ nor $b_{v,i}$ depends on the pairing, so Lemma~\ref{lem:random-pairing} gives $\E[G\mid O]=\hat g^b_{\mathrm{all}}$.

  \emph{Total covariance.}
  $\Cov(G)=\Cov(\E[G\mid O])+\E[\Cov(G\mid O)]$, and the second term is positive semidefinite.
  Since $\hat g^b_\sigma\overset{d}{=}G$, $\E[\hat g^b_\sigma]=\E[G]=\E[\hat g^b_{\mathrm{all}}]$ and Eq.~(\ref{eq:rb}) follows.
  The traces coincide only if $\Cov(G\mid O)=0$ almost surely, which means that the estimate does not depend on the pairing at all.
\end{proof}

By Lemma~\ref{lem:random-pairing}, $\hat g^b_{\mathrm{all}}=\E[\hat g^b_\Pi\mid O]$ for a uniformly random pairing $\Pi$.
A fixed pairing therefore has the same mean as $\hat g^b_{\mathrm{all}}$ and adds the covariance $\E[\Cov(\hat g^b_\Pi\mid O)]$ of Eq.~(\ref{eq:rb}).
$\hat g^b_{\mathrm{all}}$ is its Rao--Blackwellization~\cite{blackwell1947conditional} over the pairing, and $\hat R_v$ is a U-statistic~\cite{hoeffding1948class}.
Eq.~(\ref{eq:rb}) states this in matrix form, so the variance does not increase in any direction of the parameter space.
Zhou~\etal~\cite{zhou2026demystifying} show that the GRPO gradient with a leave-one-out baseline and without division by the standard deviation is a second-order U-statistic over pairs of rollouts of one input, and they derive its mean squared error.
The per-version term of CrossRPO is that estimator applied to each version.
The joint term is a U-statistic of another kind, because its kernel takes one rollout from every version and so combines rollouts across inputs.
Theorem~\ref{thm:rb} does not cover pairings chosen after observing outcomes, and such pairings change the target.
For example, matching correct rollouts with each other turns the fraction of all-correct combinations into $\min_v\hat p_v$ instead of $\prod_v\hat p_v$.
For the set accuracy itself, which is the joint statistic for $K\le3$, the reduction in variance has a closed form.

\begin{corollary}[Estimating set accuracy]
  \label{cor:set-accuracy}
  Let $P=\prod_vp_v$, $a_v=p_v(1-p_v)$, $\hat P_{\mathrm{idx}}=\frac1n\sum_k\prod_vc_{v,k}$ and $\hat P_{\mathrm{all}}=\prod_v\hat p_v$.
  Both are unbiased for $P$, $\Var(\hat P_{\mathrm{all}})\le\Var(\hat P_{\mathrm{idx}})$ for every $n$, and
  \begin{align}
    \Var(\hat P_{\mathrm{idx}})&=\frac{P(1-P)}{n},
    \label{eq:var-idx}\\
    \Var(\hat P_{\mathrm{all}})&=\prod_v\Big(p_v^2+\frac{a_v}{n}\Big)-P^2=\sum_{\emptyset\neq T\subseteq\mathcal V}\frac{1}{n^{|T|}}\prod_{v\in T}a_v\prod_{v\notin T}p_v^2.
    \label{eq:var-all}
  \end{align}
  If all $p_v>0$ and $P<1$, the ratio of the leading terms in $1/n$ is
  \begin{equation}
    \rho=\frac{P^2\sum_v(1-p_v)/p_v}{P(1-P)}\le1,
    \label{eq:var-ratio}
  \end{equation}
  For equal accuracies $p_v\equiv p\in(0,1)$, this ratio is $\rho_K(p)=\frac{Kp^{K-1}(1-p)}{1-p^K}$, and $\rho_K(p)\to0$ as $K\to\infty$.
  Moreover, $\Pr(\hat P_{\mathrm{idx}}=0)=(1-P)^n$ and $\Pr(\hat P_{\mathrm{all}}=0)=1-\prod_v\big(1-(1-p_v)^n\big)$.
\end{corollary}

\begin{proof}
  The $n$ index combinations use disjoint rollouts, so their indicators are i.i.d.\ Bernoulli$(P)$.
  This gives the mean, Eq.~(\ref{eq:var-idx}) and $\Pr(\hat P_{\mathrm{idx}}=0)=(1-P)^n$.
  The $\hat p_v$ are independent with $\E[\hat p_v]=p_v$ and $\E[\hat p_v^2]=\Var(\hat p_v)+p_v^2=a_v/n+p_v^2$, so $\E[\hat P_{\mathrm{all}}]=P$ and $\E[\hat P_{\mathrm{all}}^2]=\prod_v(p_v^2+a_v/n)$.
  We expand the product over the subsets $T$ of factors that contribute $a_v/n$.
  Removing the term for $T=\emptyset$, which equals $P^2$, gives Eq.~(\ref{eq:var-all}).
  For the inequality, let $\hat P_\Pi=\frac1n\sum_k\prod_vc_{v,\Pi_v(k)}$ for a random pairing $\Pi$ as in Lemma~\ref{lem:random-pairing}.
  By the re-indexing step of Theorem~\ref{thm:rb}, $\hat P_\Pi\overset{d}{=}\hat P_{\mathrm{idx}}$.
  Averaging over $\Pi$ gives $\E[\hat P_\Pi\mid O]=\hat P_{\mathrm{all}}$, so the law of total variance yields $\Var(\hat P_{\mathrm{all}})\le\Var(\hat P_{\mathrm{idx}})$.
  The $|T|=1$ part of Eq.~(\ref{eq:var-all}) is $\frac1n\sum_va_v\prod_{u\neq v}p_u^2=\frac{P^2}{n}\sum_v\frac{1-p_v}{p_v}$, which gives $\rho$.
  Multiplying both variances by $n$ and letting $n\to\infty$ in the exact inequality gives $\rho\le1$.
  For $p_v\equiv p$, we have $\rho_K(p)=\frac{Kp^{K}(1-p)/p}{1-p^K}$.
  Since $1-p^K=(1-p)\sum_{j=0}^{K-1}p^j$, this gives $\rho_K(p)=Kp^{K-1}/\sum_{j=0}^{K-1}p^j\le1$, because $p^j\ge p^{K-1}$ for $j<K$.
  As $K\to\infty$, $Kp^{K-1}\to0$ while the denominator is at least 1.
  Finally, $\hat P_{\mathrm{all}}=0$ if and only if some version has no correct rollout.
  This event has probability $(1-p_v)^n$ for version $v$, independently across versions.
\end{proof}

When the accuracies are below 1, $P$ decays geometrically in $K$, so pairing by index finds no all-correct combination in most sets.
In contrast, the estimate from all combinations vanishes only if some version has no correct rollout.
For $p_v=1/2$, $K=3$ and $n=4$, we have $\Var(\hat P_{\mathrm{idx}})=0.0273$, $\Var(\hat P_{\mathrm{all}})=0.0149$ and $\rho_3(1/2)=3/7$, and the two estimates vanish in 58.6\% and 17.6\% of sets.
For $K=2$ and $n=6$, the two estimates vanish in 17.8\% and 3.1\% of the sets.

\paragraph{(D2) Per-version versus pooled baseline.}
Let $K\ge2$ and $N=Kn$.
A baseline pooled over versions compares $(v,i)$ with $b^{\mathrm{sh}}_{v,i}$, which is the mean reward of the other $N-1$ rollouts of the set.
GRPO on a pooled group uses $r_{v,i}-\frac1N\sum_{u,j}r_{u,j}=\frac{N-1}{N}(r_{v,i}-b^{\mathrm{sh}}_{v,i})$ and divides it by the group standard deviation.

\begin{proposition}[Per-version versus pooled baseline]
  \label{prop:baseline}
  (a)~$r_{v,i}-b^{\mathrm{sh}}_{v,i}=(r_{v,i}-\bar r_{v,-i})+d_{v,i}$, where $d_{v,i}=\bar r_{v,-i}-b^{\mathrm{sh}}_{v,i}$ is independent of $o_{v,i}$, so $\E[\psi_{v,i}d_{v,i}]=0$.

  (b)~With
  \begin{equation*}
    \Delta_v=V_v-\E[b^{\mathrm{sh}}_{v,i}]=\frac{n(K-1)}{N-1}\Big(V_v-\frac1{K-1}\sum_{u\neq v}V_u\Big),
  \end{equation*}
  the mean squared coefficients are
  \begin{align*}
    \E\big[(r_{v,i}-\bar r_{v,-i})^2\big]&=\frac{n}{n-1}\,\sigma_v^2,\\
    \E\big[(r_{v,i}-b^{\mathrm{sh}}_{v,i})^2\big]&=\sigma_v^2+\frac{(n-1)\sigma_v^2+n\sum_{u\neq v}\sigma_u^2}{(N-1)^2}+\Delta_v^2,
  \end{align*}
  so the per-version baseline is smaller if and only if $\Delta_v^2>\frac{\sigma_v^2}{n-1}-\Var(b^{\mathrm{sh}}_{v,i})$.
\end{proposition}

\begin{proof}
  (a)~Neither mean contains $o_{v,i}$, so Corollary~\ref{cor:baseline} applies.
  (b)~For a baseline $b$ independent of $r_{v,i}$,
  \begin{align*}
    \E[(r_{v,i}-b)^2]
    &=\E\big[\big((r_{v,i}-V_v)+(V_v-\E b)+(\E b-b)\big)^2\big]\\
    &=\sigma_v^2+(V_v-\E b)^2+\Var(b),
  \end{align*}
  because the three summands are independent or constant and the first and last have mean 0.
  The leave-one-out mean has $\E b=V_v$ and $\Var(b)=\sigma_v^2/(n-1)$.
  The pooled baseline averages $n-1$ rollouts of version $v$ and $n$ of each other version, so $\E b=\frac{(n-1)V_v+n\sum_{u\neq v}V_u}{N-1}$ and $\Var(b)=\frac{(n-1)\sigma_v^2+n\sum_{u\neq v}\sigma_u^2}{(N-1)^2}$.
  Since $(N-1)-(n-1)=n(K-1)$, we get $V_v-\E b=\frac{n(K-1)V_v-n\sum_{u\neq v}V_u}{N-1}=\Delta_v$.
\end{proof}

In (b), the mean squared coefficient of the per-version baseline exceeds $\sigma_v^2$ by $\Var(\bar r_{v,-i})=\sigma_v^2/(n-1)$, because it averages only $n-1$ rollouts.
That of the pooled baseline exceeds $\sigma_v^2$ by $\Var(b^{\mathrm{sh}}_{v,i})+\Delta_v^2$, and $\Delta_v\neq0$ whenever $V_v\neq\frac1{K-1}\sum_{u\neq v}V_u$.
Standard variance analyses~\cite{greensmith2004variance} assume that the squared norm of the score is uncorrelated with the squared coefficient.
Under this assumption, the variance contributed by a rollout is proportional to its mean squared coefficient, and (b) compares the two estimates.
On contrast sets, the versions can differ strongly in difficulty, for instance when the original is answerable from the prior and a variant is not.
As a result, $\Delta_v$ is often large.
For rewards in $[0,1]$, we have $\sigma_v^2\le\frac14$, and with $n=4$ and $K=3$ the condition $|V_v-\frac12\sum_{u\neq v}V_u|\ge0.40$ suffices.
Part~(a) shows that pooling adds the zero-mean term $\psi_{v,i}d_{v,i}$.
Because its mean is zero, this term does not change the expected gradient.
However, it changes each update, and the next requirement \mbox{examines}~this~change.

\subsection{Requirement 3: credit assignment}
\label{sec:theory-credit}

A step increases $\log P_\theta(o_{v,i}\mid s_v)$ for every $A_{v,i}>0$ and decreases it for every $A_{v,i}<0$, so two estimates with the same expectation can reinforce different behavior in each step, even though they~agree~on~\mbox{average}.

\begin{proposition}[Even treatment of versions]
  \label{prop:balance}
  (a)~For every set-level reward $R$ and every sample, $\sum_iA^R_{v,i}=0$ for each $v$, and $\E[A^R_{v,i}]=0$.

  (b)~With the pooled baseline, $\E[r_{v,i}-b^{\mathrm{sh}}_{v,i}]=\Delta_v$.
  With GRPO on a pooled group, $\sum_iA^{\mathrm{GRPO}}_{v,i}=n(\bar r_v-\mu)/\sigma$, where $\mu$ and $\sigma$ are the mean and standard deviation of the rewards in the group.
\end{proposition}

\begin{proof}
  (a)~The sum vanishes by Lemma~\ref{lem:loo}(b) with $x_i=\hat R_v(o_{v,i})$, and the expectation vanishes because the rollouts of version $v$ are exchangeable.
  (b)~The first claim is part of the proof of Proposition~\ref{prop:baseline}(b), and the second follows by summing $(r_{v,i}-\mu)/\sigma$ over $i$.
\end{proof}

By (a), the cross-input advantages of each version sum to zero, so no version gains or loses credit as a whole.
In contrast, by (b), GRPO on a pooled group gives version $v$ the total $n(\bar r_v-\mu)/\sigma$, which is positive if and only if $\bar r_v>\mu$.
The most important case is a set answered from the prior.

\begin{proposition}[No credit for the prior]
  \label{prop:prior}
  Let $K\ge2$.
  (a)~If at most one version has a correct rollout, the joint term of Eq.~(\ref{eq:advantage}) vanishes for every rollout of the set.
  For $K\le3$, it already vanishes if $\hat p_u=0$ for some $u$.
  (b)~Write $r=c+e$, and let $c_{v^\star,i}=1$ for all $i$ and $c_{v,i}=0$ for all $v\neq v^\star$ and all $i$.
  This is the outcome of answering every version with $y_{v^\star}$ when the answers are pairwise distinct.
  Then $A_{v,i}=e_{v,i}-\bar e_{v,-i}$ for all $(v,i)$, and all advantages vanish if $e$ is constant on the set.
  More generally, $A^R_{v,i}=0$ for all $i$ whenever $\hat R_v(o_{v,i})$ does not depend on $i$.
  (c)~In the situation of (b) with constant $e$, consider GRPO on the pooled group of $N=Kn$ rollouts with the population standard deviation.
  It gives $\sqrt{K-1}$ to every rollout of $v^\star$ and $-1/\sqrt{K-1}$ to every other rollout.
\end{proposition}

\begin{proof}
  (a)~Let $u^\star$ be the only version with a correct rollout, if there is one.
  For $v\neq u^\star$, all rollouts of $v$ are wrong, so $c_{v,i}-\bar c_{v,-i}=0$.
  For $v=u^\star$, every product in $w_v$ ranges over $k-1\ge1$ other versions, which all have $\hat p=0$, so $w_v=0$.
  For $K\le3$, $w_v=\prod_{u'\neq v}\hat p_{u'}$ contains the factor $\hat p_u=0$ for every $v\neq u$, and $c_{u,i}-\bar c_{u,-i}=0$.
  (b)~By (a), the joint term vanishes.
  Since $c$ is constant within each version, $r_{v,i}-\bar r_{v,-i}=e_{v,i}-\bar e_{v,-i}$.
  The general statement follows directly from Eq.~(\ref{eq:general-advantage}).
  (c)~The group has $n$ rewards $1+e$ and $(K-1)n$ rewards $e$, so $\mu=e+\frac1K$ and $\sigma^2=\frac1K\big(1-\frac1K\big)$.
  Then $(1-\frac1K)/\sigma=\sqrt{K-1}$ and $-\frac1K/\sigma=-1/\sqrt{K-1}$.
\end{proof}

\begin{example}
  \label{ex:prior}
  In Example~\ref{ex:running}, let all rollouts answer ``1987'' with the same number of zooms.
  Pooled GRPO gives $+1$ to every rollout on A and $-1$ to every rollout on B.
  The step therefore raises the probability of the sampled guesses on A, together with the reasoning and the absence of zooming that produced them.
  In contrast, CrossRPO gives 0 to every rollout.
  For $K=3$ the pooled advantages are $+1.41$ and $-0.71$.
\end{example}

When the rollouts of $v^\star$ are only partly correct, CrossRPO still credits the correct ones, but only relative to other rollouts of $v^\star$.

\subsection{Requirement 4: scale}
\label{sec:theory-scale}

Besides correctness, a reward may contain small terms, such as a cost for each extra zoom.
The effect of such a term on an update should stay small relative to the effect of correctness.
Choice (D3) decides whether it does.
For a transformation $T$ of rewards, call an advantage map $A(\cdot)$ \emph{invariant} if $A(T(r))=A(r)$ and \emph{equivariant} if $A(T(r))=aA(r)$.

\begin{proposition}[Scale]
  \label{prop:scale}
  (a)~The GRPO advantage is invariant under every \emph{per-group} positive affine map $r_{g,i}\mapsto\alpha_gr_{g,i}+\beta_g$ with $\alpha_g>0$.
  Moreover, every group whose rewards are not all equal satisfies $\frac1{N_g}\sum_i(A^{\mathrm{GRPO}}_{g,i})^2=1$.

  (b)~The cross-input advantage is linear in the set-level reward: $A^{\alpha R+\beta}=\alpha A^R$ for constants $\alpha,\beta$, and $A^{R_1+R_2}=A^{R_1}+A^{R_2}$.
  After division by the batch scale, it is invariant under $R\mapsto\alpha R+\beta$ with one $\alpha>0$ for the whole batch.
  In general, it is not invariant under a different $\alpha$ for each set.

  (c)~Let all $N$ rollouts of a group be correct, and let one of them pay an extra cost $\epsilon>0$.
  For every $\epsilon>0$, its GRPO advantage is $-\sqrt{N-1}$ and the others get $1/\sqrt{N-1}$.
  For $N=12$, this is $-3.32$, which is below the advantage $-1$ of a wrong answer in a group with half of the answers correct.

  (d)~In the situation of (c) on a version with $n$ rollouts, CrossRPO gives $-\epsilon$ to the costly rollout and $\epsilon/(n-1)$ to the others before scaling.
  The batch scale preserves the ratio between any two advantages in the batch.
\end{proposition}

\begin{proof}
  (a)~The mean of group $g$ becomes $\alpha_g\mu_g+\beta_g$ and its standard deviation $\alpha_g\sigma_g$, so $(r-\mu)/\sigma$ is unchanged.
  The second claim follows from the definition of $\sigma_g^2$.
  (b)~$\hat R_v$ in Eq.~(\ref{eq:empirical-projection}) is linear in $R$, the constant $\beta$ cancels in the leave-one-out difference, and the batch scale divides all advantages by one number.
  (c)~The deviations from the mean are $\epsilon/N$ for $N-1$ rollouts and $-\epsilon(N-1)/N$ for one, so $\sigma^2=\frac1N\big[(N-1)\frac{\epsilon^2}{N^2}+\frac{\epsilon^2(N-1)^2}{N^2}\big]=\frac{\epsilon^2(N-1)}{N^2}$ and $\sigma=\epsilon\sqrt{N-1}/N$.
  Dividing by $\sigma$ gives the two values.
  With half of the answers correct and equal costs, $\mu=\sigma=\frac12$ and the advantages are $\pm1$.
  (d)~All rollouts are correct, so the joint term vanishes and the per-version term gives $-\epsilon$ and $\frac{\epsilon}{n-1}$.
  The scale divides every advantage by $s$.
\end{proof}

With the sample standard deviation, the values in (c) are multiplied by $\sqrt{(N-1)/N}$, which gives $-3.18$ for $N=12$.
By (a), the GRPO advantage does not depend on the scale $\alpha_g$ of the rewards in a group.
By (c), a cost $\epsilon>0$ of any size therefore gives the advantage $-\sqrt{N-1}$, which is $-3.32<-1$ for $N=12$, so an extra zoom is penalized more than a wrong answer.
Linearity (b) keeps advantages proportional to reward differences across the whole batch, and the batch scale only changes the step length (Remark~\ref{rem:unbiased}).
In contrast, per-group scales reweight the sets against each other.

\subsection{Generalizations and special cases}
\label{sec:theory-general}

\paragraph{Correctness-based set rewards.}
The construction of Appendix~\ref{sec:theory-advantage} applies to any $R$.
When $R$ depends on the rollouts only through their correctness, it has a closed form.
For $\Phi:\{0,1\}^K\to\mathbb R$, let
\begin{equation}
  \tilde\Phi(x)=\sum_{z\in\{0,1\}^K}\Phi(z)\prod_{v}x_v^{z_v}(1-x_v)^{1-z_v},\qquad x\in[0,1]^K,
  \label{eq:multilinear}
\end{equation}
be its multilinear extension, and write $(x_{-v},t)$ for $x$ with the $v$-th coordinate replaced by $t$.

\begin{proposition}[Multilinear form]
  \label{prop:multilinear}
  Let $R_\Phi(o_{1:K})=\Phi(c(o_1),\dots,c(o_K))$ and $p=(p_v)_v$.
  (a)~$J_{R_\Phi}=\tilde\Phi(p)$ and $\bar R_v(o)=\tilde\Phi(p_{-v},c(o))$.
  (b)~$\nabla_\theta J_{R_\Phi}=\sum_v\partial_v\tilde\Phi(p)\,\nabla_\theta p_v$ with $\partial_v\tilde\Phi(p)=\tilde\Phi(p_{-v},1)-\tilde\Phi(p_{-v},0)$.
  (c)~$\hat R_v(o_{v,i})=\tilde\Phi(\hat p_{-v},c_{v,i})$ and $A^{R_\Phi}_{v,i}=w^\Phi_v\,(c_{v,i}-\bar c_{v,-i})$ with $w^\Phi_v=\tilde\Phi(\hat p_{-v},1)-\tilde\Phi(\hat p_{-v},0)$.
\end{proposition}

\begin{proof}
  (a)~The $c(o_v)$ are independent Bernoulli$(p_v)$, so $\Pr\big((c(o_v))_v=z\big)=\prod_vp_v^{z_v}(1-p_v)^{1-z_v}$.
  Both claims follow by taking the expectation of $\Phi$ without conditioning and given $c(o_v)$.
  (b)~The function $\tilde\Phi$ is affine in each coordinate, with $\tilde\Phi(x)=x_v\tilde\Phi(x_{-v},1)+(1-x_v)\tilde\Phi(x_{-v},0)$, which gives $\partial_v\tilde\Phi$.
  The chain rule applied to $J=\tilde\Phi(p(\theta))$ gives the gradient.
  (c)~In Eq.~(\ref{eq:empirical-projection}), the tuple $(c_{u,\iota_u})_{u\neq v}$ over uniform $\iota$ has independent coordinates with $\Pr(c_{u,\iota_u}=1)=\hat p_u$, so the average equals the expectation in (a) with $p_{-v}$ replaced by $\hat p_{-v}$.
  The affine form of (b) gives $\hat R_v(o_{v,i})=\tilde\Phi(\hat p_{-v},0)+w^\Phi_vc_{v,i}$, and the first term does not depend on $i$, so it cancels in the leave-one-out difference.
\end{proof}

The weight $w^\Phi_v$ is the estimated marginal value of being correct on version $v$.
Natural choices are the threshold rewards $\Phi_m(z)=\mathbf 1[\sum_vz_v\ge m]$, for which $w^{\Phi_m}_v$ is the probability that exactly $m-1$ of the other versions are correct under independent Bernoulli$(\hat p_u)$:
\begin{itemize}
  \item $m=K$ gives $\Phi_K(z)=\prod_vz_v$ and $w_v=\prod_{u\neq v}\hat p_u$, which is the joint term of CrossRPO;
  \item $m=1$ gives ``at least one version correct'', with $w_v=\prod_{u\neq v}(1-\hat p_u)$;
  \item $\Phi(z)=\sum_vz_v$ gives $w_v\equiv1$, which is independent per-version accuracy.
\end{itemize}

\begin{corollary}[Choice of the threshold]
  \label{cor:threshold}
  On a set with pairwise distinct answers, a blind policy has $J_{R_{\Phi_m}}=\Pr(N\ge m)\le1/m$, with the sharper bound $K^{-K}$ for $m=K$.
  For $m=1$ the bound is attained by a deterministic blind policy that always gives one of the $y_v$.
\end{corollary}

\begin{proof}
  The bounds follow from Proposition~\ref{prop:blind}(b,c).
  A policy that always answers $y_1$ has $N=1$ almost surely.
\end{proof}

So ``at least one correct'' rewards guessing, whereas the strictest threshold $m=K$ bounds blind policies most tightly.
However, the weight of $\Phi_K$ is a product of $K-1$ accuracies, so it vanishes for large $K$.
The normalized counts of all-correct $k$-subsets, $\Phi^{(k)}(z)=\binom Nk\big/\binom{K-1}{k-1}$ with $N=\sum_vz_v$, lie between the two.
By Proposition~\ref{prop:multilinear}, their weight is $w^{(k)}_v=e_{k-1}(\hat p_{-v})/\binom{K-1}{k-1}$, which is the fraction of all-correct combinations of rollouts on $k-1$ other versions drawn at random.
The order $k=1$ gives $w\equiv1$, which is per-version accuracy, $k=2$ gives the pairs of Corollary~\ref{cor:pairwise}, and $k=K$ gives $\Phi_K$.

\begin{proposition}[Order of the joint term]
  \label{prop:order}
  Let $1\le k\le K$.
  (a)~If the correct answers are pairwise distinct, a blind policy attains at most a fraction $K^{-k}$ of the maximum $K/k$ of $J_{R_{\Phi^{(k)}}}$.
  (b)~If all other versions have accuracy $p$, then $\partial_v\tilde\Phi^{(k)}(p)=p^{k-1}$ for every $K\ge k$.
  In contrast, if all other versions have accuracy at most $\bar p<1$, the weight of the set accuracy is $\prod_{u\neq v}p_u\le\bar p^{\,K-1}$.
  (c)~For $k\ge3$, no sum of per-version and pairwise terms reproduces $\Phi^{(k)}$.
\end{proposition}

\begin{proof}
  (a)~By Proposition~\ref{prop:multilinear}(a), $J_{R_{\Phi^{(k)}}}=e_k(p)/\binom{K-1}{k-1}$.
  By Proposition~\ref{prop:blind}(e), this is at most $\binom KkK^{-k}/\binom{K-1}{k-1}=K^{-k}\cdot K/k$, and by Proposition~\ref{prop:agreement}(a), its maximum is $K/k$.
  (b)~$e_{k-1}$ of $K-1$ entries equal to $p$ is $\binom{K-1}{k-1}p^{k-1}$, and each factor of the product is at most $\bar p$.
  (c)~$\binom Nk=\sum_{|T|=k}m_T$ has nonzero coefficients on monomials of degree $k\ge3$, so the argument of Proposition~\ref{prop:not-pairwise}(a) applies to it as well.
\end{proof}

The order therefore trades two properties.
A larger $k$ bounds blind policies more tightly, but the weight that carries the joint credit decays like $p^{k-1}$.
CrossRPO uses $k=\min(K,3)$ in Eq.~(\ref{eq:set-reward}).
This is the smallest order that no pairwise reward reproduces, and its weight is of order $p^2$ for every $K$.
For $K\le3$, it is the set accuracy $\Phi_K$.
Prior agreement rewards depend on the answers rather than on correctness and fall outside this family (Proposition~\ref{prop:agreement}).
Two further extensions hold verbatim.
\emph{Soft correctness}: for scores $c\in[0,1]$ such as a judge score, Proposition~\ref{prop:multilinear} holds with $\Phi$ replaced by $\tilde\Phi$ applied to the scores.
The reason is that the expectation of a multilinear function of independent variables is the function of their means.
\emph{Unequal sample sizes and set weights}: Remark~\ref{rem:unbiased}(i).

\paragraph{Pairwise rewards.}
Pair-based methods credit two versions at a time.
Applied to every pair of a set, they use a set-level reward of the form
\begin{equation}
  R_{\mathrm{pair}}(o_{1:K})=\sum_{u<v}R_{uv}(o_u,o_v),
  \label{eq:pair-reward}
\end{equation}
for example $R_{uv}=\frac{r(o_u)+r(o_v)}{K-1}+\lambda\,c(o_u)c(o_v)$, which gives $\sum_vr(o_v)+\lambda\binom{N}{2}$.
Two questions arise.
The first is whether the advantage of such a reward is the sum of the pairwise advantages, and the second is whether some choice of $R_{uv}$ reproduces the joint term of Eq.~(\ref{eq:set-reward}).
The first holds and the second fails for $K\ge3$.

\begin{lemma}[Additivity]
  \label{lem:additivity}
  (a)~$A^{\alpha R_1+\beta R_2}=\alpha A^{R_1}+\beta A^{R_2}$ for all set-level rewards $R_1,R_2$ and constants $\alpha,\beta$.
  (b)~Let $R_T$ depend only on $(o_u)_{u\in T}$ for some $T\subseteq\mathcal V$.
  Then $A^{R_T}_{v,i}=0$ for $v\notin T$.
  For $v\in T$, $A^{R_T}_{v,i}$ equals the cross-input advantage of $R_T$ computed with the same rollouts on the set restricted to the versions in $T$.
  (c)~Consequently, for Eq.~(\ref{eq:pair-reward}), $A^{R_{\mathrm{pair}}}_{v,i}=\sum_{u\neq v}A^{R_{uv}}_{v,i}$, where each term is computed on the pair $\{u,v\}$ alone.
\end{lemma}

\begin{proof}
  (a)~$\hat R_v$ in Eq.~(\ref{eq:empirical-projection}) is linear in $R$, and so is the leave-one-out difference in Eq.~(\ref{eq:general-advantage}).
  (b)~If $v\notin T$, the summands of Eq.~(\ref{eq:empirical-projection}) do not depend on $o_{v,i}$, so $\hat R_v(o_{v,i})$ is the same for all $i$ and the leave-one-out difference vanishes.
  If $v\in T$, each summand depends on $\iota$ only through $(\iota_u)_{u\in T\setminus\{v\}}$, and every value of these coordinates occurs $n^{K-|T|}$ times.
  The average over $[n]^{\mathcal V\setminus\{v\}}$ therefore equals the average over $[n]^{T\setminus\{v\}}$, which is the empirical projection on the restricted set.
  (c)~Apply (a) to the sum and (b) to each $R_{uv}$.
\end{proof}

\begin{proposition}[The joint term is not pairwise]
  \label{prop:not-pairwise}
  Let $K\ge3$, so that $k=3$, and let $N=\sum_vz_v$.
  (a)~There are no functions $\phi_v:\{0,1\}\to\mathbb R$ and $\phi_{uv}:\{0,1\}^2\to\mathbb R$ such that $\binom N3=\sum_v\phi_v(z_v)+\sum_{u<v}\phi_{uv}(z_u,z_v)$ for all $z\in\{0,1\}^K$.
  The same holds for the set accuracy $\prod_vz_v$.
  (b)~More generally, let $0<p_v<1$ for all $v$.
  For any bounded $r_v:\mathcal O\to\mathbb R$ and $R_{uv}:\mathcal O^2\to\mathbb R$, the reward $\sum_vr_v(o_v)+\sum_{u<v}R_{uv}(o_u,o_v)$ differs from $\lambda\binom{N}{3}$ and from $\lambda\prod_vc(o_v)$ with positive probability under $\bigotimes_vP_v$ for every $\lambda\neq0$, where $N=\sum_vc(o_v)$.
\end{proposition}

\begin{proof}
  (a)~The monomials $m_T(z)=\prod_{v\in T}z_v$ for $T\subseteq\mathcal V$ form a basis of the functions on $\{0,1\}^K$.
  There are $2^K$ of them, and if $f=\sum_T\beta_Tm_T$, then $f(\mathbf 1_S)=\sum_{T\subseteq S}\beta_T$ for the indicator vector $\mathbf 1_S$ of $S\subseteq\mathcal V$.
  Möbius inversion therefore gives $\beta_T=\sum_{S\subseteq T}(-1)^{|T\setminus S|}f(\mathbf 1_S)$ uniquely.
  Since $\phi_v(z_v)=\phi_v(0)+(\phi_v(1)-\phi_v(0))z_v$ and $\phi_{uv}(z_u,z_v)$ is a combination of $1,z_u,z_v,z_uz_v$, the right-hand side lies in the span of $\{m_T:|T|\le2\}$.
  In contrast, $\binom N3=\sum_{|T|=3}m_T$ and $\prod_vz_v=m_{\mathcal V}$ with $|\mathcal V|=K\ge3$ have nonzero coefficients on monomials of degree at least 3.
  (b)~Under the product measure, every square-integrable $f(o_{1:K})$ has a unique decomposition $f=\sum_{T\subseteq\mathcal V}f_T$ into mutually orthogonal components, where $f_T$ depends only on $(o_u)_{u\in T}$ and has zero conditional mean given any $|T|-1$ of these coordinates~\cite{hoeffding1948class}.
  A function of at most two coordinates has $f_T=0$ for $|T|\ge3$, so the reward in (b) has no component of order 3 or more.
  Writing $c(o_v)=p_v+(c(o_v)-p_v)$ and expanding the products shows two things.
  The order-$K$ component of $\lambda\prod_vc(o_v)$ is $\lambda\prod_v(c(o_v)-p_v)$, whose variance $\lambda^2\prod_vp_v(1-p_v)$ is positive.
  For each $T$ with $|T|=3$, the component of $\lambda\binom N3=\lambda\sum_{|T|=3}\prod_{u\in T}c(o_u)$ for $T$ is $\lambda\prod_{u\in T}(c(o_u)-p_u)$, whose variance $\lambda^2\prod_{u\in T}p_u(1-p_u)$ is positive.
  In both cases, the difference of the two rewards has a nonzero component and is not almost surely zero.
\end{proof}

\begin{corollary}[Pairwise versus all-correct]
  \label{cor:pairwise}
  Let the correct answers of a set be pairwise distinct and $K\ge2$, and let $\binom N2=\sum_{u<v}c(o_u)c(o_v)$ be the number of correctly answered pairs.
  (a)~A blind policy has $\E\big[\binom N2\big]\le\frac1{K^2}\binom K2$, which is a fraction $K^{-2}$ of the value of a policy that answers every version.
  In contrast, the joint term of CrossRPO is a fraction at most $K^{-k}$ (Proposition~\ref{prop:blind}(e)), which is $K^{-3}<K^{-2}$ for $K\ge3$.
  (b)~The cross-input weight of $\binom N2$ is $w^{(2)}_v=\sum_{u\neq v}\hat p_u$, which stays positive when another version is never answered correctly, whereas for $K=3$ the weight $w_v=\prod_{u\neq v}\hat p_u$ vanishes.
\end{corollary}

\begin{proof}
  (a)~By independence, $\E\big[\binom N2\big]=\sum_{u<v}p_up_v=\frac12\big((\sum_vp_v)^2-\sum_vp_v^2\big)$.
  With $\sum_vp_v^2\ge\frac1K(\sum_vp_v)^2$ and $\sum_vp_v\le1$ (Proposition~\ref{prop:blind}(a)), this is at most $\frac{K-1}{2K}=\frac1{K^2}\binom K2$.
  For $K\ge3$, $k=3$ and $K^{-3}<K^{-2}$.
  (b)~Apply Proposition~\ref{prop:multilinear} with $\tilde\Phi(x)=\sum_{u<v}x_ux_v$, whose partial derivative in $x_v$ is $\sum_{u\neq v}x_u$.
\end{proof}

Applying a pair-based method to every pair is therefore the special case $R=R_{\mathrm{pair}}$ of the framework, whose advantage is the sum of the pairwise advantages (Lemma~\ref{lem:additivity}).
For $K=2$, it coincides with CrossRPO.
In contrast, for $K\ge3$, the joint term of CrossRPO is an interaction of order three, which no sum of pairwise terms reproduces (Proposition~\ref{prop:not-pairwise}).
Corollary~\ref{cor:pairwise} shows that it bounds blind policies more tightly, and for $K=3$ it gives no joint credit while some version is never~\mbox{answered}~\mbox{correctly}.

\paragraph{Computation.}
CrossRPO uses the same rollouts and the same policy loss as GRPO and changes only the advantages, which require the version and the correctness of each rollout.
For $K\le3$, each $w_v$ is a product of at most two numbers.
For $K>3$, $w_v=\big((S-\hat p_v)^2-(Q-\hat p_v^2)\big)/\big((K-1)(K-2)\big)$ with $S=\sum_u\hat p_u$ and $Q=\sum_u\hat p_u^2$.
In both cases, all $w_v$ cost $O(K)$ and all advantages of a set cost $O(N)$.
In contrast, listing the $n^K$ combinations costs $O(Kn^K)$.
For a symmetric $\Phi$, $w^\Phi_v$ is a Poisson--binomial probability, which dynamic programming computes in $O(K^2)$ per version.
Sets with different $K$ share one batch and one formula.
For $K=1$, the empty product gives $w=1$, so the joint term repeats the accuracy term and only weights single inputs by $1+\lambda$ relative to the versions of sets.
Our experiments instead set $w=0$ for $K=1$, which trains single inputs with RLOO (Proposition~\ref{prop:special}(b)).

\paragraph{Special cases.}
CrossRPO contains the group-based methods of Definition~\ref{def:group}.

\begin{proposition}[Special cases]
  \label{prop:special}
  (a)~For $K=1$, $J=\E[r+\lambda c]$ and $A_i=(r_i+\lambda c_i)-\frac1{n-1}\sum_{j\neq i}(r_j+\lambda c_j)$, the RLOO advantage for the reward $r+\lambda c$.
  (b)~For $\lambda=0$, $A_{v,i}=r_{v,i}-\bar r_{v,-i}$: RLOO on each version separately.
  (c)~Replace the batch scale by the standard deviation of each version's rewards and multiply by $\frac{n-1}{n}$.
  This turns (a) into GRPO with the reward $r+\lambda c$ and (b) into GRPO with one group per version.
  Without the standard deviation, (a) is $\frac{n}{n-1}$ times Dr.~GRPO.
  (d)~For $K\ge2$, GRPO on a group that pools the versions is not of the form Eq.~(\ref{eq:advantage}) for any $\lambda$ and any positive scale per set or per batch.
\end{proposition}

\begin{proof}
  (a)~With $w=1$, the two terms of Eq.~(\ref{eq:advantage}) combine, and Eq.~(\ref{eq:objective}) has one version.
  (b)~This follows directly from Eq.~(\ref{eq:advantage}) with $\lambda=0$.
  (c)~Lemma~\ref{lem:loo}(b).
  (d)~In the situation of Proposition~\ref{prop:prior}(b) with constant $e$, pooled GRPO gives the nonzero advantages of Proposition~\ref{prop:prior}(c), while every CrossRPO advantage is zero for every $\lambda$ and stays zero after any scaling of the batch.
\end{proof}

CrossRPO therefore reduces to GRPO only after reinstating the per-group division by the standard deviation, and then only on single inputs or with $\lambda=0$ and one group per version, as in (b).

\subsection{Which inputs to group}
\label{sec:theory-grouping}

The joint term needs sets, and so far the versions of a set were photos that share a question and differ in the evidence.
This subsection shows that an arbitrary grouping, such as three training questions drawn at random, does not serve, and it identifies what the members of a set must share.

An \emph{item} $x=(s,q,y)$ is an input with its own question and correct answer.
A rollout on $x$ is generated from $(s,q)$, and $p(x)$ and $V(x)$ denote its accuracy and expected reward as in Eq.~(\ref{eq:version-stats}).
Nothing in Appendix~\ref{sec:theory-advantage} uses that the versions of a set share their question.
Eqs.~(\ref{eq:objective}) and~(\ref{eq:advantage}) therefore apply to any $K$-tuple of items $(x_1,\dots,x_K)$ with $p_v=p(x_v)$ and $V_v=V(x_v)$, where each rollout is checked against its own answer $y_v$.

\begin{definition}[Grouping]
  \label{def:grouping}
  Let $\mathcal D$ be a finite distribution of items, such as the uniform distribution on a training set.
  A \emph{grouping} is a distribution $\mathcal G$ of $K$-tuples of items whose marginal at every position is $\mathcal D$.
  CrossRPO maximizes $\E_{\mathcal G}[J]$ with $J$ from Eq.~(\ref{eq:objective}).
  The grouping is \emph{independent} if $\mathcal G=\mathcal D^{\otimes K}$, which means that the members of a set are drawn independently.
  We write $\bar p=\E_{\mathcal D}[p(x)]$ and $\bar V=\E_{\mathcal D}[V(x)]$.
\end{definition}

Contrast sets are the opposite extreme, because their members are versions of one photo and each member determines the others.

\paragraph{Independent members.}
If the members are unrelated, the probability that all of them are answered correctly factorizes in expectation, and the joint term then adds no information.

\begin{proposition}[Independent grouping]
  \label{prop:independent}
  Let the grouping be independent and let Assumption~\ref{asm:sampling} hold given the items.
  (a)~$\E_{\mathcal G}[J]=K\bar V+\lambda\frac Kk\bar p^{\,k}$.
  For $r=c$, this is $\varphi(\bar p)$ with $\varphi(t)=Kt+\lambda\frac Kkt^k$.
  Since $\varphi$ is strictly increasing on $[0,1]$, $\E_{\mathcal G}[J]$ ranks policies exactly as mean accuracy does.

  (b)~$\nabla_\theta\E_{\mathcal G}[J]=K\big(\nabla_\theta\bar V+\lambda\bar p^{\,k-1}\nabla_\theta\bar p\big)$.
  For $r=c$, this is $K\big(1+\lambda\bar p^{\,k-1}\big)\nabla_\theta\bar p$, which is a positive multiple of the gradient of the mean accuracy.

  (c)~Let $\mathcal F_v$ be generated by the item $x_v$ and its rollouts $O_v$.
  For every rollout,
  \begin{equation}
    \E\big[A_{v,i}\mid\mathcal F_v\big]=\big(r_{v,i}-\bar r_{v,-i}\big)+\lambda\bar p^{\,k-1}\big(c_{v,i}-\bar c_{v,-i}\big),
    \label{eq:independent-advantage}
  \end{equation}
  This is the RLOO advantage on the single item $x_v$ for the reward $r+\lambda\bar p^{\,k-1}c$, with the same weight for every item.

  (d)~Let $g_v=\sum_iA_{v,i}\psi_{v,i}$ be the contribution of member $v$ to Eq.~(\ref{eq:ghat}), and let $g^\circ_v$ be the same sum with $w_v$ replaced by $\bar p^{\,k-1}$.
  Then $\E[g_v]=\E[g^\circ_v]$ and
  \begin{equation}
    \Cov(g_v)=\Cov(g^\circ_v)+\lambda^2\Var(w_v)\,\E\big[d_vd_v^\top\big]\succeq\Cov(g^\circ_v),
    \label{eq:independent-variance}
  \end{equation}
  where $d_v=\sum_i(c_{v,i}-\bar c_{v,-i})\psi_{v,i}$ and $m_2=\E_{\mathcal D}\big[p(x)^2+p(x)(1-p(x))/n\big]$.
  Moreover, $\Var(w_v)=m_2^{K-1}-\bar p^{\,2(K-1)}$ for $K\le3$, and $\Var(w_v)=\big(2(K-3)\,\bar p^{\,2}(m_2-\bar p^{\,2})+m_2^2-\bar p^{\,4}\big)\big/\binom{K-1}2$ for $K>3$.
\end{proposition}

\begin{proof}
  (a)~By linearity $\E_{\mathcal G}\big[\sum_vV(x_v)\big]=K\bar V$.
  By independence of the members, each of the $\binom Kk$ monomials of $e_k$ has expectation $\bar p^{\,k}$, and $\binom Kk/\binom{K-1}{k-1}=K/k$.
  For $r=c$, $V=p$ and $\varphi'(t)=K(1+\lambda t^{k-1})>0$.

  (b)~$\mathcal D$ is finite, so the gradient passes through $\E_{\mathcal D}$, and $\nabla_\theta\frac Kk\bar p^{\,k}=K\bar p^{\,k-1}\nabla_\theta\bar p$.

  (c)~In Eq.~(\ref{eq:advantage}) the two differences are functions of $O_v$, and $w_v=e_{k-1}(\hat p_{-v})/\binom{K-1}{k-1}$ is a function of $(x_u,O_u)_{u\neq v}$, which is independent of $\mathcal F_v$.
  Each monomial of $e_{k-1}(\hat p_{-v})$ has independent factors, and $\E[\hat p_u]=\E_{\mathcal D}\big[\E[\hat p_u\mid x_u]\big]=\E_{\mathcal D}[p(x_u)]=\bar p$, so $\E[w_v\mid\mathcal F_v]=\E[w_v]=\bar p^{\,k-1}$.

  (d)~The scores $\psi_{v,i}$ are functions of $\mathcal F_v$, so (c) gives $\E[g_v\mid\mathcal F_v]=g^\circ_v$, and the first claim follows by taking expectations.
  Given $\mathcal F_v$, $g_v-g^\circ_v=\lambda\big(w_v-\bar p^{\,k-1}\big)d_v$ with $d_v$ fixed and $w_v$ independent of $\mathcal F_v$, so $\Cov(g_v\mid\mathcal F_v)=\lambda^2\Var(w_v)\,d_vd_v^\top$.
  By the law of total covariance, $\Cov(g_v)=\Cov(\E[g_v\mid\mathcal F_v])+\E[\Cov(g_v\mid\mathcal F_v)]$, and this gives Eq.~(\ref{eq:independent-variance}).
  Finally, the $\hat p_u$ with $u\neq v$ are i.i.d.\ with mean $\bar p$.
  Given $x_u$, $n\hat p_u$ is binomial, so $\E[\hat p_u^2\mid x_u]=p(x_u)^2+p(x_u)(1-p(x_u))/n$ and $\E[\hat p_u^2]=m_2$.
  For $K\le3$, $w_v=\prod_{u\neq v}\hat p_u$ and $\E[w_v^2]=m_2^{K-1}$.
  For $K>3$, $w_v$ is the U-statistic of order two with kernel $h(a,b)=ab$ on the $M=K-1$ values $\hat p_u$.
  Its variance is $\big(2(M-2)\zeta_1+\zeta_2\big)/\binom M2$ with $\zeta_1=\Var(\bar p\,\hat p_u)=\bar p^{\,2}(m_2-\bar p^{\,2})$ and $\zeta_2=\Var(\hat p_u\hat p_{u'})=m_2^2-\bar p^{\,4}$~\cite{hoeffding1948class}.
\end{proof}

By (b), with independent members and $r=c$, the gradient of the expected objective is $K\big(1+\lambda\bar p^{\,k-1}\big)\nabla_\theta\bar p$.
The joint term therefore multiplies the gradient of every item by the same factor $1+\lambda\bar p^{\,k-1}$, which depends on $\theta$ only through $\bar p$ and increases with it.
In addition, the random partners add the covariance $\lambda^2\Var(w_v)\E[d_vd_v^\top]$ of Eq.~(\ref{eq:independent-variance}).
Here $\Var(w_v)>0$ for $K\ge2$ unless all items have the same accuracy and $\hat p$ has no sampling noise.
The joint term also does not penalize guessing.
By (a), the objective depends on the policy only through $\bar p$ and $\bar V$, so a policy that guesses an item correctly with probability $p(x)$ is valued exactly as a policy that reads it with the same accuracy.
In Example~\ref{ex:running}, pair variant A with an unrelated item $x'$, which is a different photo with its own question.
The guess ``1987'' and reading the sign both have accuracy 1 on A, so the pair has the same set accuracy $p(x')$ under both policies.
In contrast, on the contrast set $(\mathrm A,\mathrm B)$, the guess has set accuracy 0 and reading has 1 (Proposition~\ref{prop:blind}(d)).

\paragraph{Shared shortcuts.}
A contrast set adds a constraint.
A policy that does not inspect the region in which the versions differ must answer all versions alike.
The same argument applies to every shared feature that can give the answer.

\begin{definition}[Shortcut]
  \label{def:shortcut}
  A \emph{shortcut} is a map $z$ on items, for example the image outside $\Omega$, the question, or both.
  An answer rule \emph{uses only} $z$ if its distribution of answers on an item $x$ is $\rho_{z(x)}$ for some family $(\rho_\zeta)_\zeta$ of distributions indexed by the values of $z$.
\end{definition}

\begin{proposition}[What the joint term rules out]
  \label{prop:shortcut}
  Let the correct answers $y_1,\dots,y_K$ of a set be pairwise distinct.
  (a)~Suppose that $z(x_1)=\dots=z(x_K)$, so the members share the shortcut.
  Then every answer rule that uses only $z$ satisfies (a)--(d) of Proposition~\ref{prop:blind}.
  In particular, $\sum_vp_v\le1$ and $\prod_vp_v\le K^{-K}$, and the joint term is at most a fraction $K^{-k}$ of its maximum.
  In contrast, reading attains $\prod_vp_v=1$ and the maximum of the joint term.

  (b)~Suppose that $z(x_u)\neq z(x_v)$ for $u\neq v$, so $z$ separates the members.
  Then for every $a\in[0,1]^K$, there is an answer rule that uses only $z$ with $p_v=a_v$ for all $v$.
  In particular, such a rule can attain $\prod_vp_v=1$.
\end{proposition}

\begin{proof}
  (a)~The proof of Proposition~\ref{prop:blind} uses only that the distribution of answers is the same on every version, and here it is $\rho_\zeta$ with $\zeta=z(x_1)$.
  (b)~Let $\rho_{z(x_v)}$ put mass $a_v$ on $y_v$ and $1-a_v$ on giving no answer.
  These choices do not conflict, because the values $z(x_v)$ are distinct.
\end{proof}

Table~\ref{tab:grouping} summarizes the consequences.
The requirement on a set is therefore not that its members come from one image, but that they share the shortcut to be removed and have pairwise different correct answers.
Contrast sets meet it for the shortcut that matters most for a zoom agent, which is answering from the scene without inspecting the region.
Sets of different photos with the same question meet it for another shortcut, which is answering from the question alone.
Unrelated items meet it for no shortcut.
In expectation, CrossRPO then behaves as RLOO on single items with a slightly larger and noisier accuracy reward, so its benefit over per-item training should appear only on data in which the sets share a shortcut.

\begin{table*}[htbp]
\caption{Which inputs to group (Appendix~\ref{sec:theory-grouping}).
The joint term separates reading from a shortcut only on sets whose members share the shortcut and have pairwise different answers.}
\label{tab:grouping}
\centering
\scriptsize
\begin{tabular*}{\textwidth}{@{\extracolsep{\fill}}>{\raggedright\arraybackslash}m{3.7cm}>{\raggedright\arraybackslash}m{3.0cm}>{\raggedright\arraybackslash}m{4.6cm}>{\raggedright\arraybackslash}m{4.3cm}@{}}
\toprule
\midrule
Members of a set & Shared by the members & Shortcut whose set accuracy is bounded & Effect of the joint term \\
\midrule
Versions of one photo that differ in the evidence (contrast sets) & scene outside $\Omega$, question & answering from the scene and common sense without inspecting the region & bounds the shortcut by a fraction $K^{-k}$ (Proposition~\ref{prop:shortcut}(a)) \\
\midrule[0.3pt]
Different photos with the same question and different answers~\cite{goyal2017making} & question & answering from the question alone & bounds the shortcut by a fraction $K^{-k}$ (Proposition~\ref{prop:shortcut}(a)) \\
\midrule[0.3pt]
Transformed copies with the same answer & everything but the transform & none: the answers are not distinct & rewards answering every copy correctly, but a shortcut can do so \\
\midrule[0.3pt]
Independent items & nothing & none (Proposition~\ref{prop:shortcut}(b)) & in expectation a constant reweighting of RLOO, plus variance (Proposition~\ref{prop:independent}) \\
\midrule
\bottomrule
\end{tabular*}
\end{table*}

\paragraph{Empirical check.}
Our data meet both conditions of Proposition~\ref{prop:shortcut}(a).
The versions of a set share the scene and the question.
Moreover, any two versions have different correct answers in every multi-version set of the training data (1,664 sets with $K=2$ or $3$, 3,638 pairs of versions) and of ViRA-Bench (762 sets, 2,286 pairs).
The training data also contain 1,978 single images, so one batch mixes sets with $K=1$, 2 and 3.
The ablation with $\lambda=0$ removes the joint term (Appendix~\ref{sec:supp-ablation-settings}).
Every image then forms its own group (Proposition~\ref{prop:special}(b)).
By Propositions~\ref{prop:independent} and~\ref{prop:shortcut}, the joint term affects only answers that use the shortcut shared by the versions of a set.
Set accuracy reveals such answers, whereas single-image benchmarks cannot.
The measurements agree (Table~\ref{tab:lambda0}).
On ViRA-Bench, set accuracy with $\lambda=0.5$ is higher at every budget, by 4.6 to 7.4 points.
An agent that answers from the scene is more accurate on the original than on its variants.
This gap shrinks from 14.3 to 12.5 points at 0.5\,MP (O / A / B: 69.2 / 56.9 / 52.9 with $\lambda=0$, 73.4 / 62.4 / 59.5 with $\lambda=0.5$) and from 12.4 to 9.5 points at 1\,MP.
On the 78 photographed sets of ViRA-Bench-Real, set accuracy is higher at every budget, by 0.9 to 7.7 points.
On V$^*$ and HR-Bench, the two models are within 1.9 points.
As Proposition~\ref{prop:independent} predicts, a control that computes $w_v$ in the same advantage from images of other questions drawn at random behaves like $\lambda=0$ on ViRA-Bench (33.8 / 42.1 / 47.5 / 51.7 / 57.3 against 31.3 / 41.2 / 50.0 / 52.9 / 58.7, Table~\ref{tab:ablation}).
Every number is the mean of three evaluation runs.

\begin{table}[htbp]
\caption{Effect of the joint term.
CrossRPO ($\lambda=0.5$, ViRA-9B) is compared with one group per image ($\lambda=0$), and both are trained from Qwen3.5-9B for 227 steps.
Top: set accuracy (\%) on ViRA-Bench (762 sets) and ViRA-Bench-Real (78 photographed sets).
Bottom: accuracy (\%) on the general benchmarks V$^*$ (191 questions) and HR-Bench (800 each).
All numbers are means of three evaluation runs.
As Propositions~\ref{prop:independent} and~\ref{prop:shortcut} predict, the gain appears on sets and not~on~single~images.}
\label{tab:lambda0}
\centering
\scriptsize
\setlength{\tabcolsep}{7.05pt}
\settowidth{\numw}{\textbf{00.0}}
\begin{tabular}{llccccc}
\toprule
\midrule
Set accuracy & $\lambda$ & 0.5\,MP & 1\,MP & 2\,MP & 4\,MP & Max \\
\midrule
\shaderow{vira9bg}ViRA-Bench & 0.5 & \nb{\textbf{38.7}} & \nb{\textbf{46.9}} & \nb{\textbf{55.6}} & \nb{\textbf{57.5}} & \nb{\textbf{63.3}} \\
           & 0   & \nb{31.3} & \nb{41.2} & \nb{50.0} & \nb{52.9} & \nb{58.7} \\
\addlinespace
\shaderow{vira9bg}ViRA-Bench-Real & 0.5 & \nb{\textbf{31.2}} & \nb{\textbf{49.6}} & \nb{\textbf{57.7}} & \nb{\textbf{68.8}} & \nb{\textbf{77.4}} \\
                & 0   & \nb{23.5} & \nb{41.9} & \nb{56.8} & \nb{65.4} & \nb{70.1} \\
\midrule
\bottomrule
\end{tabular}

\vspace{4pt}
\settowidth{\numw}{\textbf{00.0}}
\begin{tabular}{lcccccc}
\toprule
\midrule
 & \multicolumn{3}{c}{Single images, 1\,MP} & \multicolumn{3}{c}{Single images, max} \\
\cmidrule(lr){2-4}\cmidrule(l){5-7}
$\lambda$ & V$^*$ & HR-4K & HR-8K & V$^*$ & HR-4K & HR-8K \\
\midrule
\shaderow{vira9bg}0.5 & \nb{92.0} & \nb{87.5} & \nb{84.7} & \nb{\textbf{96.2}} & \nb{91.0} & \nb{\textbf{90.1}} \\
0   & \nb{\textbf{93.0}} & \nb{\textbf{87.8}} & \nb{\textbf{85.7}} & \nb{95.6} & \nb{\textbf{92.9}} & \nb{89.8} \\
\midrule
\bottomrule
\end{tabular}
\end{table}

\subsection{Prior methods within the framework}
\label{sec:theory-related}

We follow the formulas of each paper and state our reading where a paper leaves a detail open.
Rewards \emph{optimize} $J_R$ if they give an unbiased estimate of $\nabla_\theta J_R$ up to a positive factor when they are used with the score-function estimate and any baseline independent of the rollout's own sample.
Group normalization, length normalization and clipping change the estimator but not the objective.

\begin{proposition}[Objectives of prior methods]
  \label{prop:related}
  The rewards of the following methods optimize $J$ in Eq.~(\ref{eq:objective}) for the stated $K$, $\lambda$, $r$ and $c$.
  (a)~GRPO, RLOO and Dr.~GRPO: $K=1$, $\lambda=0$.
  (b)~DNA-Train: $K=2$ (a real and an edited video with different answers), $\lambda=0$, $r$ the sum of correctness and a format reward.
  (c)~TC-GRPO of COMET: $K=2$ (a video and its reversal with the same answer), $\lambda=0$, each version weighted by its number of rollouts (Remark~\ref{rem:unbiased}(i)).
  (d)~ConsistRoll: $K=2$ (an image and its rotated or noisy copy with the same answer), $r=c$, $\lambda$ its consistency weight, provided that two correct answers to a question are parsed as equal.
  (e)~CSPO: $K=2$ (an image and its object-removed copy), $c_o=\mathrm{Det}(y_o)$, $c_c=\mathrm{Abs}(y_c)$, $\lambda=1/(1+\gamma)$, $r_o=c_o+(R_{\mathrm{mask}}+\alpha\,\mathrm{Fmt}(y_o))/(1+\gamma)$, $r_c=c_c+\alpha\,\mathrm{Fmt}(y_c)/(1+\gamma)$, provided that every output either detects or abstains.
  (f)~CRPO with $w_{\mathrm{aug}}=1$: $K=2$ (a video and its flipped or reversed copy), $\lambda=\lambda_z$, $r=c+R_{\mathrm{fmt}}$, with $c$ on the original and the relational check $b$ on the copy.
\end{proposition}

\begin{proof}
  By Lemma~\ref{lem:product-score}, rewards of the form $\psi_v(o_v)R^{(v)}$ with $\E[\psi_v(o_v)R^{(v)}]=\E[\psi_v(o_v)R]$ optimize $J_R$.
  (a)--(c)~The rewards have no cross-version term, so the update is $\sum_vn_v\nabla_\theta V_v$.
  (d)~Both rollouts of pair $i$ receive $c_{v,i}+\lambda S_i$.
  When correct answers parse equally, the bonus condition (correct and agreeing) holds exactly when both are correct, so $S_i=c_{0,i}c_{1,i}$.
  This is $R^{(v)}=c(o_v)+\lambda c(o_0)c(o_1)$, which differs from $R_\star$ by the term $c(o_u)$ with $u\neq v$.
  This term is independent of $o_v$ and contributes nothing by Corollary~\ref{cor:baseline}.
  (e)~If every output either detects or abstains, then $\mathrm{Abs}=1-\mathrm{Det}$ and $R_{\mathrm{CPR}}=(1+\gamma)(c_o+c_c-1)+c_oc_c$.
  Moreover, $R_{\mathrm{mask}}=\mathrm{IoU}_o\,c_o$ depends only on the original's rollout.
  The pair reward is $(1+\gamma)\big(r_o+r_c+\lambda c_oc_c\big)-(1+\gamma)$, which is a positive multiple of $R_\star$ plus a constant, and it is given to both rollouts of the pair.
  (f)~This is shown in Proposition~\ref{prop:crpo} below.
\end{proof}

ConsistRoll and CSPO assign to rollout $(v,i)$ the reward of its index-paired combination.
Their estimators are therefore fixed-pairing estimators $\hat g^b_{\mathrm{id}}$ of their own set-level rewards, and Theorem~\ref{thm:rb} applies to them with the baseline held fixed.
If any of the two-version methods is applied to every pair of a set with $K\ge3$ versions, it optimizes $J_{R_{\mathrm{pair}}}$ of Eq.~(\ref{eq:pair-reward}).
By Lemma~\ref{lem:additivity}, this is CrossRPO with $R=R_{\mathrm{pair}}$, and by Proposition~\ref{prop:not-pairwise}, it cannot express the joint term of CrossRPO.
The closest method is CRPO, which makes the same choices as CrossRPO for the design questions (D1) and (D2).

\begin{proposition}[CRPO is CrossRPO with two versions]
  \label{prop:crpo}
  Ignoring the stabilizer in its standard deviation, CRPO assigns rollout $i$ of branch $v$ the advantage
  \begin{equation*}
    \hat A^{\mathrm{CRPO}}_{v,i}=\alpha_v\,\frac{n-1}{n\,\sigma_{\mathrm{shared}}}\,A_{v,i},
  \end{equation*}
  where $A_{v,i}$ is Eq.~(\ref{eq:advantage}) with $K=2$, the choices of Proposition~\ref{prop:related}(f) and $n$ rollouts per branch, and $\sigma_{\mathrm{shared}}$ is CRPO's standard deviation pooled over both branches.
  The factor $\alpha_v$ is 1 on the original and $w_{\mathrm{aug}}$ on the copy.
\end{proposition}

\begin{proof}
  On the original, CRPO's reward is $R_i=(1+\lambda\bar b)\,c_i+f_i$, where $f$ is the format reward and $\bar b=\hat p_{\mathrm{CF}}$ is the mean of $b$ over the copy's rollouts.
  Its deviation from the branch mean is $(1+\lambda\hat p_{\mathrm{CF}})(c_i-\bar c)+(f_i-\bar f)$.
  By Lemma~\ref{lem:loo}(b), this equals $\frac{n-1}{n}\big[(r_i-\bar r_{-i})+\lambda\hat p_{\mathrm{CF}}(c_i-\bar c_{-i})\big]$ with $r=c+f$.
  Because $w_{\mathrm F}=\hat p_{\mathrm{CF}}$ for $K=2$, this is $\frac{n-1}{n}A_{\mathrm F,i}$.
  The copy is symmetric with the extra factor $w_{\mathrm{aug}}$, and both are divided by $\sigma_{\mathrm{shared}}$.
\end{proof}

CRPO is thus CrossRPO for $K=2$, except that it uses one scale per set instead of one per batch (Proposition~\ref{prop:scale}) and weights the copy by $w_{\mathrm{aug}}$.
CrossRPO extends its coupling to any $K$ through $w_v$ (Propositions~\ref{prop:closed-form} and~\ref{prop:order}), and to any correctness-based set reward through $w^\Phi_v$ (Proposition~\ref{prop:multilinear}).
The other methods differ in at least one further choice (Table~\ref{tab:related}).
ConsistRoll and CSPO pair rollouts by index (Theorem~\ref{thm:rb}), and ConsistRoll and COMET pool the versions (Propositions~\ref{prop:baseline}, \ref{prop:balance} and~\ref{prop:prior}).
GRPO, COMET, ConsistRoll and DNA-Train divide by a standard deviation per group or version (Proposition~\ref{prop:scale}).
In addition, CSPO gives both rollouts of a pair the whole pair reward, so each rollout is credited with its partner's terms.
By Proposition~\ref{prop:baseline}(a), these terms add zero-mean noise.
CSPO also uses a length-normalized joint ratio.

Two methods fall outside Eq.~(\ref{eq:objective}).
NoisyRollout samples rollouts on a noisy image but evaluates their log-probabilities on the clean image, so its update is not a score-function estimate on either input.
Its target is $K=1$, $\lambda=0$ on the clean image.
BPO rewards a rollout for agreeing or disagreeing with the answer of the anchor rollout of its pack.
Its reward is therefore not a function of one rollout per version in the sense of Definition~\ref{def:objective}, and the anchor receives no policy-gradient term.
Without these contract terms and its pack penalty, BPO optimizes $\lambda=0$ over the non-anchor views of a pack.

\subsection{Offline check}
\label{sec:theory-check}

We check three of the results on 1,376 sets (16,090 rollouts) from a run trained with one GRPO group per set.
The version explains 39.5\% of the reward variance within a set, and the per-version baseline has 74.9\% of the squared residual of the pooled baseline (Proposition~\ref{prop:baseline}).
Averaging the index-paired joint term over all pairings removes 63.8\% of its squared size (Theorem~\ref{thm:rb}).
Finally, consider rollouts that answer the original correctly while every rollout of the variants is wrong (Proposition~\ref{prop:prior}).
They receive a mean advantage of $+1.69$ under GRPO and $+0.11$ under CrossRPO (both scaled to a unit root mean square~over~the~batch).

\subsection{Summary}
\label{sec:theory-summary}

The section reduces to one chain of implications.

\emph{Objective.}
On a single version, a blind policy can be perfect, so per-version accuracy cannot certify that answers come from the evidence.
The number $N$ of correctly answered versions can, because $\Pr(N=K)\le K^{-K}$ and $\E\big[\binom Nk\big]\le\binom KkK^{-k}$ for every blind policy (Proposition~\ref{prop:blind}).
Among correctness-based set rewards, the threshold $m=K$ bounds blind policies most tightly (Corollary~\ref{cor:threshold}), and rewards for agreement are maximized by blind policies when the answers differ (Proposition~\ref{prop:agreement}).
However, the weight of $m=K$ vanishes as $K$ grows, and the order $k=\min(K,3)$ keeps a weight of order $p^2$ (Proposition~\ref{prop:order}).
This gives the objective $J=\sum_vV_v+\lambda\,e_k(p)/\binom{K-1}{k-1}$ of Eq.~(\ref{eq:objective}), which is $\sum_vV_v+\lambda\prod_vp_v$ for $K\le3$.

\emph{Estimator.}
For every set-level reward, $\nabla_\theta J_R=\sum_v\E[\psi_v\bar R_v]$ (Lemma~\ref{lem:projection}).
We estimate the projection $\bar R_v$ by the average over all combinations of the other versions' rollouts and center it within the version.
The resulting estimator is unbiased for every $R$ (Theorem~\ref{thm:unbiased}).
For $R_\star$, its closed form is the CrossRPO advantage (Proposition~\ref{prop:closed-form}),
\begin{equation*}
  A_{v,i}=\underbrace{\big(r_{v,i}-\bar r_{v,-i}\big)}_{\text{RLOO within version }v}
  +\lambda\,\underbrace{w_v}_{\widehat{\partial J/\partial p_v}}\,\big(c_{v,i}-\bar c_{v,-i}\big),
\end{equation*}
where $w_v=\prod_{u\neq v}\hat p_u$ for $K\le3$ and $w_v$ is the mean of $\hat p_u\hat p_{u'}$ over the pairs of other versions for $K>3$.

\emph{Design choices.}
Each of (D1)--(D3) meets one requirement that the alternative used in prior work does not (Table~\ref{tab:summary}).
Averaging over all combinations removes the variance of an arbitrary pairing.
The per-version baseline removes the difficulty gap between versions from the variance and gives no credit to answers from the prior.
The batch scale keeps advantages proportional to reward differences.

\emph{Generality.}
The construction holds for any number of versions and any set-level reward.
With correctness-based rewards, $w_v$ is the estimated marginal value $\partial_v\tilde\Phi$ of being correct on version $v$ (Proposition~\ref{prop:multilinear}).
It reduces to RLOO for $K=1$, and it reduces to GRPO only if the division by each group's standard deviation is reinstated (Proposition~\ref{prop:special}).
The objectives of prior multi-version methods are special cases with $K\le2$, and CRPO coincides with CrossRPO for $K=2$ up to its scale (Propositions~\ref{prop:related} and~\ref{prop:crpo}).
Applying them to every pair is the special case $R=R_{\mathrm{pair}}$, whose advantage is the sum of the pairwise advantages (Lemma~\ref{lem:additivity}).
For $K\ge3$, the joint term is an interaction of order three that no sum over pairs reproduces (Proposition~\ref{prop:not-pairwise}), and it bounds blind policies by $K^{-3}$ instead of $K^{-2}$ (Corollary~\ref{cor:pairwise}).

\emph{Grouping.}
The joint term separates reading from a shortcut only on sets whose members share the shortcut and have pairwise different answers (Proposition~\ref{prop:shortcut}).
Contrast sets meet this condition for answering from the scene, and sets of photos with one question meet it for the question prior.
For members drawn independently, the expected objective is an increasing function of the mean accuracy.
The expected update is then RLOO with the accuracy reward scaled by $1+\lambda\bar p^{\,k-1}$, plus variance from the random partners (Proposition~\ref{prop:independent}).

In short, the CrossRPO update of a batch $\mathcal B$ is
\begin{equation*}
  \frac1s\sum_{\mathcal S\in\mathcal B}\sum_{v,i}\psi_{v,i}\Big(\hat R_v(o_{v,i})-\frac1{n-1}\sum_{j\neq i}\hat R_v(o_{v,j})\Big),
\end{equation*}
where $\hat R_v$ averages $R_\star$ over all combinations of the other versions' rollouts (Eq.~(\ref{eq:empirical-projection})) and $s$ is the root mean square of the advantages in the batch.
CrossRPO thus projects the set reward onto each version, averages the projection over all combinations, centers it within the version, and scales it by one statistic per batch.
Each of these four steps answers one requirement of Appendices~\ref{sec:theory-unbiased}, \ref{sec:theory-variance},~\ref{sec:theory-credit}~and~\ref{sec:theory-scale}.

\section{Data}
\label{sec:supp-data}

This section describes ViRA-Bench (Appendix~\ref{sec:supp-bench}) and tests whether its edits leave traces that guide the agents (Appendix~\ref{sec:supp-traces}).
It then describes ViRA-Bench-Real (Appendix~\ref{sec:supp-real}) and the training~data~(Appendix~\ref{sec:supp-train}).

\subsection{ViRA-Bench}
\label{sec:supp-bench}

\paragraph{Photos.}
The photos come from Wikimedia Commons and have at least 24\,MP, a long side of at most 14,000\,px and an aspect ratio between 1.2 and 2.4.
All photos are in color.
All licenses allow modification (CC BY, CC BY-SA, CC0 or public domain), and every photo keeps its source page, author and license so that the release can credit each image.

\paragraph{Objects.}
A vision-language model proposes objects whose appearance could plausibly be different, such as a sign, a light, a small device or a part of a vehicle.
SAM~3~\cite{carion2025sam3} then segments them at full resolution.
Every object has a longer side of at most 13\,px at 1\,MP and a shorter side of at least 28\,px in the photo, so it can be read after zooming but not without zooming.
It is a whole physical object or a printed design such as a pictogram, and not text, a face, a reflection, a shadow or one of a mass of similar things such as leaves.
Other objects of the same kind may appear if the question singles out the target.

\paragraph{Questions and edits.}
For each object, a language model writes a question and an edit instruction for each of two variants (A, B).
The question has four options, which are the answers for the original (O), A and B and a distractor.
The answers of the three versions are pairwise different, so a set tests whether the agent's answer changes with the object.
The four options belong to one category, such as arrows that point up, up-left and up-right.
Changes of symbol, shape, direction and state are preferred over color, because models tend to assume the usual look of an object without looking closely.
The question first locates the object and only then asks about it.
It usually names a large part of the photo first and then the item that carries the object, and it names neither the answer nor a property that reveals it.
Its answer requires no reading of text, and any reasoning uses only common knowledge.
Besides questions about the object itself, some questions add one step of common-sense reasoning, such as what a sign allows, and some compare the object with an unedited object elsewhere in the photo.
Before editing, the plan is checked on the original, so that the answer for O is correct.
An image editing model then edits an enlarged crop around the object according to each instruction.
The edited crop is registered to the original, and the change is restricted to the object mask.
The result is then pasted back into the full-resolution photo, so the versions differ only in the region of the old and the new object.
For 108 sets, both variants are chosen to contradict common sense or another cue in the photo, so a guess gives the original's answer.
Table~\ref{tab:data-models} names the models of each step.
Several models propose the objects and write the questions, so that the preferences of one model do not run through all sets.

\begin{table}[htbp]
\caption{Models in the pipeline of ViRA-Bench and the number of sets that each one handles.}
\label{tab:data-models}
\centering
\scriptsize
\begin{tabular}{@{}llr@{}}
\toprule
\midrule
Step & Model & Sets \\
\midrule
Propose the object & Claude Sonnet~5.5 & 480 \\
 & Gemini~3.1 Pro & 49 \\
 & GPT-6.1-Terra & 233 \\
\midrule
Write the question and the edits & Claude Opus~5.5 & 529 \\
 & GPT-6-Astra & 233 \\
\midrule
Edit the crop & GPT-5.4 Image~2 & 480 \\
 & GPT Image~2 & 282 \\
\midrule
Check the set & three models and Gemini~3.1 Pro & 480 \\
 & the models of the first two steps & 282 \\
\midrule
\bottomrule
\end{tabular}
\end{table}

\paragraph{Review.}
Vision-language models answer each version from a crop around the object and check that the question identifies the object in the photo.
A realism check also compares each variant with the original.
These checks only flag candidates.
Human reviewers then check every question against the three versions and the object at full resolution.
They accept, correct or rewrite each question.

\paragraph{Statistics.}
Table~\ref{tab:data-stats} summarizes the three datasets.
The 762 sets of ViRA-Bench come from 466 photos, because one photo can have several edited objects and one object can have~two~\mbox{questions}.

\begin{table}[htbp]
\caption{Dataset statistics (medians for the region and the object; the region and the object's share are percentages of the photo, and its longer side is measured at 1\,MP). Real: ViRA-Bench-Real.
An image set is one object with its versions, and two questions can share one image set.}
\label{tab:data-stats}
\centering
\scriptsize
\setlength{\tabcolsep}{12.10pt}
\begin{tabular}{@{}lccc@{}}
\toprule
\midrule
 & ViRA-Bench & Real & ViRA-Train \\
\midrule
Contrast sets (questions) & 762 & 78 & 6,457 \\
Image sets & 544 & 78 & 4,462 \\
Photos & 466 & 234 & 3,797 \\
Photo size (MP) & 24--108 & 45.4 & 21--120 \\
Versions per set & 3 & 3 & 2--3 \\
Region that differs & 0.010\% & -- & 0.022\% \\
Object (share of photo) & 0.0075\% & 0.0017\% & 0.013\% \\
Object (longer side) & 10\,px & 6\,px & -- \\
Questions about text & 0 & 32 & 5,329 \\
Options & 4 & 4 & 4 \\
\midrule
\bottomrule
\end{tabular}
\end{table}

\subsection{Editing traces and the first zoom}
\label{sec:supp-traces}

The three versions of a set share the question and the position of the object, and A and B differ from O only in the region of the old and the new object.
If editing left traces that showed an agent where to look, its first zoom would cover the object more often on A and B than on the unedited O.
Table~\ref{tab:traces} compares the two for eight agents in the runs of Table~\ref{tab:full}.
As in Appendix~\ref{sec:supp-errors}, a zoom covers the object if its box on the photo contains at least half of the object's box.
For every agent and budget, the share on the variants is between 2.4 points below and 1.2 points above the share on O.
Averaged over the five budgets, the difference is 0.0 points for ViRA-9B and at most 0.9 points for the other agents.
ViRA and $\lambda=0$ were trained on edited sets, but their differences are no larger than those of the agents that never saw such sets.
Moreover, 1,416 of the 1,664 contrast sets that train ViRA-9B and 1,160 of the 1,338 that train ViRA-27B were edited by Qwen-Image-Edit-2511, a different model from the editors of ViRA-Bench.
The open agents decode greedily, and in 35\% to 73\% of the cases they draw exactly the same first box on O and on a variant.
Editing therefore does not guide the search.
Since the first zoom finds the object equally often in all three versions, the higher accuracy on O does not come from the search either.
For ViRA-9B at 0.5\,MP, the first zoom covers the object on 87.6\%, 87.6\% and 86.8\% of the O, A and B images, but the accuracy is 73.4\%, 62.4\% and 59.5\%.
ViRA-Bench-Real has no editing traces, because its photos are not~edited~(Appendix~\ref{sec:supp-real}).

\begin{table*}[htbp]
\caption{\textbf{Editing traces and the first zoom} on ViRA-Bench (762 sets).
For each budget, the first column gives the share (\%) of original (O) images whose first zoom covers the object, and the second gives the share on the variants (A, B) minus the share on O, in points.
The three versions share the question and the position of the object, and only A and B are edited.
ViRA and $\lambda=0$ are means of three evaluation runs.
Claude Haiku~5.5 takes at most 3.8\,MP, so its 4\,MP columns repeat~its~max~columns.}
\label{tab:traces}
\centering
\scriptsize
\setlength{\tabcolsep}{5.4pt}
\settowidth{\numw}{\textbf{00.0}}
\begin{tabular}{lcccccccccc}
\toprule
\midrule
 & \multicolumn{2}{c}{0.5\,MP} & \multicolumn{2}{c}{1\,MP} & \multicolumn{2}{c}{2\,MP} & \multicolumn{2}{c}{4\,MP} & \multicolumn{2}{c}{Max} \\
\cmidrule(lr){2-3}\cmidrule(lr){4-5}\cmidrule(lr){6-7}\cmidrule(lr){8-9}\cmidrule(l){10-11}
Agent & O & A/B $-$ O & O & A/B $-$ O & O & A/B $-$ O & O & A/B $-$ O & O & A/B $-$ O \\
\midrule
Qwen3.5-9B + tool & \nb{80.1} & \nb{$-$0.3} & \nb{80.6} & \nb{$-$0.3} & \nb{84.0} & \nb{$-$0.3} & \nb{86.5} & \nb{$-$0.1} & \nb{85.7} & \nb{+0.5} \\
ZwZ2-9B + tool & \nb{77.8} & \nb{$-$0.5} & \nb{78.7} & \nb{+0.4} & \nb{83.1} & \nb{$-$0.1} & \nb{85.0} & \nb{+1.2} & \nb{85.8} & \nb{$-$0.1} \\
$\lambda=0$ & \nb{86.5} & \nb{+0.7} & \nb{89.6} & \nb{+0.3} & \nb{91.2} & \nb{$-$0.3} & \nb{92.9} & \nb{$-$0.1} & \nb{94.3} & \nb{+0.4} \\
\shaderow{vira9bg}\textbf{ViRA-9B} & \nb{87.6} & \nb{$-$0.4} & \nb{89.9} & \nb{+0.4} & \nb{92.4} & \nb{0.0} & \nb{94.2} & \nb{$-$0.1} & \nb{96.2} & \nb{+0.2} \\
\addlinespace
Qwen3.8-27B + tool & \nb{75.5} & \nb{$-$1.1} & \nb{81.5} & \nb{$-$0.3} & \nb{86.9} & \nb{$-$0.3} & \nb{89.5} & \nb{$-$0.3} & \nb{90.6} & \nb{+0.1} \\
\shaderow{vira27bg}\textbf{ViRA-27B} & \nb{94.4} & \nb{+0.3} & \nb{95.4} & \nb{+0.3} & \nb{95.7} & \nb{0.0} & \nb{96.9} & \nb{+0.1} & \nb{97.5} & \nb{+0.2} \\
\addlinespace
Claude Haiku~5.5, max thinking & \nb{78.1} & \nb{+1.0} & \nb{78.3} & \nb{$-$0.5} & \nb{83.2} & \nb{$-$0.6} & \nb{82.9} & \nb{+1.0} & \nb{82.9} & \nb{+1.0} \\
GPT-6-Luna, max thinking & \nb{67.5} & \nb{$-$0.3} & \nb{71.5} & \nb{$-$0.9} & \nb{85.2} & \nb{$-$1.4} & \nb{73.8} & \nb{+0.6} & \nb{29.4} & \nb{$-$2.4} \\
\midrule
\bottomrule
\end{tabular}
\end{table*}

\subsection{ViRA-Bench-Real}
\label{sec:supp-real}

ViRA-Bench-Real contains 78 sets of three photos each, 234 photos in total.
The three photos of a set show the same scene, and only the object differs between them, while the camera, the light and everything else stay in place.
The objects are printed posters and signs, stickers, labels and small everyday objects, and some variants contradict common sense.
The photos were taken in classrooms, laboratories, a library and atriums of university buildings, with a Nikon Z8 camera (8256$\times$5504\,px) and a NIKKOR~Z 14-24mm f/2.8~S lens on a tripod.
All photos use manual exposure at ISO~64 without flash, focal lengths of 14 to 24\,mm and apertures of f/2.8 to f/6.3, and the three photos of a set share all settings.
The questions are written from the photos and reviewed by humans, and the photos carry no metadata.

\subsection{ViRA-Train and the training data}
\label{sec:supp-train}

\paragraph{ViRA-Train.}
Of its 6,457 contrast sets, 5,329 concern text, such as the words on a sign or a label, and 1,128 concern objects.
For a text set, we detect the text in the photo, and Qwen2.5-VL-72B writes a question and alternative words.
Qwen-Image-Edit-2511 then replaces the words in the variants, and the result must pass an OCR check, a check by Qwen2.5-VL-72B and a review by Claude.
Of the text questions, 2,347 ask for reasoning about the words, for example what a sign allows.
Claude Sonnet~5.5 writes these questions, and they have answers for A and B only.
Of the object questions, 347 come from sets built like ViRA-Bench with Claude Sonnet~5.5, Claude Opus~5.5 and GPT-5.4 Image~2 (Table~\ref{tab:data-models}).
For the other 781, Qwen2.5-VL-72B finds small objects, proposes their changes and writes the questions.
Qwen-Image-Edit-2511 then edits the objects, and Claude Opus~5.5 checks the result.
No ViRA-Train photo is a ViRA-Bench photo or a near-duplicate of one, such as a photo of the same series or a visually similar photo by the same author.
ViRA-Train contains no contrast set that Qwen3.5-9B or Qwen2.5-VL-7B answers correctly on every version at 1\,MP without a tool and with greedy decoding, since such sets give no learning signal.

\paragraph{Training data.}
Table~\ref{tab:train-mix} lists the training data of ViRA-9B and ViRA-27B.
For ViRA-9B, the contrast sets are the 1,664 sets of ViRA-Train that Qwen3.5-9B does not answer entirely correctly when it views the photo at 12.8\,MP without the tool.
The single images serve two purposes: they keep the agent's skills on ordinary questions, and they show that CrossRPO can train on sets and single images in one batch.
The open questions about Commons photos were written by Claude Opus~5.5 for the six regions of a $2\times3$ grid on each photo and checked by Gemini~3.7 Flash.
For the multi-evidence questions, Gemini~3.7 Flash lists the objects in each region, and Claude Opus~5.5 writes questions of six types (position, chained reference, order, counting, attribute or absence, and reading combined with reasoning).
A program computes the options of the closed types from the object boxes, so that their wording cannot reveal the answer.
Gemini then checks that every object named in a question occurs exactly once in the photo.
Every single image is neither too easy nor unsolvable for the base model, which answers it correctly in at least one and at most six of eight attempts.
The training data of ViRA-27B are built from the same sources with the same rules applied to Qwen3.8-27B, which gives 1,338 contrast sets and 2,063 single images.

\begin{table}[htbp]
\caption{Training data of ViRA-9B and ViRA-27B: number of items per source.
Math: 70 geometry, 40 table and 40 arithmetic problems for ViRA-9B, and 70, 29 and 40 for ViRA-27B.
$K$: number of versions of an item. \emph{Answer}: form of its answer.}
\label{tab:train-mix}
\centering
\scriptsize
\setlength{\tabcolsep}{3.00pt}
\begin{tabular}{@{}llrrl@{}}
\toprule
\midrule
Source & $K$ & ViRA-9B & ViRA-27B & Answer \\
\midrule
Contrast sets & 2--3 & 1,664 & 1,338 & multiple choice \\
Commons detail questions & 1 & 599 & 600 & open \\
Multi-evidence, photos & 1 & 859 & 949 & choice or open \\
Multi-evidence, maps & 1 & 170 & 175 & choice or open \\
Synthetic charts (5 types) & 1 & 200 & 200 & open \\
Math, ThinkLite-VL~\cite{wang2025thinklite} & 1 & 150 & 139 & open \\
\midrule
Total & & 3,642 & 3,401 & \\
\midrule
\bottomrule
\end{tabular}
\end{table}

\section{Agent and implementation}
\label{sec:supp-impl}

This section gives the zoom tool (Appendix~\ref{sec:supp-tool}), the prompts (Appendix~\ref{sec:supp-prompts}), the rollout loop and the CrossRPO advantage as we run them (Appendix~\ref{sec:supp-loop}), and the training setup (Appendix~\ref{sec:supp-training}).
Listings~\ref{lst:tool}, \ref{lst:episode} and~\ref{lst:advantage} are shortened from our implementation.

\begin{lstlisting}[style=code, label={lst:tool}, caption={\textbf{Zoom tool} of ViRA (Python, shortened from our code). \texttt{clamp} sorts the box, clips it to the image, widens each side to at least 32\,px around its center and rejects an empty box or one with an aspect~ratio~above~100.}]
PATCH, MIN_PIXELS, MAGNIFY = 32, 65536, 4.0   # Qwen3.5 / Qwen3.8: one visual token is 32 x 32 px

def grid(w, h, budget):
    """Size of an image on the 32-px grid with at most `budget` and at least MIN_PIXELS pixels."""
    W, H = max(PATCH, round(w / PATCH) * PATCH), max(PATCH, round(h / PATCH) * PATCH)
    if W * H > budget:
        s = sqrt(w * h / budget)
        W, H = max(PATCH, floor(w / s / PATCH) * PATCH), max(PATCH, floor(h / s / PATCH) * PATCH)
    elif W * H < MIN_PIXELS:
        s = sqrt(MIN_PIXELS / (w * h))
        W, H = ceil(w * s / PATCH) * PATCH, ceil(h * s / PATCH) * PATCH
    return W, H

def shown_size(w, h, budget):
    """A crop larger than the budget is shrunk to it; a smaller crop is enlarged to fill it, by at most MAGNIFY per side."""
    m = min(MAGNIFY, sqrt(budget / (w * h)))
    if m > 1:
        w, h = round(w * m), round(h * m)
    return grid(w, h, budget)

def zoom(photo, frames, bbox_2d, source, budget):
    """photo: the full-resolution image. frames[k] = (region of the photo, size shown) of image k,
    where k = 0 is the first view and k >= 1 is observation_k."""
    (rx0, ry0, rx1, ry1), (fw, fh) = frames[source]
    x0, y0, x1, y1 = (v * s / 1000 for v, s in zip(bbox_2d, (fw, fh, fw, fh)))   # 0-1000 -> pixels of image k
    x0, y0, x1, y1 = clamp(x0, y0, x1, y1, fw, fh, min_side=PATCH)
    sx, sy = (rx1 - rx0) / fw, (ry1 - ry0) / fh                                  # pixels of image k -> photo
    box = (floor(rx0 + x0 * sx), floor(ry0 + y0 * sy), ceil(rx0 + x1 * sx), ceil(ry0 + y1 * sy))
    crop = photo.crop(box)                                  # always from the full-resolution photo
    size = shown_size(crop.width, crop.height, budget)
    frames.append((box, size))                              # the crop becomes observation_{len(frames) - 1}
    return crop.resize(size, Image.BICUBIC)
\end{lstlisting}

\begin{lstlisting}[style=code, label={lst:episode}, caption={\textbf{One rollout} of ViRA (Python, shortened). Training runs the same loop inside verl. \texttt{FIRST\_TURN} and \texttt{TOOL\_RESPONSE} are the user turns of Listing~\ref{lst:turns}.}]
MAX_CALLS, MAX_NEW_TOKENS = 6, 4096

def rollout(policy, photo, question, budget):
    view = photo.resize(grid(photo.width, photo.height, budget), Image.BICUBIC)   # the first view
    frames = [((0, 0, photo.width, photo.height), view.size)]
    msgs = [system(SYSTEM_PROMPT), user(view, FIRST_TURN.format(question=question))]
    for turn in range(MAX_CALLS + 1):
        reply = policy(msgs, max_new_tokens=MAX_NEW_TOKENS, stop="</tool_call>")
        if has_answer(reply) or "<tool_call>" not in reply:
            return last_answer(reply)          # the text of the last <answer>...</answer>, or None
        if turn == MAX_CALLS:
            return None                        # a seventh call: no answer
        msgs.append(assistant(reply))
        try:
            call = parse_tool_call(reply)      # the last <tool_call>, as JSON
            k = source_index(call.get("source", "original_image"), len(frames) - 1)
            crop = zoom(photo, frames, call["bbox_2d"], k, budget)
        except ValueError as e:                # an invalid call: the error goes back, and the agent may try again
            msgs.append(user(f"<tool_response>Error: {e}</tool_response>"))
            continue
        msgs.append(user(crop, TOOL_RESPONSE.format(k=len(frames) - 1, question=question)))
    return None
\end{lstlisting}

\begin{lstlisting}[style=code, label={lst:advantage}, caption={\textbf{CrossRPO advantage} (Python, shortened from our code) for sets of up to three versions, as all our sets. It computes Eq.~(\ref{eq:method-advantage}) and divides all advantages of the batch by their root mean square.}]
def crossrpo_advantages(batch, lam=0.5):
    """batch: a list of sets. A set maps each version to the correctness (1 or 0) of its rollouts.
    A single image is a set with one version, and its advantage is plain RLOO."""
    adv = {}
    for j, s in enumerate(batch):
        p = {v: mean(c) for v, c in s.items()}                        # accuracy of each version
        for v, c in s.items():
            w = prod(p[u] for u in s if u != v) if len(s) > 1 else 0.0  # all other versions correct
            n = len(c)
            for i in range(n):
                loo = (sum(c) - c[i]) / (n - 1)                          # the other rollouts of version v
                adv[j, v, i] = (1 + lam * w) * (c[i] - loo)
    rms = sqrt(mean(a * a for a in adv.values()))
    return {key: a / rms for key, a in adv.items()}                   # one scale for the whole batch
\end{lstlisting}

\subsection{Zoom tool}
\label{sec:supp-tool}

At a budget of $x$\,MP, the agent first sees the photo resized to at most $x$\,MP with its aspect ratio kept.
Both sides are multiples of 32\,px, the size of one visual token of Qwen3.5 and Qwen3.8, and every image has at least 65,536 pixels.
The agent zooms by calling \texttt{image\_zoom\_in\_tool} with a box \texttt{bbox\_2d} $=[x_1,y_1,x_2,y_2]$ and a \texttt{source}.
The source names the image the box is drawn on.
It is either \texttt{original\_image}, the first view, or \texttt{observation\_}$k$, the $k$-th crop returned so far.
The box is given on a 0--1000 scale of that image.
The tool maps the box to the pixels of that image, sorts and clips it, and widens each side to at least 32\,px.
It rejects an empty box and a box with an aspect ratio above 100.
It then converts the box to coordinates of the full-resolution photo, using the region of the photo that the image shows.
The crop is always cut from the full-resolution photo, so a crop of a crop loses no resolution.
A crop larger than the budget is shrunk to $x$\,MP.
A smaller crop is enlarged until it fills the budget, but by at most $4\times$ per side.
Listing~\ref{lst:tool} gives the code.

\subsection{Prompts}
\label{sec:supp-prompts}

The system prompt (Listing~\ref{lst:system}) is the prompt of DeepEyes~\cite{zheng2025deepeyes} with three additions.
The box description states the 0--1000 scale, the tool takes the \texttt{source} parameter, and two sentences ask the agent to zoom in step by step.
The first user turn holds the first view, the question with its lettered options and the answer format of DeepEyes (Listing~\ref{lst:turns}).
After each zoom, the next user turn holds the crop and names it.
It then suggests how to go on and asks the question again.
A failed call returns an error message, and the agent may try again within the same limit.

\begin{lstlisting}[style=prompt, breakatwhitespace=false, label={lst:system}, caption={\textbf{System prompt} of ViRA-9B and ViRA-27B in training and evaluation. It is the prompt of DeepEyes~\cite{zheng2025deepeyes} with the 0--1000 scale in the box description, the \texttt{source} parameter and two sentences on zooming in step by step. The \texttt{required} field is kept from DeepEyes, which names \texttt{bbox}~instead~of~\texttt{bbox\_2d}.}]
You are a helpful assistant.

# Tools
You may call one or more functions to assist with the user query.
You are provided with function signatures within <tools></tools> XML tags:
<tools>
{"type":"function","function":{"name":"image_zoom_in_tool","description":"Zoom in on a specific region of an image by cropping it based on a bounding box (bbox) and an optional object label. You can zoom in step by step: zoom in again on the latest observation with a tighter box, until the detail you need is clearly visible. If the question needs several parts of the image, zoom in on each of them in the original image.","parameters":{"type":"object","properties":{"bbox_2d":{"type":"array","items":{"type":"number"},"minItems":4,"maxItems":4,"description":"The bounding box of the region to zoom in, as [x1, y1, x2, y2], where (x1, y1) is the top-left corner and (x2, y2) is the bottom-right corner. Coordinates are relative to the image named by source, on a 0-1000 scale."},"source":{"type":"string","description":"The image to zoom in on: \"original_image\" (the first image, the default) or \"observation_i\", the i-th zoom-in image returned so far (observation_1 is the first). Zooming in on an observation looks closer at part of it; every crop is cut from the full-resolution original."},"label":{"type":"string","description":"The name or label of the object in the specified bounding box (optional)."}},"required":["bbox"]}}}
</tools>

# How to call a tool
Return a json object with function name and arguments within <tool_call></tool_call> XML tags:
<tool_call>
{"name": <function-name>, "arguments": <args-json-object>}
</tool_call>

**Example**:
<tool_call>
{"name": "image_zoom_in_tool", "arguments": {"bbox_2d": [10, 20, 100, 200], "label": "the apple on the desk"}}
</tool_call>
\end{lstlisting}

\begin{lstlisting}[style=prompt, label={lst:turns}, caption={\textbf{User turns.} \texttt{<image>} marks an image, and text in braces is filled in. For an open question, the options are replaced by ``Answer the question using a single word or phrase.''}]
--- First user turn: the first view and the question ---
<image>{question}
A. {option A}
B. {option B}
C. {option C}
D. {option D}
Think first, call **image_zoom_in_tool** if needed, then answer. Format strictly as:  <think>...</think>  <tool_call>...</tool_call> (if tools needed)  <answer>...</answer>

--- After the k-th successful zoom: the crop, a hint and the question again ---
<tool_response>observation_{k}:<image>
This is observation_{k}, a zoom-in of your last box. If the detail you need is still small or unclear, zoom in again on it ("source": "observation_{k}") with a tighter box; repeat until you can see it clearly. If the question also needs another part of the image, zoom in on that part of the original image ("source": "original_image"). When you have seen what you need, answer the question.
Question: {question}
A. {option A}
...
Think first, call **image_zoom_in_tool** if needed, then answer. Format strictly as:  <think>...</think>  <tool_call>...</tool_call> (if tools needed)  <answer>...</answer> </tool_response>

--- After a failed call ---
<tool_response>Error: {message}</tool_response>

--- Proprietary models: the text after the question (system prompt: "You are a helpful assistant.") ---
Think first, call image_zoom_in_tool if needed, then answer. Put your final answer, the letter of one option, inside <answer> and </answer>.
GPT-6-Luna at 30.7 MP also gets: You can call image_zoom_in_tool at most 6 times in total; after that you must give your final answer.
\end{lstlisting}

\subsection{Rollout and advantage}
\label{sec:supp-loop}

A rollout allows six tool calls, and a seventh call ends it without an answer (Listing~\ref{lst:episode}).
A reply with an \texttt{<answer>} span ends the rollout, and so does a reply without a tool call.
The answer is the text of the last \texttt{<answer>} span.
For a multiple-choice question, it is the option letter at the start of that span.
Each reply has at most 4,096 new tokens in training and in evaluation.
The models run with thinking disabled and write their reasoning inside \texttt{<think>} tags, as the prompt asks.
Listing~\ref{lst:advantage} computes the CrossRPO advantage of a batch.

\subsection{Training and compute}
\label{sec:supp-training}

\paragraph{Training.}
Table~\ref{tab:hyper} lists the hyperparameters.
Each model is trained with RL on its base model, using CrossRPO.
ViRA-9B makes one pass over its 3,642 items (227 steps), and ViRA-27B makes one pass over its 3,401 items (212 steps).
Every step samples 16 items, draws one budget per item and generates 12 rollouts per item with vLLM, split evenly over its versions.
The advantages of Eq.~(\ref{eq:method-advantage}) are computed per set and divided by their root mean square over the batch, and the policy is updated once per step.
Open answers are judged by GPT-4o against the reference answer with the prompt of Mini-o3~\cite{lai2025minio3} (Listing~\ref{lst:judge}).
We use the last checkpoint of each run.

\begin{lstlisting}[style=prompt, label={lst:judge}, caption={\textbf{Judge prompt} for open answers, taken verbatim from Mini-o3~\cite{lai2025minio3}. GPT-4o (gpt-4o-2024-11-20) gets the system prompt and the query, and an answer is correct if the reply~says~``Score:~1''.}]
--- System ---
You are an intelligent chatbot designed for evaluating the correctness of generative outputs for question-answer pairs.
Your task is to compare the predicted answer with the correct answer and determine if they match meaningfully. Here's how you can accomplish the task:
------
##INSTRUCTIONS:
- Focus on the meaningful match between the predicted answer and the correct answer.
- Consider synonyms or paraphrases as valid matches.
- Evaluate the correctness of the prediction compared to the answer.

--- User ---
I will give you a question related to an image and the following text as inputs:

1. **Question Related to the Image**: {question}
2. **Ground Truth Answer**: {ground_truth}
3. **Model Predicted Answer**: {prediction}

Your task is to evaluate the model's predicted answer against the ground truth answer, based on the context provided by the question related to the image. Consider the following criteria for evaluation:
- **Relevance**: Does the predicted answer directly address the question posed, considering the information provided by the given question?
- **Accuracy**: Compare the predicted answer to the ground truth answer. You need to evaluate from the following two perspectives:
(1) If the ground truth answer is open-ended, consider whether the prediction accurately reflects the information given in the ground truth without introducing factual inaccuracies. If it does, the prediction should be considered correct.
(2) If the ground truth answer is a definitive answer, strictly compare the model's prediction to the actual answer. Pay attention to unit conversions such as length and angle, etc. As long as the results are consistent, the model's prediction should be deemed correct.
**Output Format**:
Your response should include an integer score indicating the correctness of the prediction: 1 for correct and 0 for incorrect. Note that 1 means the model's prediction strictly aligns with the ground truth, while 0 means it does not.
The format should be "Score: 0 or 1"
\end{lstlisting}

\begin{table}[htbp]
\caption{Training hyperparameters of ViRA.}
\label{tab:hyper}
\centering
\scriptsize
\begin{tabular}{@{}ll@{}}
\toprule
\midrule
Setting & Value \\
\midrule
Base model & Qwen3.5-9B / Qwen3.8-27B \\
Thinking & off \\
Framework & verl~0.10~\cite{sheng2025hybridflow}, vLLM~0.32 rollouts \\
Hardware & 8 H200 GPUs \\
Training time & 12.7\,h (9B) / 29.6\,h (27B) \\
Steps & 227 (9B) / 212 (27B) \\
Items per step & 16 \\
Rollouts per item & 12 ($n=12/K$ per version) \\
$\lambda$ & 0.5; $w=0$ for $K=1$ \\
Advantage scale & batch root mean square \\
Policy loss & clipped ratio, $\epsilon=0.2$, token mean \\
Old policy & log-probabilities of the vLLM rollout \\
Updates per step & 1 \\
Optimizer & AdamW, $\beta=(0.9,0.999)$, weight decay 0.01 \\
Learning rate & $1\times10^{-6}$, constant, no warmup \\
Gradient clipping & global norm 1.0 \\
Precision & bf16, fp32 gradient reduction \\
KL penalty, entropy bonus & none \\
Sampling & temperature 1.0, top-$p$ 1.0 \\
Turns & up to 7 (6 zooms) \\
Tokens per reply & 4,096 \\
Tokens per rollout & 20,480 prompt, 40,960 generated \\
Pixel budget & log-uniform, 0.5--16.8\,MP \\
Crop magnification & up to $4\times$ \\
Reward & correctness \\
Judge of open answers & GPT-4o \\
27B parallelism & tensor 2, sequence 2 \\
\midrule
\bottomrule
\end{tabular}
\end{table}

\paragraph{Compute.}
Training takes 12.7 hours for ViRA-9B and 29.6 hours for ViRA-27B on eight H200 GPUs, which is about 100 and 240 GPU-hours.
Each ablation takes 13 to 22 hours~on~eight~GPUs.

\section{Baselines and benchmarks}
\label{sec:supp-baselines}

This appendix describes the 15 baselines and the benchmarks of Section~\ref{sec:exp}.

\subsection{Baselines}
\label{sec:supp-baseline-list}

Table~\ref{tab:models} lists the checkpoints and the training data that each baseline reports, and Appendix~\ref{sec:supp-eval} gives the tool and the settings of each model.

\paragraph{Open base models.}
\begin{itemize}
  \item \textbf{Qwen3.5-9B} and \textbf{Qwen3.8-27B} are the open vision-language models from which ViRA-9B and ViRA-27B are trained.
  Without a tool, they answer from the first view.
  With a zoom tool (\emph{+\,tool}), they show what the base model reaches before it is trained with CrossRPO.
  \item \textbf{ZwZ2-9B} and \textbf{ZwZ2-27B}~\cite{wei2026zoomingv2} are trained from Qwen3.5-9B and Qwen3.6-27B to read fine details and to connect several pieces of evidence in the whole image, without a zoom tool.
  We run them without a tool and with the zoom tool.
  \item \textbf{Vision-OPD-9B}~\cite{yuan2026visionopd} is trained from Qwen3.5-9B by on-policy self-distillation to see fine details without a tool.
  With the zoom tool, it zooms on 4.3\% of the ViRA-Bench images at 1\,MP (Table~\ref{tab:reading-full}).
\end{itemize}

\paragraph{Released zoom agents (7--9B).}
These agents are based on Qwen2.5-VL-7B~\cite{bai2025qwen25vl} and use their own tools and prompts.
\begin{itemize}
  \item \textbf{DeepEyes}~\cite{zheng2025deepeyes} is trained with RL to call a zoom tool that crops a box of the image, and a tool call adds to its reward.
  \item \textbf{DeepEyesV2}~\cite{hong2025deepeyesv2} extends DeepEyes to an agent that writes and runs code on the image.
  \item \textbf{Mini-o3}~\cite{lai2025minio3} is trained for visual search over many turns, and we allow the 32 turns of its~own~\mbox{evaluation}.
  \item \textbf{Thyme}~\cite{zhang2025thyme} writes code that crops, rotates or enhances the image and runs it in a sandbox.
  \item \textbf{Pixel Reasoner}~\cite{wang2025pixelreasoner} is trained with a curiosity reward to use pixel-space operations such as zooming into a region.
  \item \textbf{ZoomEye}~\cite{shen2025zoomeye} requires no training.
  It searches a tree of crops with Qwen2.5-VL-7B in our runs and answers from the crop it selects or from the whole image.
\end{itemize}

\paragraph{Released zoom agents (27--32B).}
\begin{itemize}
  \item \textbf{LBS-27B}~\cite{yao2026lbs} is trained from Qwen3.8-27B.
  It transfers training on GUI grounding to visual observation and crops a region with a text action, up to three times.
  \item \textbf{SenseNova-MARS-32B}~\cite{chng2025sensenovamars} is trained with RL to reason with a crop tool and two search tools.
  We run it offline, so its searches return no results.
\end{itemize}

\paragraph{Proprietary models.}
\begin{itemize}
  \item \textbf{Claude Haiku~5.5} and \textbf{GPT-6-Luna} are proprietary models that we query through their APIs.
  They call our zoom tool by function calling and run at three thinking levels.
  Their largest inputs are 3.8\,MP and 30.7\,MP.
\end{itemize}

\paragraph{Baseline settings in Figure~\ref{fig:bench}.}
The counts below each set in Figure~\ref{fig:bench} are over 15 baseline settings at 1\,MP: Claude Haiku~5.5 and GPT-6-Luna at their highest and medium thinking levels (four settings), the five open base models with the zoom tool (Qwen3.5-9B, Qwen3.8-27B, ZwZ2-9B, ZwZ2-27B and Vision-OPD-9B), and the six released 7--9B agents (DeepEyes, DeepEyesV2, Mini-o3, Thyme, Pixel Reasoner and ZoomEye).
A setting is counted if it gives the original's answer to all three versions~of~the~set.

\subsection{Benchmarks}
\label{sec:supp-benchmark-list}

\paragraph{Contrast-set benchmarks.}
The metric is set accuracy.
\begin{itemize}
  \item \textbf{ViRA-Bench} (Section~\ref{sec:data}) has 762 contrast sets on 466 photos of 24 to 108\,MP.
  Each set has an original and two edited variants with different answers.
  \item \textbf{ViRA-Bench-Real} (Appendix~\ref{sec:supp-real}) has 78 contrast sets that were\newline photographed~without~\mbox{editing}.
\end{itemize}

\paragraph{General benchmarks.}
Each question is about a single image, and the metric is accuracy.
\begin{itemize}
  \item \textbf{V$^*$}~\cite{wu2024vstar} has 191 multiple-choice questions about the attributes and the relative positions of small objects in high-resolution photos.
  It tests visual search.
  \item \textbf{HR-Bench}~\cite{wang2025hrbench} has 800 questions on 4K images and 800 on 8K images.
  They ask about one small object or about the relation between several objects.
  \item \textbf{MME-RealWorld-Lite}~\cite{zhang2025mmerealworld} has 1,919 multiple-choice questions from MME-RealWorld.
  Its tasks cover perception and reasoning on high-resolution images from five domains (text in the wild, remote sensing, diagrams and tables, monitoring, and autonomous driving).
  \item \textbf{TreeBench}~\cite{wang2026treebench} has 405 multiple-choice questions on dense photos, in which the target covers about 3\% of the image on average.
  It tests perception and reasoning that depend on a small region.
  \item \textbf{O3-Bench}~\cite{li2026insighto3} has 345 questions on large charts and maps with a median of 19\,MP.
  A question requires finding and combining several regions.
  \item \textbf{MMStar}~\cite{chen2024mmstar} has 1,500 multiple-choice questions that were selected so that they cannot be answered without the image.
  It tests general perception and reasoning, and most of its images are~below~1\,MP.
  \item \textbf{HallusionBench}~\cite{guan2024hallusionbench} asks yes-or-no questions about images that contradict common knowledge or contain visual illusions.
  It tests whether a model answers from the image or from its prior, and we use its 951 questions that come with an image.
\end{itemize}

\paragraph{Text benchmarks.}
\begin{itemize}
  \item \textbf{ConFiQA}~\cite{bi2024contextdpo} pairs a question about Wikidata facts with an LLM-written passage in which the answer entity is replaced, so the passage contradicts what the model remembers.
  We build text contrast sets~from~it~\mbox{(Section~\ref{sec:exp-general})}.
  \item \textbf{NQ-Swap}~\cite{longpre2021entity} and \textbf{FaithEval}~\cite{ming2025faitheval} are used only for testing.
  NQ-Swap replaces the answer entity in Wikipedia passages of Natural Questions, and FaithEval pairs multiple-choice science questions with passages that state counterfactual facts.
\end{itemize}

\section{Evaluation protocol}
\label{sec:supp-eval}

\paragraph{Budgets.}
At a budget of $x$\,MP, the first view is the photo resized to at most $x$\,MP with its aspect ratio kept, and every crop is also resized to at most $x$\,MP.
A model's maximum is the largest input it accepts: 16.8\,MP for Qwen3.5, Qwen3.8 and Qwen3-VL and the models trained from them, 12.8\,MP for models based on Qwen2.5-VL, 3.8\,MP for Claude Haiku~5.5 and 30.7\,MP for GPT-6-Luna.
Budgets above a model's maximum use the maximum.

\paragraph{Zoom tools.}
Every model in our study zooms by drawing a box, and the box is mapped to the full-resolution photo.
The tools differ in where the box may be drawn, in how the crop is resized and in how many calls are allowed.
ViRA-9B and ViRA-27B use our zoom tool (Appendix~\ref{sec:supp-tool}).
It accepts a box on the photo or on an earlier crop, cuts the region from the full-resolution photo and enlarges it by up to $4\times$ within the budget, so a small object fills more of the returned image.
A model may call it up to six times.
The proprietary models call the same tool through the function calling of their APIs, with one difference.
Their crops are shrunk to fit the budget but not enlarged.
The \emph{+\,tool} baselines (Qwen3.5-9B, ZwZ2-9B, Vision-OPD-9B, Qwen3.8-27B and ZwZ2-27B) use a zoom tool in the format of DeepEyes~\cite{zheng2025deepeyes}.
It accepts a box on the photo and returns the crop resized to the budget without magnification.
A model may call it up to six times, and each reply has at most 2,048 tokens.
The released agents use the tools, prompts and turn limits they were trained with: DeepEyes, DeepEyesV2, Thyme, Pixel Reasoner and Mini-o3 (32 turns) call their own crop or code tools, and ZoomEye~\cite{shen2025zoomeye} searches a tree of crops without training.
LBS-27B~\cite{yao2026lbs} crops with a text action, at most three times, at its highest reasoning effort.
SenseNova-MARS-32B~\cite{chng2025sensenovamars} has a crop tool and two search tools.
We run it offline, so its searches return no results.
It may call tools for six turns and must then answer, although its own evaluation allows ten.

\paragraph{Proprietary models.}
We query the proprietary models of Table~\ref{tab:models} through their APIs.
Claude Haiku~5.5 and GPT-6-Luna run with our zoom tool at the highest, medium and no thinking level, which are the reasoning efforts \texttt{max}, \texttt{medium} and \texttt{none} of the API.
All calls use the default temperature of each API.
The system prompt is ``You are a helpful assistant.'', and the user turn holds the first view and the question with its options, followed by the instruction of Listing~\ref{lst:turns}.
Each model draws boxes in the format it was trained with.
Claude Haiku~5.5 gives the box in pixels of the image named by \texttt{source}, and the turn that holds an image also states its size.
GPT-6-Luna gives $[x_1,y_1,x_2,y_2]$ on a 0--1000 scale.
Each crop follows in the next user turn as an image named \texttt{observation\_}$k$.
A reply may have up to 16,384 tokens, and up to 32,768 at the highest thinking level.

\begin{table}[htbp]
\caption{\textbf{Evaluated models}: checkpoints, API model names and training data.
Training data: the number of samples that train a model from its base model, as its paper reports it. $a+b$: supervised fine-tuning and RL. $\geq$: only the supervised stage is reported.}
\label{tab:models}
\centering
\scriptsize
\resizebox{\columnwidth}{!}{%
\begin{tabular}{@{}llr@{}}
\toprule
\midrule
Model & Checkpoint or API model & Training data \\
\midrule
\multicolumn{3}{l}{\emph{Ours}} \\
ViRA-9B & trained from \texttt{Qwen/Qwen3.5-9B} & 3,642 \\
ViRA-27B & trained from \texttt{Qwen/Qwen3.8-27B} & 3,401 \\
\midrule[0.3pt]
\multicolumn{3}{l}{\emph{Open models}} \\
Qwen3.5-9B & \texttt{Qwen/Qwen3.5-9B}& -- \\
ZwZ2-9B & \texttt{inclusionAI/ZwZ2-9B}& 20.4K \\
Vision-OPD-9B & \texttt{yuanqianhao/Vision-OPD-9B}& 6.2K \\
Qwen3.8-27B & \texttt{Qwen/Qwen3.8-27B}& -- \\
ZwZ2-27B & \texttt{inclusionAI/ZwZ2-27B}& 20.4K \\
\midrule[0.3pt]
\multicolumn{3}{l}{\emph{Released zoom agents}} \\
Mini-o3 & \texttt{Mini-o3/Mini-o3-7B-v1}& 6K + 12K \\
DeepEyes & \texttt{ChenShawn/DeepEyes-7B}& 47K \\
DeepEyesV2 & \texttt{honglyhly/DeepEyesV2\_7B\_1031}& not reported \\
Thyme & \texttt{Kwai-Keye/Thyme-RL}& $\geq$500K \\
Pixel Reasoner & \texttt{TIGER-Lab/PixelReasoner-RL-v1}& 7.5K + 15K \\
ZoomEye & search with \texttt{Qwen/Qwen2.5-VL-7B-Instruct}& none \\
LBS-27B & \texttt{inclusionAI/LBS-27B}& $\geq$10K \\
SenseNova-MARS-32B & \texttt{sensenova/SenseNova-MARS-32B}& 13.4K \\
\midrule[0.3pt]
\multicolumn{3}{l}{\emph{Proprietary models}} \\
Claude Haiku~5.5 & \texttt{anthropic/claude-haiku-5.5}& -- \\
GPT-6-Luna & \texttt{openai/gpt-6-luna}& -- \\
\midrule[0.3pt]
\multicolumn{3}{l}{\emph{Judge of open answers}} \\
GPT-4o & \texttt{gpt-4o-2024-11-20}& -- \\
\midrule
\bottomrule
\end{tabular}}
\end{table}

\paragraph{Open models and agents.}
Table~\ref{tab:models} lists the checkpoints of all models.
Without a zoom tool, a model answers from the first view.
Decoding is greedy.

\paragraph{Scoring.}
We score the last answer of each rollout.
On ViRA-Bench and ViRA-Bench-Real, a set counts as solved if all three versions are answered correctly, and per-image accuracy counts the versions separately.
Tokens per question are the prompt tokens of all turns plus the generated tokens, summed over the three versions.
For the proprietary models, the generated tokens include the hidden reasoning tokens, which the APIs count as output tokens.
On ViRA-Bench, ViRA-Bench-Real, ViRA and the ablations are evaluated three times, and we report the mean.
The other models are run once.
The three runs differ slightly from one another despite greedy decoding, because inference is~not~\mbox{deterministic}.

\section{Main results}
\label{sec:supp-full}

\begin{figure*}[htbp]
  \centering
  \includegraphics[width=\textwidth]{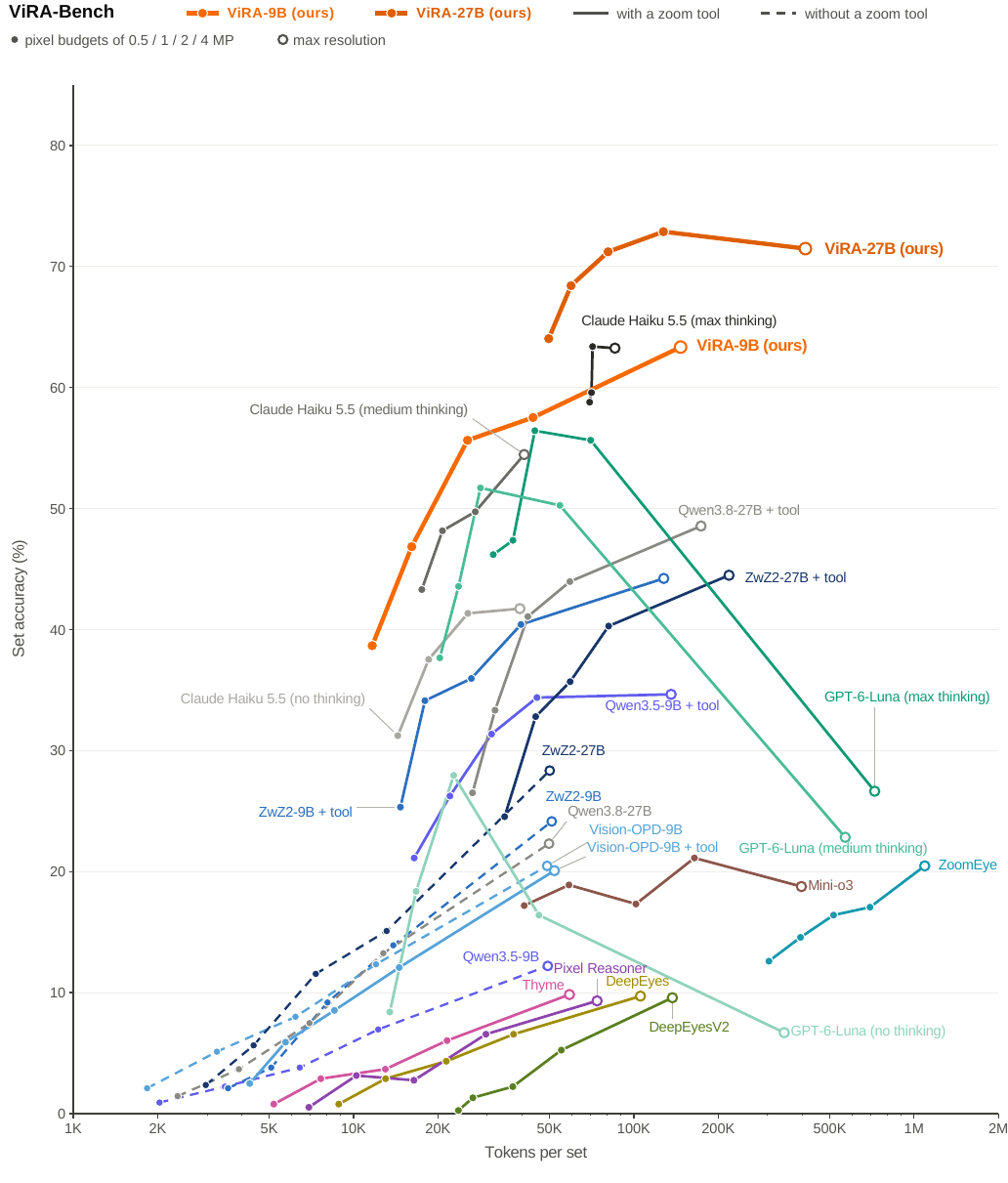}
  \caption{\textbf{Set accuracy vs.\ tokens for every model on ViRA-Bench.}
  Each model is labeled at its last marker, and the labels of proprietary models give the thinking level.
  Filled markers are pixel budgets of 0.5--4\,MP, hollow markers are each model's max resolution, and dashed lines are open models without a zoom tool.
  Claude Haiku~5.5 has no 4\,MP marker because its max resolution is 3.8\,MP.
  Tokens are summed over all turns and versions of a set.}
  \label{fig:token-acc-bench}
\end{figure*}

\begin{figure*}[htbp]
  \centering
  \includegraphics[width=\textwidth]{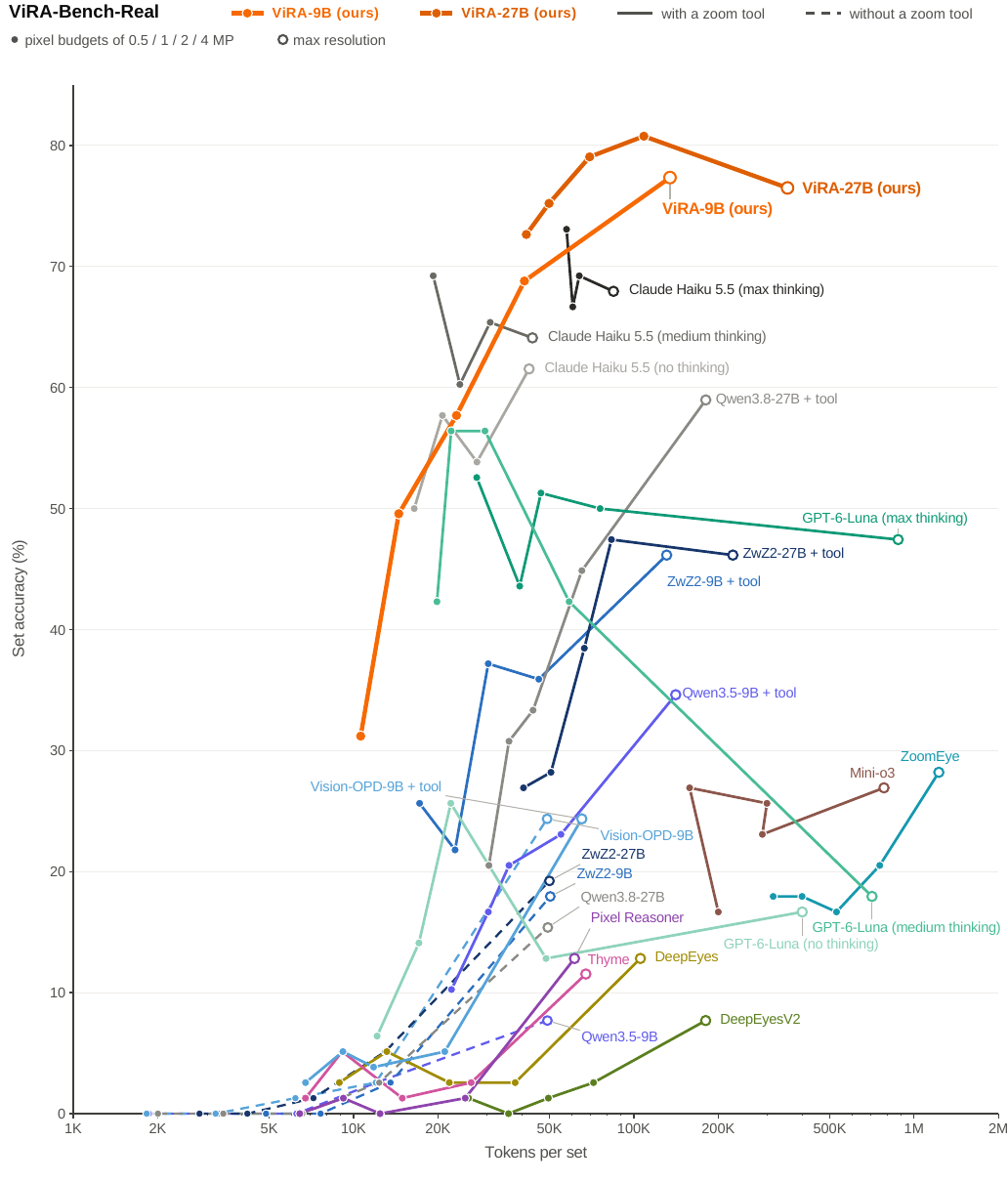}
  \caption{\textbf{Set accuracy vs.\ tokens for every model on ViRA-Bench-Real.}
  Each model is labeled at its last marker, and the labels of proprietary models give the thinking level.
  Filled markers are pixel budgets of 0.5--4\,MP, hollow markers are each model's max resolution, and dashed lines are open models without a zoom tool.
  Claude Haiku~5.5 has no 4\,MP marker because its max resolution is 3.8\,MP.
  Tokens are summed over all turns and versions of a set.}
  \label{fig:token-acc-real}
\end{figure*}

Figures~\ref{fig:token-acc-bench} and~\ref{fig:token-acc-real} show the set accuracy and the tokens per set of every model on the~two~\mbox{benchmarks}.

Table~\ref{tab:image-acc} gives the per-image accuracy of every model on the two benchmarks, which counts each version of a set separately.

\begin{table*}[htbp]
\caption{\textbf{Per-image accuracy (\%), all models}, on ViRA-Bench and ViRA-Bench-Real. Each version of a set counts separately. The models and the budgets are as in Table~\ref{tab:main}. ViRA results are means of three runs. Bold: best in each column within each group.}
\label{tab:image-acc}
\centering
\scriptsize
\setlength{\tabcolsep}{7.5pt}
\begin{tabular}{l*{10}{c}}
\toprule
\midrule
 & \multicolumn{5}{c}{ViRA-Bench} & \multicolumn{5}{c}{ViRA-Bench-Real} \\
\cmidrule(lr){2-6}\cmidrule(l){7-11}
Model & 0.5\,MP & 1\,MP & 2\,MP & 4\,MP & Max & 0.5\,MP & 1\,MP & 2\,MP & 4\,MP & Max \\
\midrule
\multicolumn{11}{l}{\emph{Open 7--9B models}} \\
Qwen3.5-9B & 31.3 & 33.8 & 37.0 & 39.9 & 46.2 & 28.6 & 29.5 & 32.5 & 35.0 & 45.7 \\
ZwZ2-9B & 30.7 & 34.3 & 39.6 & 44.2 & 54.3 & 27.8 & 27.8 & 29.9 & 35.5 & 52.6 \\
Vision-OPD-9B & 32.6 & 36.3 & 40.0 & 44.2 & 52.7 & 27.8 & 28.2 & 32.5 & 35.9 & 56.4 \\
\midrule[0.3pt]
\multicolumn{11}{l}{\emph{Open 7--9B models + tool}} \\
Qwen3.5-9B + tool & 50.8 & 55.2 & 58.3 & 61.8 & 62.9 & 38.5 & 47.0 & 54.7 & 56.8 & 64.5 \\
ZwZ2-9B + tool & 55.6 & 61.7 & 64.5 & 67.2 & 70.0 & 56.4 & 57.7 & 67.9 & 67.1 & 72.6 \\
Vision-OPD-9B + tool & 32.5 & 35.7 & 40.5 & 44.1 & 51.4 & 34.6 & 35.0 & 37.2 & 40.6 & 56.8 \\
\midrule[0.3pt]
\multicolumn{11}{l}{\emph{Released zoom agents (7--9B)}} \\
Mini-o3 & 44.8 & 48.1 & 48.0 & 50.7 & 47.8 & 54.3 & 59.0 & 56.0 & 58.1 & 55.1 \\
DeepEyes & 32.2 & 35.9 & 35.2 & 38.8 & 42.2 & 34.6 & 36.3 & 35.0 & 38.0 & 45.7 \\
DeepEyesV2 & 29.7 & 32.3 & 34.2 & 38.8 & 42.7 & 29.9 & 29.5 & 32.1 & 34.2 & 44.4 \\
Thyme & 31.5 & 34.5 & 35.4 & 37.8 & 42.9 & 30.8 & 36.8 & 34.2 & 37.6 & 41.0 \\
Pixel Reasoner & 31.4 & 35.0 & 35.4 & 39.3 & 43.1 & 30.3 & 31.6 & 32.9 & 34.6 & 45.3 \\
ZoomEye & 43.5 & 45.8 & 46.4 & 48.4 & 50.9 & 49.6 & 50.9 & 50.9 & 53.4 & 59.8 \\
\midrule[0.3pt]
\multicolumn{11}{l}{\emph{Ours + our zoom tool}} \\
\shaderow{vira9bg}\textbf{ViRA-9B} (ours) & \textbf{65.1} & \textbf{71.1} & \textbf{75.8} & \textbf{78.2} & \textbf{81.3} & \textbf{61.5} & \textbf{73.1} & \textbf{79.6} & \textbf{86.8} & \textbf{89.9} \\
\midrule[1pt]
\multicolumn{11}{l}{\emph{Open 27B models}} \\
Qwen3.8-27B & 33.6 & 36.6 & 41.0 & 45.9 & 54.5 & 32.1 & 31.6 & 31.2 & 36.8 & 49.6 \\
ZwZ2-27B & 33.8 & 38.5 & 44.1 & 48.6 & 59.6 & 32.5 & 31.2 & 35.9 & 38.0 & 52.6 \\
\midrule[0.3pt]
\multicolumn{11}{l}{\emph{Open 27B models + tool}} \\
Qwen3.8-27B + tool & 45.5 & 55.0 & 62.4 & 66.2 & 69.4 & 45.3 & 52.1 & 57.3 & 62.4 & 74.8 \\
ZwZ2-27B + tool & 48.8 & 54.8 & 59.7 & 64.8 & 67.2 & 50.9 & 55.6 & 61.1 & 68.8 & 71.8 \\
\midrule[0.3pt]
\multicolumn{11}{l}{\emph{Released zoom agents (27--32B)}} \\
LBS-27B & 70.1 & 76.1 & 79.1 & 82.9 & 85.7 & 73.5 & 75.6 & 83.8 & 85.9 & 88.5 \\
SenseNova-MARS-32B & 57.0 & 62.5 & 65.7 & 66.7 & 68.5 & 56.4 & 64.1 & 69.2 & 67.5 & 71.4 \\
\midrule[0.3pt]
\multicolumn{11}{l}{\emph{Ours + our zoom tool}} \\
\shaderow{vira27bg}\textbf{ViRA-27B} (ours) & \textbf{82.0} & \textbf{84.7} & \textbf{86.7} & \textbf{87.3} & \textbf{86.8} & \textbf{87.9} & \textbf{89.9} & \textbf{91.9} & \textbf{92.2} & \textbf{90.5} \\
\midrule[1pt]
\multicolumn{11}{l}{\emph{Proprietary models + our zoom tool}} \\
Claude Haiku~5.5, max thinking & \textbf{78.2} & \textbf{79.1} & \textbf{81.5} & \textbf{82.2} & \textbf{82.2} & \textbf{86.3} & \textbf{85.0} & \textbf{84.6} & \textbf{84.2} & \textbf{84.2} \\
Claude Haiku~5.5, medium thinking & 70.2 & 71.9 & 73.9 & 77.4 & 77.4 & 85.0 & 82.5 & 84.2 & 83.3 & 83.3 \\
Claude Haiku~5.5, no thinking & 61.0 & 64.1 & 68.3 & 68.6 & 68.6 & 73.9 & 81.2 & 76.5 & 81.2 & 81.2 \\
GPT-6-Luna, max thinking & 71.1 & 73.0 & 78.1 & 78.3 & 63.6 & 73.5 & 71.4 & 75.2 & 71.8 & 70.9 \\
GPT-6-Luna, medium thinking & 67.7 & 71.4 & 76.8 & 75.3 & 58.8 & 72.6 & 77.8 & 76.5 & 72.6 & 43.6 \\
GPT-6-Luna, no thinking & 42.2 & 50.4 & 59.1 & 52.3 & 42.3 & 47.4 & 52.1 & 61.5 & 50.0 & 49.6 \\
\midrule
\bottomrule
\end{tabular}
\end{table*}

\section{Ablations}
\label{sec:supp-ablation}

Table~\ref{tab:ablation} in Section~\ref{sec:exp-ablation} gives the accuracy of the ablations.
Table~\ref{tab:ablation-bench} adds their per-image accuracy and tokens per set on ViRA-Bench and ViRA-Bench-Real.
Table~\ref{tab:ablation-mmerw} splits MME-RealWorld-Lite by task and domain.
The tables of this appendix also contain three runs that are not in Table~\ref{tab:ablation}.
Two of them are per-version GRPO with the cross-input weight, which multiplies the advantage of per-version GRPO by the factor $1+\lambda w_v$ of CrossRPO, for $\lambda=0.5$ and $\lambda=1$ (Appendix~\ref{sec:supp-ablation-joint}).
The third run trains on one version of each contrast set with the advantage of $\lambda=0$ (Appendix~\ref{sec:supp-ablation-onever}).
On MME-RealWorld-Lite, CrossRPO is at least as accurate as every ablation of Table~\ref{tab:ablation} on the perception questions, on the reasoning questions and overall, at both resolutions.
Table~\ref{tab:ablation-reading} gives the columns of Table~\ref{tab:reading-full} for all ablations.
With random partners, with $\lambda=2$ and with per-version GRPO, the agent zooms 2.5, 2.7 and 2.5 times per image, and its last zoom covers the object less often (39.5\%, 61.0\% and 52.1\%).

\begin{table*}[htbp]
\caption{\textbf{Ablations on ViRA-Bench and ViRA-Bench-Real}: set accuracy with per-image accuracy in parentheses (\%), and tokens per set in gray.
All results are means of three runs.
The rows are the variants of Table~\ref{tab:ablation-settings}, which Appendix~\ref{sec:supp-ablation-settings} describes.
The rows under \emph{Per-version GRPO $\times\,(1+\lambda w_v)$} multiply the advantage of per-version GRPO by this factor (Appendix~\ref{sec:supp-ablation-joint}).
Appendix~\ref{sec:supp-ablation-onever} describes the row \emph{$\lambda=0$, one version per set}.}
\label{tab:ablation-bench}
\centering
\scriptsize
\setlength{\tabcolsep}{15.80pt}
\settowidth{\accw}{\textbf{00.0}}\settowidth{\tokw}{\tok{000K}}\settowidth{\numw}{\textbf{00.0}}\settowidth{\imgw}{(00.0)}
\begin{tabular}{l*{5}{c}}
\toprule
\midrule
Advantage & 0.5\,MP & 1\,MP & 2\,MP & 4\,MP & Max \\
\midrule
\multicolumn{6}{c}{ViRA-Bench} \\
\midrule[0.3pt]
\shaderow{vira9bg}\textbf{CrossRPO} ($\lambda=0.5$) & \makebox[\accw][r]{38.7}\,\makebox[\imgw][r]{(65.1)}\makebox[\tokw][l]{\tok{12K}} & \makebox[\accw][r]{46.9}\,\makebox[\imgw][r]{(71.1)}\makebox[\tokw][l]{\tok{16K}} & \makebox[\accw][r]{\textbf{55.6}}\,\makebox[\imgw][r]{(75.8)}\makebox[\tokw][l]{\tok{26K}} & \makebox[\accw][r]{\textbf{57.5}}\,\makebox[\imgw][r]{(78.2)}\makebox[\tokw][l]{\tok{44K}} & \makebox[\accw][r]{\textbf{63.3}}\,\makebox[\imgw][r]{(81.3)}\makebox[\tokw][l]{\tok{147K}} \\
\midrule[0.3pt]
\multicolumn{6}{l}{\emph{Joint term}} \\
$\lambda=0$ (per-version RLOO) & \makebox[\accw][r]{31.3}\,\makebox[\imgw][r]{(59.7)}\makebox[\tokw][l]{\tok{11K}} & \makebox[\accw][r]{41.2}\,\makebox[\imgw][r]{(66.8)}\makebox[\tokw][l]{\tok{15K}} & \makebox[\accw][r]{50.0}\,\makebox[\imgw][r]{(72.5)}\makebox[\tokw][l]{\tok{24K}} & \makebox[\accw][r]{52.9}\,\makebox[\imgw][r]{(74.7)}\makebox[\tokw][l]{\tok{42K}} & \makebox[\accw][r]{58.7}\,\makebox[\imgw][r]{(78.6)}\makebox[\tokw][l]{\tok{144K}} \\
$\lambda=0$, one version per set & \makebox[\accw][r]{31.1}\,\makebox[\imgw][r]{(59.3)}\makebox[\tokw][l]{\tok{10K}} & \makebox[\accw][r]{40.5}\,\makebox[\imgw][r]{(67.3)}\makebox[\tokw][l]{\tok{15K}} & \makebox[\accw][r]{49.3}\,\makebox[\imgw][r]{(72.5)}\makebox[\tokw][l]{\tok{23K}} & \makebox[\accw][r]{53.6}\,\makebox[\imgw][r]{(74.7)}\makebox[\tokw][l]{\tok{41K}} & \makebox[\accw][r]{58.1}\,\makebox[\imgw][r]{(78.0)}\makebox[\tokw][l]{\tok{147K}} \\
$\lambda=1$ & \makebox[\accw][r]{33.4}\,\makebox[\imgw][r]{(61.4)}\makebox[\tokw][l]{\tok{16K}} & \makebox[\accw][r]{44.2}\,\makebox[\imgw][r]{(68.9)}\makebox[\tokw][l]{\tok{22K}} & \makebox[\accw][r]{51.3}\,\makebox[\imgw][r]{(73.7)}\makebox[\tokw][l]{\tok{35K}} & \makebox[\accw][r]{56.1}\,\makebox[\imgw][r]{(76.5)}\makebox[\tokw][l]{\tok{60K}} & \makebox[\accw][r]{56.6}\,\makebox[\imgw][r]{(78.1)}\makebox[\tokw][l]{\tok{192K}} \\
$\lambda=2$ & \makebox[\accw][r]{\underline{40.9}}\,\makebox[\imgw][r]{(65.8)}\makebox[\tokw][l]{\tok{34K}} & \makebox[\accw][r]{\underline{47.9}}\,\makebox[\imgw][r]{(71.4)}\makebox[\tokw][l]{\tok{47K}} & \makebox[\accw][r]{\underline{53.8}}\,\makebox[\imgw][r]{(74.7)}\makebox[\tokw][l]{\tok{75K}} & \makebox[\accw][r]{55.7}\,\makebox[\imgw][r]{(76.2)}\makebox[\tokw][l]{\tok{125K}} & \makebox[\accw][r]{58.2}\,\makebox[\imgw][r]{(77.6)}\makebox[\tokw][l]{\tok{346K}} \\
Random partners & \makebox[\accw][r]{33.8}\,\makebox[\imgw][r]{(61.8)}\makebox[\tokw][l]{\tok{33K}} & \makebox[\accw][r]{42.1}\,\makebox[\imgw][r]{(67.6)}\makebox[\tokw][l]{\tok{43K}} & \makebox[\accw][r]{47.5}\,\makebox[\imgw][r]{(71.2)}\makebox[\tokw][l]{\tok{62K}} & \makebox[\accw][r]{51.7}\,\makebox[\imgw][r]{(74.6)}\makebox[\tokw][l]{\tok{94K}} & \makebox[\accw][r]{57.3}\,\makebox[\imgw][r]{(78.5)}\makebox[\tokw][l]{\tok{282K}} \\
\midrule[0.3pt]
\multicolumn{6}{l}{\emph{Per-version GRPO $\times\,(1+\lambda w_v)$}} \\
GRPO, $\lambda=0.5$ & \makebox[\accw][r]{31.2}\,\makebox[\imgw][r]{(59.3)}\makebox[\tokw][l]{\tok{26K}} & \makebox[\accw][r]{39.0}\,\makebox[\imgw][r]{(66.6)}\makebox[\tokw][l]{\tok{38K}} & \makebox[\accw][r]{47.1}\,\makebox[\imgw][r]{(71.2)}\makebox[\tokw][l]{\tok{60K}} & \makebox[\accw][r]{52.2}\,\makebox[\imgw][r]{(73.8)}\makebox[\tokw][l]{\tok{95K}} & \makebox[\accw][r]{54.9}\,\makebox[\imgw][r]{(76.5)}\makebox[\tokw][l]{\tok{313K}} \\
GRPO, $\lambda=1$ & \makebox[\accw][r]{37.7}\,\makebox[\imgw][r]{(63.3)}\makebox[\tokw][l]{\tok{10K}} & \makebox[\accw][r]{46.3}\,\makebox[\imgw][r]{(69.5)}\makebox[\tokw][l]{\tok{14K}} & \makebox[\accw][r]{51.6}\,\makebox[\imgw][r]{(73.5)}\makebox[\tokw][l]{\tok{23K}} & \makebox[\accw][r]{56.1}\,\makebox[\imgw][r]{(76.0)}\makebox[\tokw][l]{\tok{40K}} & \makebox[\accw][r]{58.9}\,\makebox[\imgw][r]{(78.3)}\makebox[\tokw][l]{\tok{137K}} \\
\midrule[0.3pt]
\multicolumn{6}{l}{\emph{Existing RL methods}} \\
Per-version GRPO & \makebox[\accw][r]{\textbf{44.0}}\,\makebox[\imgw][r]{(67.4)}\makebox[\tokw][l]{\tok{27K}} & \makebox[\accw][r]{\textbf{50.2}}\,\makebox[\imgw][r]{(72.0)}\makebox[\tokw][l]{\tok{37K}} & \makebox[\accw][r]{53.4}\,\makebox[\imgw][r]{(74.4)}\makebox[\tokw][l]{\tok{56K}} & \makebox[\accw][r]{\underline{56.3}}\,\makebox[\imgw][r]{(76.7)}\makebox[\tokw][l]{\tok{87K}} & \makebox[\accw][r]{59.3}\,\makebox[\imgw][r]{(79.2)}\makebox[\tokw][l]{\tok{233K}} \\
Pooled GRPO & \makebox[\accw][r]{38.0}\,\makebox[\imgw][r]{(63.8)}\makebox[\tokw][l]{\tok{18K}} & \makebox[\accw][r]{44.9}\,\makebox[\imgw][r]{(69.9)}\makebox[\tokw][l]{\tok{22K}} & \makebox[\accw][r]{51.6}\,\makebox[\imgw][r]{(74.4)}\makebox[\tokw][l]{\tok{33K}} & \makebox[\accw][r]{53.9}\,\makebox[\imgw][r]{(75.6)}\makebox[\tokw][l]{\tok{53K}} & \makebox[\accw][r]{52.6}\,\makebox[\imgw][r]{(75.4)}\makebox[\tokw][l]{\tok{157K}} \\
Pooled RLOO & \makebox[\accw][r]{37.2}\,\makebox[\imgw][r]{(61.8)}\makebox[\tokw][l]{\tok{12K}} & \makebox[\accw][r]{43.7}\,\makebox[\imgw][r]{(68.3)}\makebox[\tokw][l]{\tok{17K}} & \makebox[\accw][r]{52.2}\,\makebox[\imgw][r]{(73.7)}\makebox[\tokw][l]{\tok{26K}} & \makebox[\accw][r]{55.7}\,\makebox[\imgw][r]{(76.3)}\makebox[\tokw][l]{\tok{42K}} & \makebox[\accw][r]{\underline{61.0}}\,\makebox[\imgw][r]{(79.8)}\makebox[\tokw][l]{\tok{133K}} \\
ConsistRoll~\cite{zou2026consistency} & \makebox[\accw][r]{23.5}\,\makebox[\imgw][r]{(54.3)}\makebox[\tokw][l]{\tok{24K}} & \makebox[\accw][r]{33.4}\,\makebox[\imgw][r]{(61.6)}\makebox[\tokw][l]{\tok{32K}} & \makebox[\accw][r]{40.8}\,\makebox[\imgw][r]{(67.0)}\makebox[\tokw][l]{\tok{48K}} & \makebox[\accw][r]{47.1}\,\makebox[\imgw][r]{(70.6)}\makebox[\tokw][l]{\tok{79K}} & \makebox[\accw][r]{53.1}\,\makebox[\imgw][r]{(74.8)}\makebox[\tokw][l]{\tok{274K}} \\
\midrule[1pt]
\multicolumn{6}{c}{ViRA-Bench-Real} \\
\midrule[0.3pt]
\shaderow{vira9bg}\textbf{CrossRPO} ($\lambda=0.5$) & \makebox[\accw][r]{31.2}\,\makebox[\imgw][r]{(61.5)}\makebox[\tokw][l]{\tok{11K}} & \makebox[\accw][r]{49.6}\,\makebox[\imgw][r]{(73.1)}\makebox[\tokw][l]{\tok{15K}} & \makebox[\accw][r]{57.7}\,\makebox[\imgw][r]{(79.6)}\makebox[\tokw][l]{\tok{23K}} & \makebox[\accw][r]{68.8}\,\makebox[\imgw][r]{(86.8)}\makebox[\tokw][l]{\tok{41K}} & \makebox[\accw][r]{\textbf{77.4}}\,\makebox[\imgw][r]{(89.9)}\makebox[\tokw][l]{\tok{135K}} \\
\midrule[0.3pt]
\multicolumn{6}{l}{\emph{Joint term}} \\
$\lambda=0$ (per-version RLOO) & \makebox[\accw][r]{23.5}\,\makebox[\imgw][r]{(57.3)}\makebox[\tokw][l]{\tok{10K}} & \makebox[\accw][r]{41.9}\,\makebox[\imgw][r]{(69.8)}\makebox[\tokw][l]{\tok{14K}} & \makebox[\accw][r]{56.8}\,\makebox[\imgw][r]{(77.6)}\makebox[\tokw][l]{\tok{23K}} & \makebox[\accw][r]{65.4}\,\makebox[\imgw][r]{(84.5)}\makebox[\tokw][l]{\tok{40K}} & \makebox[\accw][r]{70.1}\,\makebox[\imgw][r]{(87.3)}\makebox[\tokw][l]{\tok{141K}} \\
$\lambda=0$, one version per set & \makebox[\accw][r]{20.5}\,\makebox[\imgw][r]{(54.6)}\makebox[\tokw][l]{\tok{10K}} & \makebox[\accw][r]{40.2}\,\makebox[\imgw][r]{(66.5)}\makebox[\tokw][l]{\tok{14K}} & \makebox[\accw][r]{58.1}\,\makebox[\imgw][r]{(79.3)}\makebox[\tokw][l]{\tok{23K}} & \makebox[\accw][r]{64.1}\,\makebox[\imgw][r]{(82.1)}\makebox[\tokw][l]{\tok{40K}} & \makebox[\accw][r]{67.1}\,\makebox[\imgw][r]{(83.5)}\makebox[\tokw][l]{\tok{147K}} \\
$\lambda=1$ & \makebox[\accw][r]{32.9}\,\makebox[\imgw][r]{(61.1)}\makebox[\tokw][l]{\tok{17K}} & \makebox[\accw][r]{49.6}\,\makebox[\imgw][r]{(76.2)}\makebox[\tokw][l]{\tok{23K}} & \makebox[\accw][r]{59.0}\,\makebox[\imgw][r]{(79.8)}\makebox[\tokw][l]{\tok{35K}} & \makebox[\accw][r]{67.1}\,\makebox[\imgw][r]{(83.9)}\makebox[\tokw][l]{\tok{60K}} & \makebox[\accw][r]{69.2}\,\makebox[\imgw][r]{(86.6)}\makebox[\tokw][l]{\tok{189K}} \\
$\lambda=2$ & \makebox[\accw][r]{\underline{52.1}}\,\makebox[\imgw][r]{(74.5)}\makebox[\tokw][l]{\tok{26K}} & \makebox[\accw][r]{\underline{59.0}}\,\makebox[\imgw][r]{(80.1)}\makebox[\tokw][l]{\tok{38K}} & \makebox[\accw][r]{\underline{68.8}}\,\makebox[\imgw][r]{(86.2)}\makebox[\tokw][l]{\tok{59K}} & \makebox[\accw][r]{69.7}\,\makebox[\imgw][r]{(85.9)}\makebox[\tokw][l]{\tok{97K}} & \makebox[\accw][r]{74.8}\,\makebox[\imgw][r]{(89.7)}\makebox[\tokw][l]{\tok{251K}} \\
Random partners & \makebox[\accw][r]{32.5}\,\makebox[\imgw][r]{(63.7)}\makebox[\tokw][l]{\tok{33K}} & \makebox[\accw][r]{48.3}\,\makebox[\imgw][r]{(73.2)}\makebox[\tokw][l]{\tok{44K}} & \makebox[\accw][r]{52.1}\,\makebox[\imgw][r]{(79.1)}\makebox[\tokw][l]{\tok{57K}} & \makebox[\accw][r]{65.0}\,\makebox[\imgw][r]{(84.0)}\makebox[\tokw][l]{\tok{85K}} & \makebox[\accw][r]{71.4}\,\makebox[\imgw][r]{(88.9)}\makebox[\tokw][l]{\tok{247K}} \\
\midrule[0.3pt]
\multicolumn{6}{l}{\emph{Per-version GRPO $\times\,(1+\lambda w_v)$}} \\
GRPO, $\lambda=0.5$ & \makebox[\accw][r]{32.9}\,\makebox[\imgw][r]{(60.5)}\makebox[\tokw][l]{\tok{27K}} & \makebox[\accw][r]{47.0}\,\makebox[\imgw][r]{(73.4)}\makebox[\tokw][l]{\tok{38K}} & \makebox[\accw][r]{60.3}\,\makebox[\imgw][r]{(81.5)}\makebox[\tokw][l]{\tok{62K}} & \makebox[\accw][r]{64.1}\,\makebox[\imgw][r]{(82.9)}\makebox[\tokw][l]{\tok{103K}} & \makebox[\accw][r]{65.0}\,\makebox[\imgw][r]{(84.2)}\makebox[\tokw][l]{\tok{308K}} \\
GRPO, $\lambda=1$ & \makebox[\accw][r]{32.9}\,\makebox[\imgw][r]{(61.3)}\makebox[\tokw][l]{\tok{10K}} & \makebox[\accw][r]{58.5}\,\makebox[\imgw][r]{(78.3)}\makebox[\tokw][l]{\tok{14K}} & \makebox[\accw][r]{65.0}\,\makebox[\imgw][r]{(84.0)}\makebox[\tokw][l]{\tok{23K}} & \makebox[\accw][r]{68.4}\,\makebox[\imgw][r]{(86.0)}\makebox[\tokw][l]{\tok{40K}} & \makebox[\accw][r]{71.4}\,\makebox[\imgw][r]{(87.7)}\makebox[\tokw][l]{\tok{136K}} \\
\midrule[0.3pt]
\multicolumn{6}{l}{\emph{Existing RL methods}} \\
Per-version GRPO & \makebox[\accw][r]{\textbf{55.6}}\,\makebox[\imgw][r]{(77.9)}\makebox[\tokw][l]{\tok{27K}} & \makebox[\accw][r]{\textbf{66.2}}\,\makebox[\imgw][r]{(84.5)}\makebox[\tokw][l]{\tok{35K}} & \makebox[\accw][r]{\textbf{74.4}}\,\makebox[\imgw][r]{(89.3)}\makebox[\tokw][l]{\tok{50K}} & \makebox[\accw][r]{\textbf{73.5}}\,\makebox[\imgw][r]{(88.6)}\makebox[\tokw][l]{\tok{82K}} & \makebox[\accw][r]{\underline{75.6}}\,\makebox[\imgw][r]{(90.7)}\makebox[\tokw][l]{\tok{190K}} \\
Pooled GRPO & \makebox[\accw][r]{44.9}\,\makebox[\imgw][r]{(71.2)}\makebox[\tokw][l]{\tok{16K}} & \makebox[\accw][r]{53.8}\,\makebox[\imgw][r]{(76.4)}\makebox[\tokw][l]{\tok{20K}} & \makebox[\accw][r]{59.8}\,\makebox[\imgw][r]{(81.1)}\makebox[\tokw][l]{\tok{29K}} & \makebox[\accw][r]{65.4}\,\makebox[\imgw][r]{(85.5)}\makebox[\tokw][l]{\tok{47K}} & \makebox[\accw][r]{70.1}\,\makebox[\imgw][r]{(88.2)}\makebox[\tokw][l]{\tok{135K}} \\
Pooled RLOO & \makebox[\accw][r]{44.9}\,\makebox[\imgw][r]{(70.2)}\makebox[\tokw][l]{\tok{14K}} & \makebox[\accw][r]{58.1}\,\makebox[\imgw][r]{(78.2)}\makebox[\tokw][l]{\tok{18K}} & \makebox[\accw][r]{68.4}\,\makebox[\imgw][r]{(82.8)}\makebox[\tokw][l]{\tok{24K}} & \makebox[\accw][r]{\underline{70.9}}\,\makebox[\imgw][r]{(86.9)}\makebox[\tokw][l]{\tok{41K}} & \makebox[\accw][r]{\underline{75.6}}\,\makebox[\imgw][r]{(89.3)}\makebox[\tokw][l]{\tok{115K}} \\
ConsistRoll~\cite{zou2026consistency} & \makebox[\accw][r]{16.2}\,\makebox[\imgw][r]{(48.6)}\makebox[\tokw][l]{\tok{27K}} & \makebox[\accw][r]{29.9}\,\makebox[\imgw][r]{(59.1)}\makebox[\tokw][l]{\tok{33K}} & \makebox[\accw][r]{49.6}\,\makebox[\imgw][r]{(72.5)}\makebox[\tokw][l]{\tok{49K}} & \makebox[\accw][r]{56.0}\,\makebox[\imgw][r]{(78.9)}\makebox[\tokw][l]{\tok{82K}} & \makebox[\accw][r]{61.5}\,\makebox[\imgw][r]{(83.2)}\makebox[\tokw][l]{\tok{269K}} \\
\midrule
\bottomrule
\end{tabular}

\end{table*}
\begin{table*}[htbp]
\caption{\textbf{Ablations on MME-RealWorld-Lite by task and domain}: accuracy (\%) at the maximum resolution and at 1\,MP.
The columns are those of Table~\ref{tab:mmerw-max}, and the rows are those of Table~\ref{tab:ablation-bench}.}
\label{tab:ablation-mmerw}
\centering
\scriptsize
\setlength{\tabcolsep}{9.90pt}
\settowidth{\accw}{\textbf{00.0}}\settowidth{\tokw}{\tok{000K}}\settowidth{\numw}{\textbf{00.0}}\settowidth{\imgw}{(00.0)}
\begin{tabular}{l*{12}{c}}
\toprule
\midrule
 & \multicolumn{6}{c}{Perception} & \multicolumn{5}{c}{Reasoning} & \\
\cmidrule(lr){2-7}\cmidrule(lr){8-12}
Advantage & OCR & RS & DT & MO & AD & All & OCR & DT & MO & AD & All & Overall \\
\midrule
\multicolumn{13}{c}{Maximum resolution} \\
\midrule[0.3pt]
\shaderow{vira9bg}\textbf{CrossRPO} ($\lambda=0.5$) & \nb{\underline{95.2}} & \nb{\textbf{71.3}} & \nb{\textbf{96.0}} & \nb{\underline{55.8}} & \nb{52.9} & \nb{\underline{68.8}} & \nb{85.0} & \nb{89.0} & \nb{\underline{60.7}} & \nb{46.0} & \nb{\underline{59.9}} & \nb{\textbf{65.3}} \\
\midrule[0.3pt]
\multicolumn{13}{l}{\emph{Joint term}} \\
$\lambda=0$ (per-version RLOO) & \nb{\underline{95.2}} & \nb{69.3} & \nb{\textbf{96.0}} & \nb{53.3} & \nb{50.3} & \nb{67.1} & \nb{84.0} & \nb{\textbf{92.0}} & \nb{\textbf{62.7}} & \nb{44.2} & \nb{59.6} & \nb{64.1} \\
$\lambda=0$, one version per set & \nb{\textbf{96.4}} & \nb{68.7} & \nb{93.0} & \nb{\textbf{57.4}} & \nb{\underline{53.1}} & \nb{\textbf{68.9}} & \nb{\underline{88.0}} & \nb{89.0} & \nb{56.7} & \nb{46.0} & \nb{59.5} & \nb{\underline{65.2}} \\
$\lambda=1$ & \nb{94.8} & \nb{\textbf{71.3}} & \nb{\underline{95.0}} & \nb{51.4} & \nb{50.6} & \nb{66.7} & \nb{86.0} & \nb{88.0} & \nb{57.3} & \nb{\underline{47.2}} & \nb{\underline{59.9}} & \nb{64.0} \\
$\lambda=2$ & \nb{92.8} & \nb{64.0} & \nb{91.0} & \nb{51.4} & \nb{49.1} & \nb{64.6} & \nb{84.0} & \nb{\textbf{92.0}} & \nb{56.0} & \nb{46.5} & \nb{59.5} & \nb{62.6} \\
Random partners & \nb{94.4} & \nb{64.7} & \nb{94.0} & \nb{50.2} & \nb{46.6} & \nb{64.2} & \nb{87.0} & \nb{89.0} & \nb{50.7} & \nb{41.2} & \nb{55.6} & \nb{60.8} \\
\midrule[0.3pt]
\multicolumn{13}{l}{\emph{Per-version GRPO $\times\,(1+\lambda w_v)$}} \\
GRPO, $\lambda=0.5$ & \nb{93.6} & \nb{62.0} & \nb{91.0} & \nb{48.3} & \nb{42.6} & \nb{61.7} & \nb{\textbf{89.0}} & \nb{90.0} & \nb{52.0} & \nb{39.8} & \nb{55.5} & \nb{59.2} \\
GRPO, $\lambda=1$ & \nb{94.4} & \nb{64.0} & \nb{91.0} & \nb{51.7} & \nb{\textbf{54.6}} & \nb{66.6} & \nb{83.0} & \nb{90.0} & \nb{58.0} & \nb{\textbf{49.5}} & \nb{\textbf{61.1}} & \nb{64.5} \\
\midrule[0.3pt]
\multicolumn{13}{l}{\emph{Existing RL methods}} \\
Per-version GRPO & \nb{94.4} & \nb{69.3} & \nb{92.0} & \nb{50.2} & \nb{45.1} & \nb{64.2} & \nb{83.0} & \nb{88.0} & \nb{54.0} & \nb{40.2} & \nb{55.1} & \nb{60.6} \\
Pooled GRPO & \nb{93.6} & \nb{\underline{70.7}} & \nb{91.0} & \nb{50.8} & \nb{48.9} & \nb{65.4} & \nb{83.0} & \nb{\underline{91.0}} & \nb{48.7} & \nb{43.8} & \nb{56.3} & \nb{61.8} \\
Pooled RLOO & \nb{93.6} & \nb{64.0} & \nb{91.0} & \nb{49.2} & \nb{49.7} & \nb{64.3} & \nb{75.0} & \nb{89.0} & \nb{48.0} & \nb{43.5} & \nb{54.7} & \nb{60.6} \\
ConsistRoll~\cite{zou2026consistency} & \nb{94.4} & \nb{66.0} & \nb{93.0} & \nb{51.7} & \nb{48.0} & \nb{65.1} & \nb{85.0} & \nb{88.0} & \nb{54.7} & \nb{45.0} & \nb{58.0} & \nb{62.3} \\
\midrule[1pt]
\multicolumn{13}{c}{1\,MP} \\
\midrule[0.3pt]
\shaderow{vira9bg}\textbf{CrossRPO} ($\lambda=0.5$) & \nb{\underline{94.4}} & \nb{61.3} & \nb{91.0} & \nb{\underline{52.7}} & \nb{52.3} & \nb{\underline{65.9}} & \nb{\underline{86.0}} & \nb{85.0} & \nb{\underline{56.7}} & \nb{\textbf{48.5}} & \nb{\textbf{60.0}} & \nb{\textbf{63.6}} \\
\midrule[0.3pt]
\multicolumn{13}{l}{\emph{Joint term}} \\
$\lambda=0$ (per-version RLOO) & \nb{93.2} & \nb{\textbf{66.0}} & \nb{\textbf{95.0}} & \nb{50.8} & \nb{50.6} & \nb{65.5} & \nb{\textbf{87.0}} & \nb{\textbf{90.0}} & \nb{54.7} & \nb{43.2} & \nb{57.6} & \nb{62.4} \\
$\lambda=0$, one version per set & \nb{94.0} & \nb{58.7} & \nb{\underline{93.0}} & \nb{\textbf{54.2}} & \nb{\underline{54.0}} & \nb{\textbf{66.6}} & \nb{84.0} & \nb{85.0} & \nb{55.3} & \nb{46.2} & \nb{58.3} & \nb{\underline{63.3}} \\
$\lambda=1$ & \nb{93.2} & \nb{60.0} & \nb{91.0} & \nb{47.6} & \nb{49.4} & \nb{63.2} & \nb{81.0} & \nb{86.0} & \nb{55.3} & \nb{41.5} & \nb{55.5} & \nb{60.2} \\
$\lambda=2$ & \nb{92.8} & \nb{58.7} & \nb{90.0} & \nb{50.5} & \nb{48.6} & \nb{63.4} & \nb{81.0} & \nb{83.0} & \nb{48.0} & \nb{45.2} & \nb{55.6} & \nb{60.3} \\
Random partners & \nb{93.6} & \nb{60.0} & \nb{89.0} & \nb{48.9} & \nb{45.4} & \nb{62.3} & \nb{82.0} & \nb{84.0} & \nb{45.3} & \nb{39.8} & \nb{52.4} & \nb{58.4} \\
\midrule[0.3pt]
\multicolumn{13}{l}{\emph{Per-version GRPO $\times\,(1+\lambda w_v)$}} \\
GRPO, $\lambda=0.5$ & \nb{92.0} & \nb{50.7} & \nb{89.0} & \nb{40.8} & \nb{43.1} & \nb{57.8} & \nb{83.0} & \nb{\underline{89.0}} & \nb{50.7} & \nb{38.0} & \nb{53.3} & \nb{56.1} \\
GRPO, $\lambda=1$ & \nb{92.0} & \nb{63.3} & \nb{89.0} & \nb{48.0} & \nb{\textbf{57.1}} & \nb{65.6} & \nb{77.0} & \nb{87.0} & \nb{\textbf{60.0}} & \nb{\underline{47.2}} & \nb{\underline{59.1}} & \nb{63.1} \\
\midrule[0.3pt]
\multicolumn{13}{l}{\emph{Existing RL methods}} \\
Per-version GRPO & \nb{91.2} & \nb{61.3} & \nb{90.0} & \nb{48.9} & \nb{46.6} & \nb{62.4} & \nb{\underline{86.0}} & \nb{84.0} & \nb{45.3} & \nb{43.5} & \nb{54.9} & \nb{59.5} \\
Pooled GRPO & \nb{\textbf{95.2}} & \nb{\underline{64.7}} & \nb{91.0} & \nb{48.6} & \nb{51.4} & \nb{65.1} & \nb{82.0} & \nb{\underline{89.0}} & \nb{47.3} & \nb{41.8} & \nb{54.5} & \nb{61.0} \\
Pooled RLOO & \nb{93.2} & \nb{58.7} & \nb{88.0} & \nb{46.4} & \nb{50.9} & \nb{62.9} & \nb{76.0} & \nb{84.0} & \nb{52.7} & \nb{42.2} & \nb{54.4} & \nb{59.6} \\
ConsistRoll~\cite{zou2026consistency} & \nb{92.4} & \nb{60.0} & \nb{87.0} & \nb{46.4} & \nb{46.0} & \nb{61.3} & \nb{85.0} & \nb{87.0} & \nb{48.7} & \nb{41.2} & \nb{54.7} & \nb{58.7} \\
\midrule
\bottomrule
\end{tabular}
\end{table*}

\begin{table*}[htbp]
\caption{\textbf{Reading the evidence, ablations} at 1\,MP on ViRA-Bench, with the columns of Table~\ref{tab:reading-full}.
\emph{O}, \emph{A} and \emph{B} give the accuracy (\%) on the original and on the two variants, \emph{Zoomed} the share of images with at least one zoom, \emph{On obj.}\ the share whose last zoom covers the object, and \emph{Zooms} the zooms per image.
The rows are those of Table~\ref{tab:ablation-bench}.
All results are means of three runs.}
\label{tab:ablation-reading}
\centering
\scriptsize
\setlength{\tabcolsep}{8.00pt}
\settowidth{\numw}{\textbf{00.0}}
\begin{tabular}{l*{6}{c}}
\toprule
\midrule
 & \multicolumn{3}{c}{Accuracy\,$\uparrow$} & & & \\
\cmidrule(lr){2-4}
Advantage at 1\,MP & O & A & B & Zoomed & On obj.\,$\uparrow$ & Zooms \\
\midrule
\shaderow{vira9bg}\textbf{CrossRPO} ($\lambda=0.5$) & \nb{77.4} & \nb{\underline{68.6}} & \nb{67.2} & \nb{99.9} & \nb{86.6} & \nb{1.07} \\
\midrule[0.3pt]
\multicolumn{7}{l}{\emph{Joint term}} \\
$\lambda=0$ (per-version RLOO) & \nb{75.0} & \nb{63.4} & \nb{61.9} & \nb{100.0} & \nb{\underline{88.3}} & \nb{1.06} \\
$\lambda=0$, one version per set & \nb{\underline{78.2}} & \nb{61.9} & \nb{61.6} & \nb{100.0} & \nb{\textbf{90.0}} & \nb{1.02} \\
$\lambda=1$ & \nb{77.6} & \nb{66.8} & \nb{62.4} & \nb{97.7} & \nb{82.1} & \nb{1.61} \\
$\lambda=2$ & \nb{77.8} & \nb{68.4} & \nb{\underline{68.1}} & \nb{100.0} & \nb{61.0} & \nb{2.72} \\
Random partners & \nb{74.2} & \nb{65.1} & \nb{63.5} & \nb{99.9} & \nb{39.5} & \nb{2.46} \\
\midrule[0.3pt]
\multicolumn{7}{l}{\emph{Per-version GRPO $\times\,(1+\lambda w_v)$}} \\
GRPO, $\lambda=0.5$ & \nb{\textbf{78.4}} & \nb{63.6} & \nb{57.8} & \nb{100.0} & \nb{84.4} & \nb{2.40} \\
GRPO, $\lambda=1$ & \nb{74.1} & \nb{68.3} & \nb{66.1} & \nb{100.0} & \nb{87.8} & \nb{1.01} \\
\midrule[0.3pt]
\multicolumn{7}{l}{\emph{Existing RL methods}} \\
Per-version GRPO & \nb{77.8} & \nb{\textbf{68.7}} & \nb{\textbf{69.4}} & \nb{100.0} & \nb{52.1} & \nb{2.52} \\
Pooled GRPO & \nb{76.5} & \nb{68.2} & \nb{64.9} & \nb{100.0} & \nb{75.5} & \nb{1.53} \\
Pooled RLOO & \nb{75.5} & \nb{65.2} & \nb{64.3} & \nb{100.0} & \nb{71.7} & \nb{1.19} \\
ConsistRoll~\cite{zou2026consistency} & \nb{73.3} & \nb{57.5} & \nb{54.1} & \nb{100.0} & \nb{86.4} & \nb{2.15} \\
\midrule
\bottomrule
\end{tabular}
\end{table*}

\subsection{Settings of the ablations}
\label{sec:supp-ablation-settings}

Every ablation trains Qwen3.5-9B on the training data of ViRA-9B with the same reward, prompts, hyperparameters and number of steps (Table~\ref{tab:hyper}).
Each step samples 16 sets, and each set receives 12 rollouts, $n=12/K$ on each of its $K$ versions.
The runs differ only in how the advantage $A_{v,i}$ of rollout $i$ on version $v$ is computed from the correctness values of the set.
One run, described last, changes the training data instead of the advantage.
We use the notation of Section~\ref{sec:method-crossrpo}: $c_{v,i}$ is the correctness of the rollout, $\hat p_v$ is the share of correct rollouts on version $v$, and $\bar c_{v,-i}$ is the mean correctness of the other $n-1$ rollouts on version $v$.
Table~\ref{tab:ablation-settings} summarizes the variants.
All runs are evaluated at the last step with the protocol of Appendix~\ref{sec:supp-eval}.

\begin{table}[htbp]
\caption{Settings of the ablations.
\emph{Baseline}: the rollouts that the baseline of a rollout averages, those of its own version or those of the whole set.
\emph{Factor}: the multiplier of the difference between the reward and the baseline.
\emph{Scale}: the divisor of the advantages, the root mean square of the batch or the standard deviation of the set or of the version.
$w_v$ is the cross-input weight (Eq.~(\ref{eq:method-weight})), and $\tilde w_v$ is this weight computed from random partners.
\emph{One version per set} changes the training data: each contrast set keeps one version.}
\label{tab:ablation-settings}
\centering
\scriptsize
\setlength{\tabcolsep}{2.40pt}
\begin{tabular}{@{}llll@{}}
\toprule
\midrule
Variant & Baseline & Factor & Scale \\
\midrule
CrossRPO ($\lambda=0.5$) & version & $1+0.5\,w_v$ & batch \\
$\lambda=0$ (per-version RLOO) & version & $1$ & batch \\
$\lambda=0$, one version per set & version & $1$ & batch \\
$\lambda=1$ & version & $1+w_v$ & batch \\
$\lambda=2$ & version & $1+2\,w_v$ & batch \\
Random partners & version & $1+0.5\,\tilde w_v$ & batch \\
Per-version GRPO & version & $1$ & version \\
Per-version GRPO, $\lambda=0.5$ & version & $1+0.5\,w_v$ & version \\
Per-version GRPO, $\lambda=1$ & version & $1+w_v$ & version \\
Pooled RLOO & set & $1$ & batch \\
Pooled GRPO & set & $1$ & set \\
ConsistRoll~\cite{zou2026consistency} & set & $1$, with a pair bonus in the reward & set \\
\midrule
\bottomrule
\end{tabular}
\end{table}

\paragraph{CrossRPO ($\lambda=0.5$).}
This run is ViRA-9B.
Its advantage is Eq.~(\ref{eq:method-advantage}), $A_{v,i}=(1+\lambda w_v)(c_{v,i}-\bar c_{v,-i})$ with $w_v=\prod_{u\neq v}\hat p_u$ and $\lambda=0.5$.
The baseline $\bar c_{v,-i}$ uses only the rollouts of the same version, and the weight $w_v$ uses only the rollouts of the other versions.
For a single image ($K=1$), $w_1=0$.
All advantages of a batch are divided by their root mean square (Eq.~(\ref{eq:method-scale})).

\paragraph{$\lambda=0$, $\lambda=1$ and $\lambda=2$.}
These runs change only the value of $\lambda$ in Eq.~(\ref{eq:method-advantage}).
With $\lambda=0$, the joint term is removed, and $A_{v,i}=c_{v,i}-\bar c_{v,-i}$.
Every version is then trained as a separate single image with RLOO, and no advantage depends on another version.
With $\lambda=1$ and $\lambda=2$, the joint term has two and four times its weight in ViRA-9B.
The factor $1+\lambda w_v$ then lies between 1 and 2 for $\lambda=1$ and between 1 and 3 for $\lambda=2$, against 1 and 1.5 for $\lambda=0.5$.

\paragraph{Random partners.}
This run keeps Eq.~(\ref{eq:method-advantage}) with $\lambda=0.5$ but computes the weight from other photos.
In each step, the versions of all contrast sets in the batch are assigned at random to new groups, which have the same sizes as the original sets.
The assignment avoids two versions of the same set in one group whenever this is possible.
The weight of version $v$ is $\tilde w_v=\prod_u\hat p_u$ over the other members $u$ of its new group.
The images, the rollouts and the baseline $\bar c_{v,-i}$ are unchanged.
Each version therefore has as many partners as before, but the partners show other photos with other questions.
Single images are not affected.
Proposition~\ref{prop:independent} analyzes this case.

\paragraph{Per-version GRPO.}
This run applies GRPO~\cite{shao2024deepseekmath} to each version separately, as implemented in verl~\cite{sheng2025hybridflow}.
The $n$ rollouts of a version form one group, and $A_{v,i}=(c_{v,i}-\mu_v)/\sigma_v$, where $\mu_v$ and $\sigma_v$ are the mean and the standard deviation of their correctness.
If all rollouts of a version have the same correctness, their advantages are zero.
The run has no joint term and no batch scale.
It differs from $\lambda=0$ in the division by the standard deviation of each version instead of the root mean square of the batch (Proposition~\ref{prop:scale}).
For a single image, it is the same as pooled GRPO.

\paragraph{Per-version GRPO with the cross-input weight.}
Two runs multiply the advantage of per-version GRPO by the factor $1+\lambda w_v$ of CrossRPO, $A_{v,i}=(1+\lambda w_v)(c_{v,i}-\mu_v)/\sigma_v$ with the cross-input weight $w_v=\prod_{u\neq v}\hat p_u$ of Eq.~(\ref{eq:method-weight}), for $\lambda=0.5$ and $\lambda=1$.
The baseline $\mu_v$ and the scale $\sigma_v$ are those of per-version GRPO.
All rollouts of a version share the factor, so a division that came after it would cancel it (Section~\ref{sec:method-crossrpo}).
The factor is therefore applied after the division by $\sigma_v$.
The runs have no batch scale.
For a single image, $w_1=0$, and the advantage is that of per-version GRPO.
These runs are not in Table~\ref{tab:ablation}, and Appendix~\ref{sec:supp-ablation-joint} gives their results.

\paragraph{Pooled RLOO.}
This run has no joint term, and its baseline is computed over the whole set instead of over one version.
All $N=Kn=12$ rollouts of a set form one group, and $A_{v,i}=c_{v,i}-\bar c_{-(v,i)}$, where $\bar c_{-(v,i)}$ is the mean correctness of the other $N-1$ rollouts of the set on all versions.
The advantages are divided by the root mean square of the batch, as in CrossRPO.
The run therefore differs from $\lambda=0$ only in the baseline (Proposition~\ref{prop:baseline}), and for a single image the two advantages are equal.

\paragraph{Pooled GRPO.}
This run applies GRPO~\cite{shao2024deepseekmath} to each set, as implemented in verl~\cite{sheng2025hybridflow}.
All 12 rollouts of a set form one group, and $A_{v,i}=(c_{v,i}-\mu)/\sigma$, where $\mu$ and $\sigma$ are the mean and the standard deviation of the correctness of these rollouts.
If all 12 rollouts have the same correctness, their advantages are zero.
The run has no joint term and no batch scale.
It differs from pooled RLOO in the division by the standard deviation of each set (Proposition~\ref{prop:scale}).
A single image is a group of its 12 rollouts with the same formula.

\paragraph{ConsistRoll.}
This run reproduces ConsistRoll~\cite{zou2026consistency} on our data.
It is pooled GRPO with a reward that adds a bonus for pairs of correct rollouts.
Rollouts on two versions are paired by their index $i$, and a pair in which both rollouts are correct adds 0.5 to the reward of each.
For $K=2$, the two versions are paired, so $r_{v,i}=c_{v,i}+0.5\,c_{1,i}c_{2,i}$.
For $K=3$, the original O is paired with each variant.
A rollout on variant A has $r_{\mathrm A,i}=c_{\mathrm A,i}+0.5\,c_{\mathrm O,i}c_{\mathrm A,i}$, and likewise for B.
A rollout on the original receives the mean of its two bonuses, $r_{\mathrm O,i}=c_{\mathrm O,i}+0.25\,(c_{\mathrm O,i}c_{\mathrm A,i}+c_{\mathrm O,i}c_{\mathrm B,i})$.
Single images receive no bonus.
The advantage is that of pooled GRPO with $r_{v,i}$ in place of $c_{v,i}$.
ConsistRoll was proposed for pairs of an image and a perturbed copy with the same answer, so the pairing for $K=3$ is our extension.
Theorem~\ref{thm:rb} compares pairing by index with the average over all combinations in $w_v$.

\paragraph{One version per set.}
This run keeps the advantage of $\lambda=0$ and changes the training data.
Each of the 1,664 contrast sets is reduced to one version.
The 987 sets with three versions keep the original, and the 677 sets with two versions keep variant A, since their questions have answers for A and B only (Appendix~\ref{sec:supp-train}).
The single images are unchanged, so the run has as many training items and steps as ViRA-9B.
All 12 rollouts of a set are on its one version, so $K=1$, $w_1=0$ and $A_{1,i}=c_{1,i}-\bar c_{1,-i}$ for every item, with the batch scale.
The run thus differs from $\lambda=0$ only in the data: $\lambda=0$ trains on all versions of a set as separate images, and this run trains on one of them.
It is not in Table~\ref{tab:ablation}, and Appendix~\ref{sec:supp-ablation-onever} gives its results.

\subsection{Zoom-box size on ViRA-Bench-Real}
\label{sec:supp-box}

On ViRA-Bench-Real, the pooled RLOO ablation is ahead of ViRA-9B at every budget below the maximum resolution (Table~\ref{tab:ablation}), although it is behind on ViRA-Bench.
Its zoom behavior explains the difference.
Both agents put the object inside their last zoom on about 95\% of the images.
However, ViRA-9B almost always zooms once, with a box of about 1.1 to 1.5\,MP of the photo regardless of the size of the object.
In contrast, the ablation draws boxes of 0.3 to 0.7\,MP and zooms again more often.
At 0.5\,MP, the object's longer side therefore spans a median of 24\,px in ViRA-9B's last view and 32\,px in the ablation's.
When the object has the same size in the last view, the two agents are equally accurate.
On ViRA-Bench, the objects are larger (a longer side of 10\,px at 1\,MP instead of 6\,px), so their longer side still spans a median of 36\,px in ViRA-9B's last view at 0.5\,MP, and the gap between the~two~agents~disappears.

To test this, we enlarged every zoom box of the ablation around its center, which keeps the object inside the box.
We evaluated the result on 128 photographed sets, the 78 sets of ViRA-Bench-Real and 50 further sets shot in the same way, with three runs each.
Doubling the box area lowers its set accuracy at 0.5, 1 and 2\,MP from 40.4 / 52.6 / 66.7 to 33.1 / 44.5 / 58.1.
This is close to ViRA-9B (30.5 / 42.4 / 53.9).
Quadrupling the box area gives 18.5 / 39.8 / 56.8.
Shrinking ViRA-9B's boxes around their centers does not help, because the smaller box often cuts the object off.

The per-version GRPO ablation shows the same effect.
It is ahead of ViRA-9B up to 4\,MP on ViRA-Bench-Real and at 0.5 and 1\,MP on ViRA-Bench (Table~\ref{tab:ablation}).
This agent zooms 2.0 to 2.7 times per image, and its last box covers a median of 0.05 to 0.23\,MP of the photo.
At 0.5\,MP on ViRA-Bench-Real, the object's longer side spans a median of 49\,px in its largest view that contains the object, twice the 24\,px of ViRA-9B.
The additional zooms cost tokens: on ViRA-Bench it uses 27K to 233K tokens per set, against 12K to 147K for ViRA-9B (Table~\ref{tab:ablation-bench}).
Its small boxes also miss the object more often.
On ViRA-Bench, at least one of its zooms covers the object on 84.5\% to 92.1\% of the images, against 88.4\% to 97.0\% for ViRA-9B, and it is behind ViRA-9B from 2\,MP on.
At half the area, coverage falls from 95\% to 78\%.
However, on the images where the smaller box still covers the object, accuracy rises from 61\% to 72\% at 0.5\,MP.
The low-budget gap on ViRA-Bench-Real is thus a matter of box size rather than of reading.

\subsection{Per-version GRPO with the cross-input weight}
\label{sec:supp-ablation-joint}

Per-version GRPO is ahead of CrossRPO at low budgets (Table~\ref{tab:ablation}).
We therefore test whether the joint term also helps per-version GRPO.
CrossRPO adds the joint term to RLOO by multiplying the advantage of each version by $1+\lambda w_v$, where $w_v=\prod_{u\neq v}\hat p_u$ is the cross-input weight (Eq.~(\ref{eq:method-weight})).
Two further runs multiply the advantage of per-version GRPO by the same factor:
\begin{align}
  \text{per-version GRPO:}\quad&A_{v,i}=\frac{c_{v,i}-\mu_v}{\sigma_v},\notag\\
  \text{CrossRPO:}\quad&A_{v,i}=(1+\lambda w_v)\,(c_{v,i}-\bar c_{v,-i}),\notag\\
  \text{the two runs:}\quad&A_{v,i}=(1+\lambda w_v)\,\frac{c_{v,i}-\mu_v}{\sigma_v}.\label{eq:supp-gj}
\end{align}
The two runs keep the baseline $\mu_v$ and the scale $\sigma_v$ of per-version GRPO and take only the factor from CrossRPO, with $\lambda=0.5$ and $\lambda=1$ (Appendix~\ref{sec:supp-ablation-settings}).
The tables of this appendix list them under \emph{Per-version GRPO $\times\,(1+\lambda w_v)$}.

On the contrast sets, the weight does not improve per-version GRPO.
On ViRA-Bench, per-version GRPO is ahead of both runs at every budget, by 4.1 to 12.8 points for $\lambda=0.5$ and by 0.2 to 6.3 points for $\lambda=1$ (Table~\ref{tab:ablation-bench}).
CrossRPO is also ahead of both runs at every budget, by 5.3 to 8.5 points and by 0.6 to 4.4 points.
On ViRA-Bench-Real, per-version GRPO is again ahead of both runs at every budget.
Table~\ref{tab:ablation-joint} gives the general benchmarks.
CrossRPO is ahead of the run with $\lambda=1$ in all columns but one and of the run with $\lambda=0.5$ in all columns but three.
On MME-RealWorld-Lite, CrossRPO is ahead of both runs on the perception questions and overall at both resolutions (Table~\ref{tab:ablation-mmerw}).

\begin{table*}[htbp]
\caption{\textbf{Per-version GRPO with the cross-input weight, general benchmarks}: accuracy (\%) at 1\,MP and at the maximum resolution (\emph{Max}), in the columns of Table~\ref{tab:ablation}.
The rows \emph{Per-version GRPO, $\lambda$} multiply the advantage of per-version GRPO by $1+\lambda w_v$ (Appendix~\ref{sec:supp-ablation-joint}).}
\label{tab:ablation-joint}
\centering
\scriptsize
\setlength{\tabcolsep}{7.10pt}
\settowidth{\numw}{\textbf{00.0}}
\begin{tabular}{l*{14}{c}}
\toprule
\midrule
 & \multicolumn{2}{c}{V$^*$} & \multicolumn{2}{c}{HR-4K} & \multicolumn{2}{c}{HR-8K} & \multicolumn{2}{c}{MME-RW} & \multicolumn{2}{c}{TreeBench} & \multicolumn{2}{c}{O3-Bench} & MMStar & Hallusion \\
\cmidrule(lr){2-3}\cmidrule(lr){4-5}\cmidrule(lr){6-7}\cmidrule(lr){8-9}\cmidrule(lr){10-11}\cmidrule(lr){12-13}\cmidrule(lr){14-14}\cmidrule(l){15-15}
Advantage & 1 & Max & 1 & Max & 1 & Max & 1 & Max & 1 & Max & 1 & Max & Max & Max \\
\midrule
\shaderow{vira9bg}\textbf{CrossRPO} ($\lambda=0.5$) & \nb{\textbf{92.0}} & \nb{\textbf{96.2}} & \nb{\textbf{87.5}} & \nb{\underline{91.0}} & \nb{\textbf{84.7}} & \nb{\textbf{90.1}} & \nb{\textbf{63.6}} & \nb{\textbf{65.3}} & \nb{\textbf{56.3}} & \nb{\textbf{57.0}} & \nb{\textbf{50.4}} & \nb{\underline{67.0}} & \nb{\underline{77.1}} & \nb{\underline{79.6}} \\
\midrule[0.3pt]
Per-version GRPO, $\lambda=0.5$ & \nb{88.7} & \nb{\underline{95.8}} & \nb{83.9} & \nb{87.8} & \nb{80.8} & \nb{88.6} & \nb{56.1} & \nb{59.2} & \nb{\underline{54.6}} & \nb{53.1} & \nb{43.5} & \nb{\textbf{67.8}} & \nb{\textbf{78.3}} & \nb{\textbf{82.9}} \\
Per-version GRPO, $\lambda=1$ & \nb{87.3} & \nb{94.2} & \nb{85.7} & \nb{\textbf{91.7}} & \nb{83.2} & \nb{\underline{89.8}} & \nb{\underline{63.1}} & \nb{\underline{64.5}} & \nb{51.4} & \nb{50.1} & \nb{35.9} & \nb{53.9} & \nb{75.0} & \nb{77.1} \\
Per-version GRPO & \nb{\underline{89.5}} & \nb{95.3} & \nb{\underline{86.0}} & \nb{90.8} & \nb{\underline{84.6}} & \nb{88.8} & \nb{59.5} & \nb{60.6} & \nb{53.6} & \nb{\underline{54.8}} & \nb{\underline{43.8}} & \nb{58.6} & \nb{75.1} & \nb{79.2} \\
\midrule
\bottomrule
\end{tabular}
\end{table*}

The two runs zoom in different ways (Table~\ref{tab:ablation-reading}).
With $\lambda=1$, the agent zooms once per image, as ViRA-9B does, and uses 10K to 137K tokens per set on ViRA-Bench, against 12K to 147K for ViRA-9B.
With $\lambda=0.5$, the agent zooms 2.4 times per image at 1\,MP, about as often as per-version GRPO (2.5 times), but it draws large boxes.
On ViRA-Bench, its last box covers a median of 1.5 to 1.9\,MP of the photo, against 0.12 to 0.23\,MP for per-version GRPO.
At 0.5\,MP, the object's longer side spans a median of 26\,px in its largest view that contains the object, against 64\,px for per-version GRPO and 34\,px for ViRA-9B, which zooms once.
The run uses 26K to 313K tokens per set.

\subsection{Training on one version per set}
\label{sec:supp-ablation-onever}

Without the joint term ($\lambda=0$), the versions of a contrast set are still in the training data, as separate images.
A further run also removes the other versions: each contrast set keeps one version, and the advantage is that of $\lambda=0$ (Appendix~\ref{sec:supp-ablation-settings}).
The three runs form a sequence.
The run with one version per set trains on one image of each set, $\lambda=0$ adds the other versions as separate images, and CrossRPO adds the joint term.
The tables of this appendix list the run as \emph{$\lambda=0$, one version per set}.

On ViRA-Bench, the other versions alone do not raise set accuracy.
The run with one version per set and $\lambda=0$ differ by at most 0.7 points at every budget (31.1 / 40.5 / 49.3 / 53.6 / 58.1 against 31.3 / 41.2 / 50.0 / 52.9 / 58.7, Table~\ref{tab:ablation-bench}).
CrossRPO is ahead of both at every budget, by 3.9 to 7.6 points over the run with one version per set.
On ViRA-Bench, the gain of CrossRPO therefore comes from the joint term and not from the additional images.
On ViRA-Bench-Real, the run is behind $\lambda=0$ by 1.3 to 3.0 points at four budgets and ahead by 1.3 points at 2\,MP.
CrossRPO is ahead of it by 4.7 to 10.7 points at four budgets and behind by 0.4 points at 2\,MP.

The difference from CrossRPO is on the variants (Table~\ref{tab:ablation-reading}).
At 1\,MP, the accuracy on the originals is 78.2 for the run with one version per set, 75.0 for $\lambda=0$ and 77.4 for CrossRPO.
On the variants A and B, it is 61.9 and 61.6 for the run with one version per set and 63.4 and 61.9 for $\lambda=0$, against 68.6 and 67.2 for CrossRPO.
All three agents zoom once per image and use about the same number of tokens per set on ViRA-Bench (10K to 147K, 11K to 144K and 12K to 147K).
The two runs without the joint term draw larger boxes than CrossRPO.
Their last box covers a median of 1.4 to 2.2\,MP of the photo with one version per set and 1.1 to 1.9\,MP with $\lambda=0$, against 0.8 to 1.4\,MP for CrossRPO.
At 0.5\,MP, the object's longer side spans a median of 27\,px and 28\,px in their largest view that contains the object, against 34\,px for CrossRPO.

Table~\ref{tab:ablation-onever} gives the general benchmarks.
CrossRPO is ahead of the run with one version per set in ten of the 14 columns and tied in one.
The largest difference is on O3-Bench, where the run is behind CrossRPO by 6.6 points at 1\,MP and by 3.8 points at the maximum resolution, and within 0.9 points of $\lambda=0$.
On V$^*$ and HR-Bench, the run is behind $\lambda=0$ by 1.8 to 3.2 points at 1\,MP and within 0.7 points of it at the maximum resolution.
On MME-RealWorld-Lite, the run is behind CrossRPO on the reasoning questions and overall (Table~\ref{tab:ablation-mmerw}).

\begin{table*}[htbp]
\caption{\textbf{Training on one version per set, general benchmarks}: accuracy (\%) at 1\,MP and at the maximum resolution (\emph{Max}), in the columns of Table~\ref{tab:ablation}.
The last row trains on one version of each contrast set with the advantage of $\lambda=0$ (Appendix~\ref{sec:supp-ablation-onever}).}
\label{tab:ablation-onever}
\centering
\scriptsize
\setlength{\tabcolsep}{7.20pt}
\settowidth{\numw}{\textbf{00.0}}
\begin{tabular}{l*{14}{c}}
\toprule
\midrule
 & \multicolumn{2}{c}{V$^*$} & \multicolumn{2}{c}{HR-4K} & \multicolumn{2}{c}{HR-8K} & \multicolumn{2}{c}{MME-RW} & \multicolumn{2}{c}{TreeBench} & \multicolumn{2}{c}{O3-Bench} & MMStar & Hallusion \\
\cmidrule(lr){2-3}\cmidrule(lr){4-5}\cmidrule(lr){6-7}\cmidrule(lr){8-9}\cmidrule(lr){10-11}\cmidrule(lr){12-13}\cmidrule(lr){14-14}\cmidrule(l){15-15}
Run & 1 & Max & 1 & Max & 1 & Max & 1 & Max & 1 & Max & 1 & Max & Max & Max \\
\midrule
\shaderow{vira9bg}\textbf{CrossRPO} ($\lambda=0.5$) & \nb{\underline{92.0}} & \nb{\textbf{96.2}} & \nb{\underline{87.5}} & \nb{91.0} & \nb{\underline{84.7}} & \nb{\underline{90.1}} & \nb{\textbf{63.6}} & \nb{\textbf{65.3}} & \nb{\textbf{56.3}} & \nb{\textbf{57.0}} & \nb{\textbf{50.4}} & \nb{\textbf{67.0}} & \nb{77.1} & \nb{\underline{79.6}} \\
\midrule[0.3pt]
$\lambda=0$ (per-version RLOO) & \nb{\textbf{93.0}} & \nb{\underline{95.6}} & \nb{\textbf{87.8}} & \nb{\textbf{92.9}} & \nb{\textbf{85.7}} & \nb{89.8} & \nb{62.4} & \nb{64.1} & \nb{\underline{54.3}} & \nb{55.3} & \nb{43.5} & \nb{\underline{64.1}} & \nb{\textbf{78.8}} & \nb{\textbf{80.2}} \\
$\lambda=0$, one version per set & \nb{90.9} & \nb{95.5} & \nb{86.0} & \nb{\underline{92.2}} & \nb{82.5} & \nb{\textbf{90.2}} & \nb{\underline{63.3}} & \nb{\underline{65.2}} & \nb{\textbf{56.3}} & \nb{\underline{56.5}} & \nb{\underline{43.8}} & \nb{63.2} & \nb{\underline{78.7}} & \nb{78.7} \\
\midrule
\bottomrule
\end{tabular}
\end{table*}

\section{Finding and reading the evidence}
\label{sec:supp-reading}

Table~\ref{tab:reading-full} gives, for every model with a zoom tool, the accuracy on each version and how the model zooms at 1\,MP on ViRA-Bench.
ViRA-9B is more accurate than Qwen3.5-9B with the zoom tool on all three versions, by 14.0 points on O and by 16.1 and 17.3 points on A and B.
Its last zoom covers the object on 86.6\% of the images instead of 62.6\%, with 1.07 instead of 1.82 zooms per image.
LBS-27B and the proprietary models zoom two to three times per image and cover the object less often than ViRA-9B.
The joint term improves reading rather than finding (Table~\ref{tab:ablation-reading}): without it ($\lambda=0$), the last zoom covers the object at least as often, but accuracy drops about twice as much on the variants A and B as on the original O.
Several models often answer without zooming.
Qwen2.5-VL-7B with the DeepEyes tool zooms on 49.0\% of the images, Pixel Reasoner on 44.6\%, Thyme on 18.1\% and Vision-OPD-9B with our tool on 4.3\%.
DeepEyesV2 zooms on 75.2\% of the images, but its last crop covers the object on only 5.7\%.
ZoomEye visits 21.8 crops per image in its tree search.
However, its final answer uses a crop on 35.0\% of the images and the whole image on the others.

\begin{table*}[htbp]
\caption{\textbf{Reading the evidence, all models} at 1\,MP on ViRA-Bench, for every model with a zoom tool.
\emph{O}, \emph{A} and \emph{B} give the accuracy (\%) on the original and on the two variants, \emph{Zoomed} the share of images with at least one zoom, \emph{On obj.}\ the share whose last zoom covers the object, and \emph{Zooms} the zooms per image.
ZoomEye is a tree search, so its \emph{Zoomed} and \emph{On obj.} refer to the crop that its final answer uses, and its \emph{Zooms} counts the crops that the search visits.
ViRA results are means of three runs.
Table~\ref{tab:ablation-reading} gives the ablations.}
\label{tab:reading-full}
\centering
\scriptsize
\setlength{\tabcolsep}{8.00pt}
\settowidth{\numw}{\textbf{00.0}}
\begin{tabular}{l*{6}{c}}
\toprule
\midrule
 & \multicolumn{3}{c}{Accuracy\,$\uparrow$} & & & \\
\cmidrule(lr){2-4}
Model at 1\,MP & O & A & B & Zoomed & On obj.\,$\uparrow$ & Zooms \\
\midrule
\multicolumn{7}{l}{\emph{Open 7--9B models + tool}} \\
Qwen3.5-9B + tool & \nb{63.4} & \nb{52.5} & \nb{49.9} & \nb{99.1} & \nb{62.6} & \nb{1.82} \\
ZwZ2-9B + tool & \nb{68.9} & \nb{60.5} & \nb{55.8} & \nb{97.1} & \nb{64.8} & \nb{1.56} \\
Vision-OPD-9B + tool & \nb{42.7} & \nb{35.3} & \nb{29.3} & \nb{4.3} & \nb{3.4} & \nb{0.08} \\
Qwen2.5-VL-7B + DeepEyes tool & \nb{47.0} & \nb{24.9} & \nb{19.0} & \nb{49.0} & \nb{18.8} & \nb{2.45} \\
\midrule[0.3pt]
\multicolumn{7}{l}{\emph{Released zoom agents (7--9B)}} \\
Mini-o3 & \nb{58.3} & \nb{43.0} & \nb{42.9} & \nb{99.7} & \nb{69.0} & \nb{3.48} \\
DeepEyes & \nb{50.4} & \nb{32.2} & \nb{25.1} & \nb{100.0} & \nb{47.0} & \nb{1.01} \\
DeepEyesV2 & \nb{43.3} & \nb{30.2} & \nb{23.5} & \nb{75.2} & \nb{5.7} & \nb{1.69} \\
Thyme & \nb{48.8} & \nb{29.4} & \nb{25.3} & \nb{18.1} & \nb{9.8} & \nb{0.18} \\
Pixel Reasoner & \nb{52.8} & \nb{30.2} & \nb{22.2} & \nb{44.6} & \nb{31.4} & \nb{0.45} \\
ZoomEye & \nb{58.7} & \nb{41.9} & \nb{36.7} & \nb{35.0} & \nb{27.1} & \nb{21.84} \\
\midrule[0.3pt]
\multicolumn{7}{l}{\emph{Ours + our zoom tool}} \\
\shaderow{vira9bg}\textbf{ViRA-9B} (ours) & \nb{\textbf{77.4}} & \nb{\textbf{68.6}} & \nb{\textbf{67.2}} & \nb{99.9} & \nb{\textbf{86.6}} & \nb{1.07} \\
\midrule[1pt]
\multicolumn{7}{l}{\emph{Open 27B models + tool}} \\
Qwen3.8-27B + tool & \nb{56.2} & \nb{55.6} & \nb{53.3} & \nb{100.0} & \nb{59.4} & \nb{3.06} \\
ZwZ2-27B + tool & \nb{58.8} & \nb{53.3} & \nb{52.2} & \nb{99.9} & \nb{55.5} & \nb{3.69} \\
\midrule[0.3pt]
\multicolumn{7}{l}{\emph{Released zoom agents (27--32B)}} \\
LBS-27B & \nb{82.0} & \nb{73.5} & \nb{72.7} & \nb{100.0} & \nb{73.8} & \nb{1.96} \\
SenseNova-MARS-32B & \nb{75.2} & \nb{60.0} & \nb{52.4} & \nb{97.4} & \nb{62.6} & \nb{3.01} \\
\midrule[0.3pt]
\multicolumn{7}{l}{\emph{Ours + our zoom tool}} \\
\shaderow{vira27bg}\textbf{ViRA-27B} (ours) & \nb{\textbf{88.5}} & \nb{\textbf{83.0}} & \nb{\textbf{82.5}} & \nb{100.0} & \nb{\textbf{91.0}} & \nb{2.38} \\
\midrule[1pt]
\multicolumn{7}{l}{\emph{Proprietary models + our zoom tool}} \\
Claude Haiku~5.5, max thinking & \nb{\textbf{85.2}} & \nb{\textbf{74.9}} & \nb{\textbf{77.2}} & \nb{100.0} & \nb{\textbf{75.9}} & \nb{2.74} \\
Claude Haiku~5.5, medium thinking & \nb{83.5} & \nb{65.4} & \nb{66.9} & \nb{97.4} & \nb{65.0} & \nb{1.48} \\
Claude Haiku~5.5, no thinking & \nb{77.6} & \nb{56.8} & \nb{58.0} & \nb{99.6} & \nb{65.0} & \nb{1.28} \\
GPT-6-Luna, max thinking & \nb{77.3} & \nb{71.1} & \nb{70.5} & \nb{98.9} & \nb{73.8} & \nb{2.48} \\
GPT-6-Luna, medium thinking & \nb{81.2} & \nb{68.5} & \nb{64.6} & \nb{98.9} & \nb{66.1} & \nb{2.06} \\
GPT-6-Luna, no thinking & \nb{63.6} & \nb{44.1} & \nb{43.6} & \nb{100.0} & \nb{45.8} & \nb{1.61} \\
\midrule
\bottomrule
\end{tabular}
\end{table*}

\begin{figure*}[tp]
  \centering
  \includegraphics[width=\textwidth]{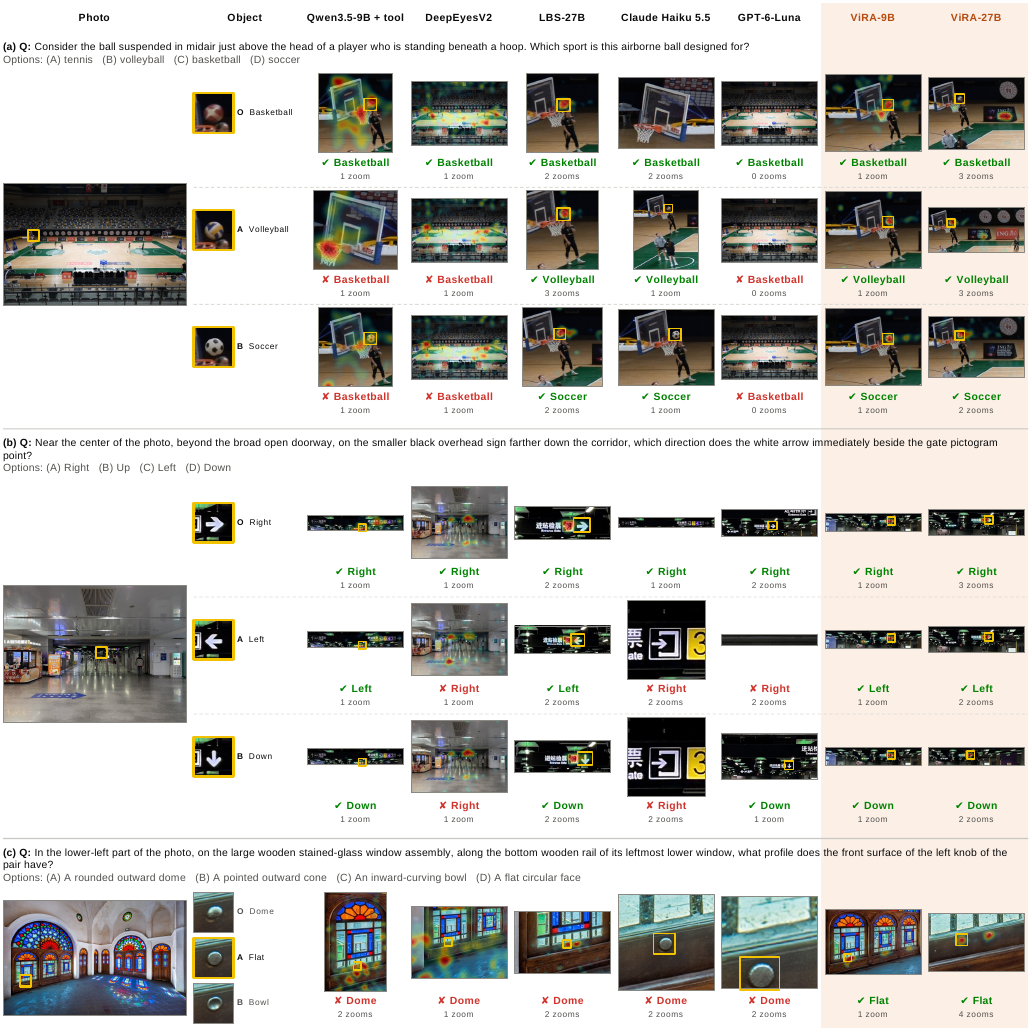}
  \caption{\textbf{Case study on ViRA-Bench at 1\,MP: seven models on 14 sets (Figures~\ref{fig:cases} to~\ref{fig:cases4}).}
  A row shows one variant or all three versions of a set.
  It gives the question, the whole photo with the object in a yellow box, the object at full resolution in each version, and for every model the last crop that it received with its answer.
  A yellow box in a crop marks the object, and a model that received no crop of the photo is shown with the whole photo.
  For the open models, colors show where the final reply looks in the shown image (Appendix~\ref{sec:supp-cases}).
  Claude Haiku~5.5 and GPT-6-Luna run at max thinking.}
  \label{fig:cases}
\end{figure*}

\begin{figure*}[tp]
  \centering
  \includegraphics[width=\textwidth]{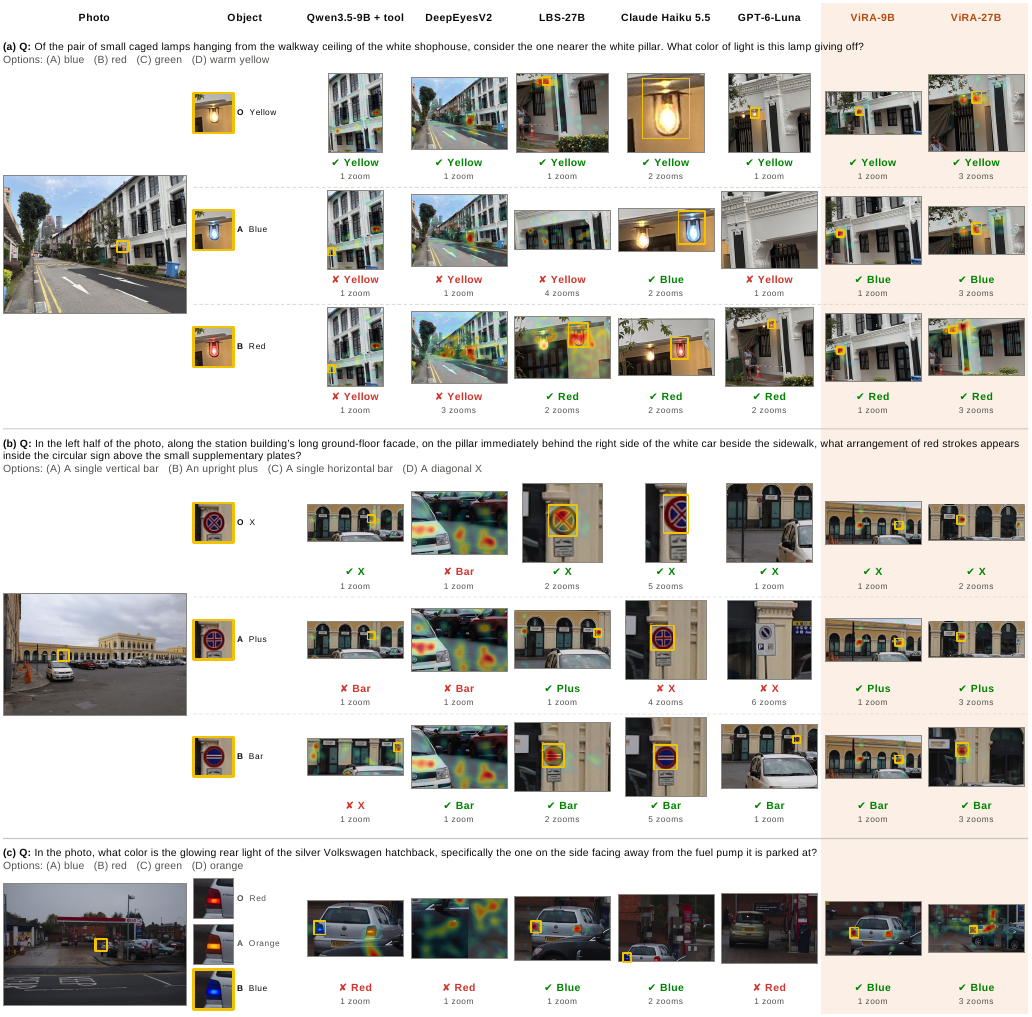}
  \caption{\textbf{Case study (continued).} The layout is as in Figure~\ref{fig:cases}.
  Baselines give the original's answer on a variant, and both ViRA agents answer every shown version correctly.}
  \label{fig:cases2}
\end{figure*}

\begin{figure*}[tp]
  \centering
  \includegraphics[width=\textwidth]{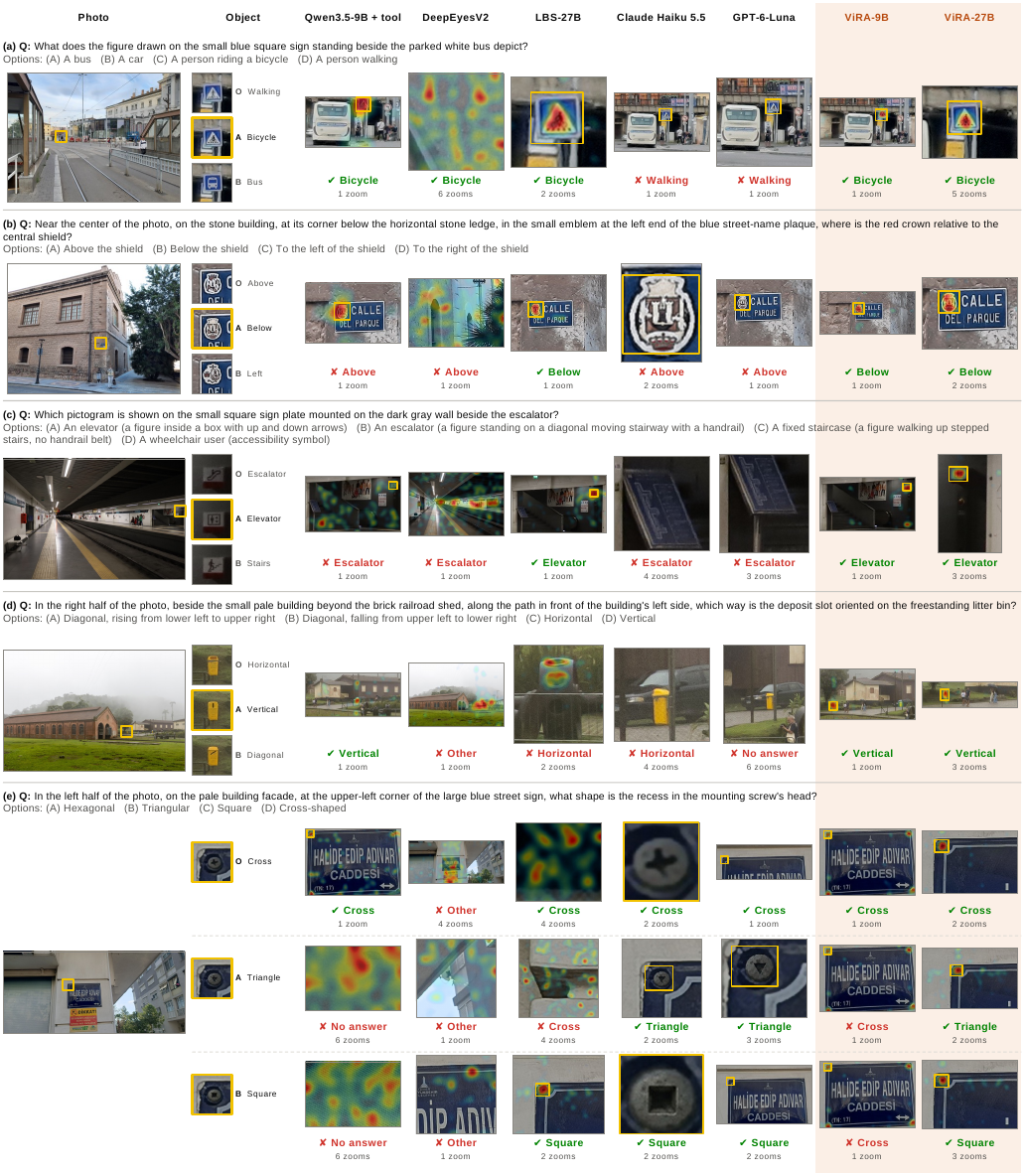}
  \caption{\textbf{Case study (continued).} The layout is as in Figure~\ref{fig:cases}.
  (a, b)~The proprietary models have the object in their last crop but give the original's answer.
  (c, d)~They zoom up to six times without covering the object.
  (e)~ViRA-9B gives this answer on every version.}
  \label{fig:cases3}
\end{figure*}

\begin{figure*}[tp]
  \centering
  \includegraphics[width=\textwidth]{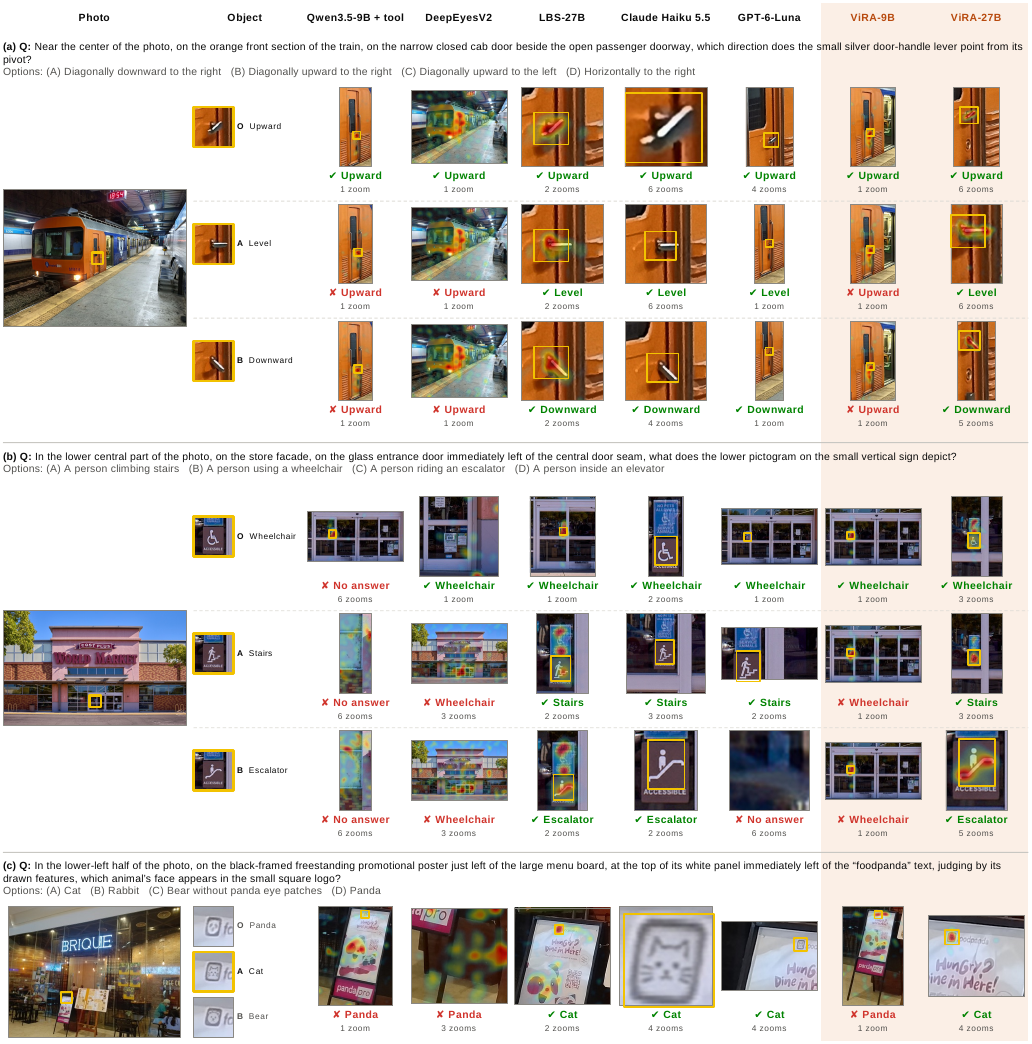}
  \caption{\textbf{Case study (continued).} The layout is as in Figure~\ref{fig:cases}.
  ViRA-9B zooms once and gives the original's answer on the variants, whereas ViRA-27B zooms several times and answers correctly.}
  \label{fig:cases4}
\end{figure*}

\subsection{Error types}
\label{sec:supp-errors}

Table~\ref{tab:errors} sorts all wrong answers of ViRA-9B on ViRA-Bench by type, and Figures~\ref{fig:cases3} and~\ref{fig:cases4} show examples.
A crop covers the object if its box on the photo contains at least half of the object's box.
Most errors happen after a crop has covered the object, 78\% at 1\,MP and 89\% at the maximum resolution.
On a variant, such an error gives the original's answer in 63\% of the cases, whereas a uniform choice among the three wrong options would give it in 33\%.
In 21\% of the errors at 1\,MP, no crop covers the object.
At this budget, ViRA-9B zooms exactly once in 94\% of the rollouts and never draws a box on an earlier crop.
Therefore, a box that misses the object or is too large is rarely corrected (Appendix~\ref{sec:supp-box}).
Qwen3.5-9B with the zoom tool makes more errors, and 16.6\% of them at 1\,MP give no answer, mostly because it reaches the limit of six calls.

\begin{table}[htbp]
\caption{\textbf{Error types} on ViRA-Bench (2,286 images of 762 sets).
A crop covers the object if its box on the photo contains at least half of the object's box.
\emph{No answer} includes rollouts that reach the limit of six calls.
ViRA-9B numbers are means of three runs, and Qwen3.5-9B + tool is run once.}
\label{tab:errors}
\centering
\scriptsize
\setlength{\tabcolsep}{6.60pt}
\settowidth{\numw}{\textbf{00.0}}
\begin{tabular}{lcccc}
\toprule
\midrule
 & \multicolumn{2}{c}{ViRA-9B} & \multicolumn{2}{c}{Qwen3.5-9B + tool} \\
\cmidrule(lr){2-3}\cmidrule(l){4-5}
 & 1\,MP & Max & 1\,MP & Max \\
\midrule
Wrong answers (\% of images) & \nb{28.9} & \nb{18.7} & \nb{44.8} & \nb{37.1} \\
\midrule
\multicolumn{5}{l}{\emph{Share of the wrong answers (\%)}} \\
No answer & \nb{0.7} & \nb{0.5} & \nb{16.6} & \nb{6.4} \\
Answered without zooming & \nb{0.4} & \nb{0.1} & \nb{1.5} & \nb{3.4} \\
No crop covers the object & \nb{21.0} & \nb{10.0} & \nb{22.8} & \nb{19.1} \\
A crop covers the object: & & & & \\
\quad the original's answer on a variant & \nb{36.1} & \nb{41.0} & \nb{25.8} & \nb{30.2} \\
\quad another wrong answer & \nb{41.9} & \nb{48.3} & \nb{33.3} & \nb{40.9} \\
\midrule
\bottomrule
\end{tabular}

\end{table}

\subsection{Case study across models}
\label{sec:supp-cases}

Figures~\ref{fig:cases} to~\ref{fig:cases4} compare seven models on 14 sets of ViRA-Bench at 1\,MP.
Seven rows show all three versions of a set, and the other rows show one variant.

\paragraph{Heatmaps.}
For the five open models, a heatmap shows where the model looks in the shown image while it writes its final reply.
The shown image is the last crop that the model received, or its first view of the whole photo if it received no crop.
We replay the saved episode and take the attention from the tokens of the final reply to the $T$ tokens of this image, averaged over heads.
For a long reply, we use 256 evenly spaced tokens of it.
We use only the layers with full attention: 8 of the 32 layers in Qwen3.5-9B and ViRA-9B, 16 of the 64 layers in LBS-27B and ViRA-27B, and all 28 layers in DeepEyesV2.
As in~\cite{zhang2025mllmsknow}, we also give the model the same image alone with the request ``Write a general description of the image.'' and take the attention from the last token of this prompt to the image.
In each layer, we normalize both attentions to sum to one over the $T$ image tokens and divide the first by the second plus $1/T$.
The division removes image tokens that receive high attention under any request, and the term $1/T$ keeps the ratio bounded where the second attention is near zero.
The heatmap is the mean of this ratio over the layers.
We clip it at its 99th percentile, upsample and smooth it to the size of the image, and draw it with the jet colormap, with low values left transparent.
The proprietary models have no heatmap, because their attention is not available.

\paragraph{Cases.}
In Figures~\ref{fig:cases} and~\ref{fig:cases2}, baselines answer from common sense.
In Figure~\ref{fig:cases}a, three baselines answer ``basketball'' on all three versions, and in Figure~\ref{fig:cases}c, all five baselines give the original's answer.
Claude Haiku~5.5 and GPT-6-Luna do so although their last crops show the edited knob, so they find the evidence but do not read it.
On the variants in Figure~\ref{fig:cases}a, Qwen3.5-9B with the tool looks at the hoop and the net, whereas LBS-27B and both ViRA agents look at the ball.
Figure~\ref{fig:cases3}a and~b show the same for a traffic sign and an emblem.
In Figure~\ref{fig:cases3}c and~d, the proprietary models zoom up to six times without covering the object.
The other rows show errors of our agents.
In Figure~\ref{fig:cases3}e and Figure~\ref{fig:cases4}, ViRA-9B zooms once and gives the original's answer on both variants, whereas ViRA-27B zooms two to six times and solves the set.

\section{General benchmarks}
\label{sec:supp-beyond}
\label{sec:supp-single}

Table~\ref{tab:mmerw-max} and Table~\ref{tab:mmerw-1mp} split MME-RealWorld-Lite by task and domain.
ViRA-9B is ahead of Qwen3.5-9B with the zoom tool in every column at both resolutions.
The largest gain at each resolution is on monitoring (11.4 points at 1\,MP and 15.4 points at the maximum resolution).
On OCR and on diagrams and tables, where Qwen3.5-9B with the zoom tool is already at 80\% or more, the gains are 4.0 to 8.0 points.
ViRA-27B is ahead of every proprietary model on monitoring, in perception and in reasoning at both resolutions, by 4.7 to 8.0 points.

\begin{table*}[htbp]
\caption{\textbf{MME-RealWorld-Lite by task and domain at the maximum resolution}: accuracy (\%) of every model in Table~\ref{tab:general-full}.
The domains are OCR in complex context (OCR), remote sensing (RS), diagrams and tables (DT), monitoring (MO) and autonomous driving (AD).
Perception has 1,169 questions (OCR 250, RS 150, DT 100, MO 319, AD 350), and reasoning has 750 (OCR 100, DT 100, MO 150, AD 400).
\emph{All} is the accuracy on all questions of the task, and \emph{Overall} is the accuracy on all 1,919 questions, as in Table~\ref{tab:general-full}.
Tools, +\,tool, $\Delta$ and the colors are as in Table~\ref{tab:main}.
Bold: best in each column within each group.
Table~\ref{tab:ablation-mmerw} gives the ablations.}
\label{tab:mmerw-max}
\centering
\scriptsize
\setlength{\tabcolsep}{9.05pt}
\settowidth{\accw}{\textbf{00.0}}\settowidth{\tokw}{\tok{000K}}\settowidth{\numw}{\textbf{00.0}}\settowidth{\imgw}{(00.0)}
\begin{tabular}{l*{12}{c}}
\toprule
\midrule
 & \multicolumn{6}{c}{Perception} & \multicolumn{5}{c}{Reasoning} & \\
\cmidrule(lr){2-7}\cmidrule(lr){8-12}
Model at Max & OCR & RS & DT & MO & AD & All & OCR & DT & MO & AD & All & Overall \\
\midrule
\multicolumn{13}{l}{\emph{Open 7--9B models}} \\
Qwen3.5-9B & \nb{91.2} & \nb{58.7} & \nb{94.0} & \nb{40.8} & \nb{44.0} & \nb{59.4} & \nb{76.0} & \nb{79.0} & \nb{42.7} & \nb{38.2} & \nb{49.6} & \nb{55.5} \\
ZwZ2-9B & \nb{92.0} & \nb{60.7} & \nb{90.0} & \nb{48.0} & \nb{\textbf{55.4}} & \nb{64.8} & \nb{\textbf{86.0}} & \nb{\textbf{90.0}} & \nb{46.0} & \nb{43.2} & \nb{55.7} & \nb{61.3} \\
Vision-OPD-9B & \nb{94.0} & \nb{60.7} & \nb{\textbf{96.0}} & \nb{51.7} & \nb{53.7} & \nb{66.3} & \nb{82.0} & \nb{75.0} & \nb{\textbf{60.7}} & \nb{39.2} & \nb{54.0} & \nb{61.5} \\
\midrule[0.3pt]
\multicolumn{13}{l}{\emph{Open 7--9B models + tool}} \\
Qwen3.5-9B + tool & \nb{90.8} & \nb{59.3} & \nb{88.0} & \nb{40.4} & \nb{46.9} & \nb{59.6} & \nb{80.0} & \nb{82.0} & \nb{52.0} & \nb{41.8} & \nb{54.3} & \nb{57.5} \\
ZwZ2-9B + tool & \nb{92.4} & \nb{68.7} & \nb{94.0} & \nb{52.0} & \nb{55.1} & \nb{67.3} & \nb{\textbf{86.0}} & \nb{89.0} & \nb{56.0} & \nb{44.8} & \nb{58.4} & \nb{63.8} \\
Vision-OPD-9B + tool & \nb{92.0} & \nb{62.0} & \nb{90.0} & \nb{44.5} & \nb{43.7} & \nb{60.6} & \nb{83.0} & \nb{\textbf{90.0}} & \nb{48.0} & \nb{38.0} & \nb{52.9} & \nb{57.6} \\
\midrule[0.3pt]
\multicolumn{13}{l}{\emph{Released zoom agents (7--9B)}} \\
Mini-o3 & \nb{86.8} & \nb{44.7} & \nb{82.0} & \nb{38.2} & \nb{34.3} & \nb{52.0} & \nb{74.0} & \nb{71.0} & \nb{24.7} & \nb{30.2} & \nb{40.4} & \nb{47.5} \\
DeepEyes & \nb{90.0} & \nb{52.7} & \nb{88.0} & \nb{48.6} & \nb{39.4} & \nb{58.6} & \nb{82.0} & \nb{74.0} & \nb{42.7} & \nb{37.8} & \nb{49.5} & \nb{55.0} \\
DeepEyesV2 & \nb{87.6} & \nb{52.7} & \nb{88.0} & \nb{48.3} & \nb{34.3} & \nb{56.5} & \nb{73.0} & \nb{72.0} & \nb{45.3} & \nb{37.8} & \nb{48.5} & \nb{53.4} \\
Thyme & \nb{90.4} & \nb{54.7} & \nb{86.0} & \nb{47.6} & \nb{37.7} & \nb{58.0} & \nb{80.0} & \nb{72.0} & \nb{42.7} & \nb{36.2} & \nb{48.1} & \nb{54.1} \\
Pixel Reasoner & \nb{88.0} & \nb{49.3} & \nb{85.0} & \nb{42.3} & \nb{39.4} & \nb{55.8} & \nb{78.0} & \nb{78.0} & \nb{42.0} & \nb{36.2} & \nb{48.5} & \nb{52.9} \\
ZoomEye & \nb{87.2} & \nb{36.7} & \nb{88.0} & \nb{35.7} & \nb{34.3} & \nb{50.9} & \nb{72.0} & \nb{60.0} & \nb{34.0} & \nb{28.0} & \nb{39.3} & \nb{46.4} \\
\midrule[0.3pt]
\multicolumn{13}{l}{\emph{Ours + our zoom tool}} \\
\shaderow{vira9bg}\textbf{ViRA-9B} (ours) & \nb{\textbf{95.2}} & \nb{\textbf{71.3}} & \nb{\textbf{96.0}} & \nb{\textbf{55.8}} & \nb{52.9} & \nb{\textbf{68.8}} & \nb{85.0} & \nb{89.0} & \nb{\textbf{60.7}} & \nb{\textbf{46.0}} & \nb{\textbf{59.9}} & \nb{\textbf{65.3}} \\
\quad $\Delta$ vs.\ Qwen3.5-9B + tool & \dgain{+4.4} & \dgain{+12.0} & \dgain{+8.0} & \dgain{+15.4} & \dgain{+6.0} & \dgain{+9.2} & \dgain{+5.0} & \dgain{+7.0} & \dgain{+8.7} & \dgain{+4.2} & \dgain{+5.6} & \dgain{+7.8} \\
\midrule[1pt]
\multicolumn{13}{l}{\emph{Open 27B models}} \\
Qwen3.8-27B & \nb{92.4} & \nb{60.7} & \nb{89.0} & \nb{39.8} & \nb{50.0} & \nb{61.0} & \nb{86.0} & \nb{84.0} & \nb{50.7} & \nb{37.2} & \nb{52.7} & \nb{57.7} \\
ZwZ2-27B & \nb{93.6} & \nb{70.0} & \nb{93.0} & \nb{53.9} & \nb{\textbf{58.6}} & \nb{\textbf{69.2}} & \nb{90.0} & \nb{87.0} & \nb{\textbf{64.7}} & \nb{43.8} & \nb{59.9} & \nb{\textbf{65.6}} \\
\midrule[0.3pt]
\multicolumn{13}{l}{\emph{Open 27B models + tool}} \\
Qwen3.8-27B + tool & \nb{92.8} & \nb{53.3} & \nb{92.0} & \nb{45.8} & \nb{48.9} & \nb{61.7} & \nb{87.0} & \nb{89.0} & \nb{49.3} & \nb{44.0} & \nb{56.8} & \nb{59.8} \\
ZwZ2-27B + tool & \nb{93.2} & \nb{66.0} & \nb{94.0} & \nb{49.2} & \nb{55.7} & \nb{66.6} & \nb{89.0} & \nb{88.0} & \nb{56.7} & \nb{43.5} & \nb{58.1} & \nb{63.3} \\
\midrule[0.3pt]
\multicolumn{13}{l}{\emph{Released zoom agents (27--32B)}} \\
LBS-27B & \nb{94.0} & \nb{71.3} & \nb{\textbf{95.0}} & \nb{\textbf{61.1}} & \nb{49.1} & \nb{68.8} & \nb{\textbf{91.0}} & \nb{89.0} & \nb{57.3} & \nb{\textbf{46.0}} & \nb{\textbf{60.0}} & \nb{65.3} \\
SenseNova-MARS-32B & \nb{94.0} & \nb{70.7} & \nb{93.0} & \nb{47.6} & \nb{53.4} & \nb{66.1} & \nb{87.0} & \nb{\textbf{91.0}} & \nb{47.3} & \nb{38.5} & \nb{53.7} & \nb{61.3} \\
\midrule[0.3pt]
\multicolumn{13}{l}{\emph{Ours + our zoom tool}} \\
\shaderow{vira27bg}\textbf{ViRA-27B} (ours) & \nb{\textbf{95.6}} & \nb{\textbf{74.0}} & \nb{92.0} & \nb{54.9} & \nb{49.4} & \nb{67.6} & \nb{87.0} & \nb{89.0} & \nb{60.7} & \nb{45.0} & \nb{59.6} & \nb{64.5} \\
\quad $\Delta$ vs.\ Qwen3.8-27B + tool & \dgain{+2.8} & \dgain{+20.7} & \dgain{+0.0} & \dgain{+9.1} & \dgain{+0.5} & \dgain{+5.9} & \dgain{+0.0} & \dgain{+0.0} & \dgain{+11.4} & \dgain{+1.0} & \dgain{+2.8} & \dgain{+4.7} \\
\midrule[1pt]
\multicolumn{13}{l}{\emph{Proprietary models + our zoom tool}} \\
\cstrut Claude Haiku~5.5, max thinking & \cellcolor{gainc!10}\nb{\textbf{93.2}} & \cellcolor{gainc!11}\nb{70.7} & \nb{92.0} & \cellcolor{gainc!21}\nb{45.5} & \cellcolor{gainc!19}\nb{41.4} & \cellcolor{gainc!15}\nb{61.7} & \cellcolor{gainc!8}\nb{86.0} & \cellcolor{lossc!11}\nb{\textbf{92.0}} & \cellcolor{gainc!27}\nb{47.3} & \cellcolor{lossc!6}\nb{45.2} & \cellcolor{gainc!10}\nb{57.3} & \cellcolor{gainc!13}\nb{60.0} \\
\cstrut Claude Haiku~5.5, medium thinking & \cellcolor{gainc!11}\nb{92.4} & \nb{\textbf{74.0}} & \cellcolor{lossc!8}\nb{93.0} & \cellcolor{gainc!14}\nb{\textbf{49.8}} & \cellcolor{gainc!14}\nb{44.0} & \cellcolor{gainc!12}\nb{\textbf{64.0}} & \nb{87.0} & \cellcolor{lossc!9}\nb{91.0} & \cellcolor{gainc!13}\nb{\textbf{56.0}} & \cellcolor{lossc!7}\nb{\textbf{45.5}} & \cellcolor{gainc!7}\nb{\textbf{59.2}} & \cellcolor{gainc!10}\nb{\textbf{62.1}} \\
\cstrut Claude Haiku~5.5, no thinking & \cellcolor{gainc!11}\nb{92.4} & \cellcolor{gainc!21}\nb{64.7} & \nb{92.0} & \cellcolor{gainc!19}\nb{46.7} & \cellcolor{gainc!21}\nb{40.0} & \cellcolor{gainc!17}\nb{60.7} & \nb{87.0} & \cellcolor{lossc!9}\nb{91.0} & \cellcolor{gainc!21}\nb{51.3} & \cellcolor{gainc!11}\nb{41.5} & \cellcolor{gainc!11}\nb{56.1} & \cellcolor{gainc!15}\nb{58.9} \\
\cstrut GPT-6-Luna, max thinking & \cellcolor{gainc!12}\nb{91.6} & \cellcolor{gainc!30}\nb{58.7} & \cellcolor{lossc!11}\nb{\textbf{95.0}} & \cellcolor{gainc!30}\nb{39.8} & \cellcolor{gainc!25}\nb{37.1} & \cellcolor{gainc!22}\nb{57.2} & \cellcolor{gainc!12}\nb{83.0} & \cellcolor{lossc!11}\nb{\textbf{92.0}} & \cellcolor{gainc!30}\nb{36.0} & \cellcolor{gainc!11}\nb{42.0} & \cellcolor{gainc!17}\nb{52.9} & \cellcolor{gainc!20}\nb{55.5} \\
\cstrut GPT-6-Luna, medium thinking & \cellcolor{gainc!12}\nb{92.0} & \cellcolor{gainc!30}\nb{53.3} & \cellcolor{lossc!8}\nb{93.0} & \cellcolor{gainc!14}\nb{49.5} & \cellcolor{gainc!10}\nb{\textbf{46.9}} & \cellcolor{gainc!15}\nb{62.0} & \cellcolor{lossc!8}\nb{\textbf{88.0}} & \cellcolor{lossc!8}\nb{90.0} & \cellcolor{gainc!24}\nb{49.3} & \cellcolor{gainc!8}\nb{43.8} & \cellcolor{gainc!10}\nb{56.9} & \cellcolor{gainc!13}\nb{60.0} \\
\cstrut GPT-6-Luna, no thinking & \cellcolor{gainc!14}\nb{90.4} & \cellcolor{gainc!30}\nb{37.3} & \cellcolor{gainc!12}\nb{88.0} & \cellcolor{gainc!30}\nb{36.7} & \cellcolor{gainc!30}\nb{32.9} & \cellcolor{gainc!30}\nb{51.5} & \cellcolor{gainc!23}\nb{76.0} & \cellcolor{gainc!14}\nb{84.0} & \cellcolor{gainc!30}\nb{38.0} & \cellcolor{gainc!15}\nb{39.2} & \cellcolor{gainc!21}\nb{49.9} & \cellcolor{gainc!27}\nb{50.9} \\
\midrule
\bottomrule
\end{tabular}
\end{table*}

\begin{table*}[htbp]
\caption{\textbf{MME-RealWorld-Lite by task and domain at 1\,MP}: accuracy (\%).
The columns, rows, colors and bold are as in Table~\ref{tab:mmerw-max}.}
\label{tab:mmerw-1mp}
\centering
\scriptsize
\setlength{\tabcolsep}{9.05pt}
\settowidth{\accw}{\textbf{00.0}}\settowidth{\tokw}{\tok{000K}}\settowidth{\numw}{\textbf{00.0}}\settowidth{\imgw}{(00.0)}
\begin{tabular}{l*{12}{c}}
\toprule
\midrule
 & \multicolumn{6}{c}{Perception} & \multicolumn{5}{c}{Reasoning} & \\
\cmidrule(lr){2-7}\cmidrule(lr){8-12}
Model at 1\,MP & OCR & RS & DT & MO & AD & All & OCR & DT & MO & AD & All & Overall \\
\midrule
\multicolumn{13}{l}{\emph{Open 7--9B models}} \\
Qwen3.5-9B & \nb{85.2} & \nb{50.7} & \nb{83.0} & \nb{39.8} & \nb{43.4} & \nb{55.7} & \nb{76.0} & \nb{72.0} & \nb{42.7} & \nb{37.5} & \nb{48.3} & \nb{52.8} \\
ZwZ2-9B & \nb{86.4} & \nb{55.3} & \nb{82.0} & \nb{44.5} & \nb{50.3} & \nb{59.8} & \nb{82.0} & \nb{76.0} & \nb{46.0} & \nb{42.2} & \nb{52.8} & \nb{57.1} \\
Vision-OPD-9B & \nb{88.0} & \nb{56.7} & \nb{84.0} & \nb{48.0} & \nb{\textbf{53.4}} & \nb{62.4} & \nb{79.0} & \nb{66.0} & \nb{50.7} & \nb{38.8} & \nb{50.1} & \nb{57.6} \\
\midrule[0.3pt]
\multicolumn{13}{l}{\emph{Open 7--9B models + tool}} \\
Qwen3.5-9B + tool & \nb{89.6} & \nb{52.0} & \nb{86.0} & \nb{42.0} & \nb{41.4} & \nb{57.1} & \nb{82.0} & \nb{81.0} & \nb{45.3} & \nb{39.5} & \nb{51.9} & \nb{55.0} \\
ZwZ2-9B + tool & \nb{92.4} & \nb{\textbf{62.7}} & \nb{89.0} & \nb{\textbf{52.7}} & \nb{50.0} & \nb{64.8} & \nb{84.0} & \nb{\textbf{86.0}} & \nb{50.0} & \nb{43.2} & \nb{55.7} & \nb{61.2} \\
Vision-OPD-9B + tool & \nb{88.8} & \nb{52.0} & \nb{86.0} & \nb{42.3} & \nb{45.4} & \nb{58.2} & \nb{79.0} & \nb{79.0} & \nb{44.7} & \nb{39.5} & \nb{51.1} & \nb{55.4} \\
\midrule[0.3pt]
\multicolumn{13}{l}{\emph{Released zoom agents (7--9B)}} \\
Mini-o3 & \nb{84.8} & \nb{46.7} & \nb{68.0} & \nb{34.8} & \nb{32.9} & \nb{49.3} & \nb{69.0} & \nb{68.0} & \nb{25.3} & \nb{28.2} & \nb{38.4} & \nb{45.0} \\
DeepEyes & \nb{83.6} & \nb{52.0} & \nb{80.0} & \nb{39.8} & \nb{34.6} & \nb{52.6} & \nb{69.0} & \nb{60.0} & \nb{42.7} & \nb{34.8} & \nb{44.3} & \nb{49.3} \\
DeepEyesV2 & \nb{84.0} & \nb{48.0} & \nb{83.0} & \nb{37.0} & \nb{32.0} & \nb{50.9} & \nb{79.0} & \nb{63.0} & \nb{42.7} & \nb{37.2} & \nb{47.3} & \nb{49.5} \\
Thyme & \nb{83.6} & \nb{56.7} & \nb{79.0} & \nb{41.7} & \nb{32.9} & \nb{53.1} & \nb{77.0} & \nb{67.0} & \nb{46.7} & \nb{31.8} & \nb{45.5} & \nb{50.1} \\
Pixel Reasoner & \nb{82.8} & \nb{48.7} & \nb{84.0} & \nb{35.1} & \nb{35.1} & \nb{51.2} & \nb{74.0} & \nb{64.0} & \nb{40.7} & \nb{34.8} & \nb{45.1} & \nb{48.8} \\
ZoomEye & \nb{82.8} & \nb{29.3} & \nb{83.0} & \nb{32.0} & \nb{32.6} & \nb{47.0} & \nb{64.0} & \nb{54.0} & \nb{34.7} & \nb{26.8} & \nb{36.9} & \nb{43.1} \\
\midrule[0.3pt]
\multicolumn{13}{l}{\emph{Ours + our zoom tool}} \\
\shaderow{vira9bg}\textbf{ViRA-9B} (ours) & \nb{\textbf{94.4}} & \nb{61.3} & \nb{\textbf{91.0}} & \nb{\textbf{52.7}} & \nb{52.3} & \nb{\textbf{65.9}} & \nb{\textbf{86.0}} & \nb{85.0} & \nb{\textbf{56.7}} & \nb{\textbf{48.5}} & \nb{\textbf{60.0}} & \nb{\textbf{63.6}} \\
\quad $\Delta$ vs.\ Qwen3.5-9B + tool & \dgain{+4.8} & \dgain{+9.3} & \dgain{+5.0} & \dgain{+10.7} & \dgain{+10.9} & \dgain{+8.8} & \dgain{+4.0} & \dgain{+4.0} & \dgain{+11.4} & \dgain{+9.0} & \dgain{+8.1} & \dgain{+8.6} \\
\midrule[1pt]
\multicolumn{13}{l}{\emph{Open 27B models}} \\
Qwen3.8-27B & \nb{82.8} & \nb{50.7} & \nb{85.0} & \nb{41.7} & \nb{48.3} & \nb{57.3} & \nb{81.0} & \nb{75.0} & \nb{48.0} & \nb{37.0} & \nb{50.1} & \nb{54.5} \\
ZwZ2-27B & \nb{88.4} & \nb{63.3} & \nb{93.0} & \nb{49.5} & \nb{\textbf{56.3}} & \nb{65.4} & \nb{84.0} & \nb{83.0} & \nb{\textbf{58.7}} & \nb{41.5} & \nb{56.1} & \nb{61.8} \\
\midrule[0.3pt]
\multicolumn{13}{l}{\emph{Open 27B models + tool}} \\
Qwen3.8-27B + tool & \nb{93.2} & \nb{49.3} & \nb{88.0} & \nb{42.3} & \nb{42.9} & \nb{58.2} & \nb{81.0} & \nb{85.0} & \nb{46.0} & \nb{38.0} & \nb{51.6} & \nb{55.6} \\
ZwZ2-27B + tool & \nb{94.0} & \nb{62.7} & \nb{91.0} & \nb{48.6} & \nb{54.0} & \nb{65.4} & \nb{89.0} & \nb{87.0} & \nb{50.7} & \nb{45.8} & \nb{58.0} & \nb{62.5} \\
\midrule[0.3pt]
\multicolumn{13}{l}{\emph{Released zoom agents (27--32B)}} \\
LBS-27B & \nb{93.6} & \nb{66.7} & \nb{92.0} & \nb{\textbf{53.6}} & \nb{52.0} & \nb{\textbf{66.6}} & \nb{\textbf{90.0}} & \nb{86.0} & \nb{58.0} & \nb{45.2} & \nb{59.2} & \nb{\textbf{63.7}} \\
SenseNova-MARS-32B & \nb{\textbf{94.4}} & \nb{64.0} & \nb{93.0} & \nb{43.9} & \nb{50.9} & \nb{63.6} & \nb{89.0} & \nb{\textbf{90.0}} & \nb{46.0} & \nb{39.5} & \nb{54.1} & \nb{59.9} \\
\midrule[0.3pt]
\multicolumn{13}{l}{\emph{Ours + our zoom tool}} \\
\shaderow{vira27bg}\textbf{ViRA-27B} (ours) & \nb{\textbf{94.4}} & \nb{\textbf{68.0}} & \nb{\textbf{94.0}} & \nb{53.0} & \nb{48.6} & \nb{66.0} & \nb{85.0} & \nb{89.0} & \nb{53.3} & \nb{\textbf{47.8}} & \nb{\textbf{59.3}} & \nb{63.4} \\
\quad $\Delta$ vs.\ Qwen3.8-27B + tool & \dgain{+1.2} & \dgain{+18.7} & \dgain{+6.0} & \dgain{+10.7} & \dgain{+5.7} & \dgain{+7.8} & \dgain{+4.0} & \dgain{+4.0} & \dgain{+7.3} & \dgain{+9.8} & \dgain{+7.7} & \dgain{+7.8} \\
\midrule[1pt]
\multicolumn{13}{l}{\emph{Proprietary models + our zoom tool}} \\
\cstrut Claude Haiku~5.5, max thinking & \cellcolor{gainc!9}\nb{92.8} & \cellcolor{gainc!18}\nb{61.3} & \cellcolor{gainc!11}\nb{91.0} & \cellcolor{gainc!30}\nb{39.8} & \cellcolor{gainc!22}\nb{40.0} & \cellcolor{gainc!20}\nb{58.3} & \cellcolor{gainc!8}\nb{84.0} & \cellcolor{lossc!10}\nb{91.0} & \cellcolor{gainc!21}\nb{\textbf{45.3}} & \cellcolor{gainc!8}\nb{46.8} & \cellcolor{gainc!10}\nb{57.3} & \cellcolor{gainc!16}\nb{57.9} \\
\cstrut Claude Haiku~5.5, medium thinking & \cellcolor{gainc!13}\nb{90.4} & \cellcolor{gainc!7}\nb{67.3} & \cellcolor{gainc!10}\nb{92.0} & \cellcolor{gainc!22}\nb{44.2} & \cellcolor{gainc!21}\nb{40.3} & \cellcolor{gainc!17}\nb{60.0} & \cellcolor{lossc!10}\nb{87.0} & \cellcolor{lossc!8}\nb{90.0} & \cellcolor{gainc!24}\nb{43.3} & \cellcolor{gainc!13}\nb{43.8} & \cellcolor{gainc!13}\nb{55.6} & \cellcolor{gainc!15}\nb{58.3} \\
\cstrut Claude Haiku~5.5, no thinking & \cellcolor{gainc!12}\nb{91.2} & \cellcolor{gainc!22}\nb{59.3} & \cellcolor{gainc!10}\nb{92.0} & \cellcolor{gainc!30}\nb{39.5} & \cellcolor{gainc!21}\nb{40.3} & \cellcolor{gainc!21}\nb{57.8} & \cellcolor{lossc!11}\nb{\textbf{88.0}} & \cellcolor{lossc!8}\nb{90.0} & \cellcolor{gainc!22}\nb{44.7} & \cellcolor{gainc!15}\nb{43.0} & \cellcolor{gainc!13}\nb{55.6} & \cellcolor{gainc!18}\nb{57.0} \\
\cstrut GPT-6-Luna, max thinking & \cellcolor{gainc!7}\nb{\textbf{94.0}} & \cellcolor{lossc!15}\nb{\textbf{72.7}} & \cellcolor{gainc!8}\nb{\textbf{93.0}} & \cellcolor{gainc!29}\nb{40.1} & \cellcolor{gainc!24}\nb{38.9} & \cellcolor{gainc!17}\nb{60.0} & \cellcolor{lossc!10}\nb{87.0} & \cellcolor{lossc!13}\nb{\textbf{93.0}} & \cellcolor{gainc!30}\nb{34.0} & \cellcolor{gainc!14}\nb{43.5} & \cellcolor{gainc!16}\nb{54.0} & \cellcolor{gainc!17}\nb{57.6} \\
\cstrut GPT-6-Luna, medium thinking & \cellcolor{gainc!10}\nb{92.0} & \cellcolor{lossc!7}\nb{68.7} & \cellcolor{gainc!11}\nb{91.0} & \cellcolor{gainc!15}\nb{\textbf{48.3}} & \cellcolor{lossc!7}\nb{\textbf{48.9}} & \cellcolor{gainc!9}\nb{\textbf{64.1}} & \cellcolor{lossc!11}\nb{\textbf{88.0}} & \cellcolor{lossc!13}\nb{\textbf{93.0}} & \cellcolor{gainc!21}\nb{\textbf{45.3}} & \nb{\textbf{47.8}} & \cellcolor{gainc!7}\nb{\textbf{58.7}} & \cellcolor{gainc!9}\nb{\textbf{62.0}} \\
\cstrut GPT-6-Luna, no thinking & \cellcolor{gainc!18}\nb{88.0} & \cellcolor{gainc!30}\nb{50.0} & \cellcolor{gainc!30}\nb{78.0} & \cellcolor{gainc!30}\nb{28.8} & \cellcolor{gainc!24}\nb{38.9} & \cellcolor{gainc!30}\nb{51.4} & \cellcolor{gainc!19}\nb{78.0} & \cellcolor{gainc!24}\nb{79.0} & \cellcolor{gainc!30}\nb{36.7} & \cellcolor{gainc!23}\nb{38.5} & \cellcolor{gainc!25}\nb{48.8} & \cellcolor{gainc!30}\nb{50.4} \\
\midrule
\bottomrule
\end{tabular}
\end{table*}

On the six benchmarks with two resolutions, ZoomEye uses 65K to 425K tokens per question, which is 4 to 28 times as many as ViRA-9B.

\end{document}